\documentclass[iicol,pdflatex,sn-mathphys-num]{sn-jnl}% Math and Physical Sciences Numbered Reference Style 

\usepackage{graphicx}%
\usepackage{multirow}%
\usepackage{amsmath,amssymb,amsfonts}%
\usepackage{amsthm}%
\usepackage{mathrsfs}%
\usepackage[title]{appendix}%
\usepackage{xcolor}%
\usepackage{textcomp}%
\usepackage{manyfoot}%
\usepackage{booktabs}%
\usepackage{algorithm}%
\usepackage{algorithmicx}%
\usepackage{algpseudocode}%
\usepackage{listings}%
\usepackage{amsmath}
\usepackage{xspace}
\usepackage{graphicx}
\usepackage{epstopdf}
\usepackage{amsmath}
\usepackage{algorithm}
\usepackage{algorithmicx}
\usepackage{algpseudocode}
\usepackage{url}
\usepackage{hyperref}
\usepackage[table,xcdraw]{xcolor}
\usepackage{caption}
\usepackage{booktabs} % 推荐使用，但不是必须
\usepackage{array}

\usepackage{array}
\usepackage{booktabs}
\usepackage{multirow}
\usepackage{caption}
\usepackage{colortbl} % 如果需要

\usepackage{graphicx}%
\usepackage{multirow}%
\usepackage{amsmath,amssymb,amsfonts}%
\usepackage{amsthm}%
\usepackage{mathrsfs}%
\usepackage[title]{appendix}%
\usepackage{xcolor}%
\usepackage{textcomp}%
\usepackage{manyfoot}%
\usepackage{booktabs}%
\usepackage{algorithm}%
\usepackage{algorithmicx}%
\usepackage{algpseudocode}%
\usepackage{listings}%
\usepackage{amsmath}
\usepackage{xspace}
\usepackage{graphicx}
\usepackage{epstopdf}
\usepackage{amsmath}
\usepackage{algorithm}
\usepackage{algorithmicx}
\usepackage{algpseudocode}
\usepackage{url}
\usepackage{hyperref}
\usepackage{adjustbox}
\usepackage{graphicx}
\usepackage{lineno}

\newcommand{\benchname}{\textcolor{black}{I$^2$EBench2.0}\xspace}

\usepackage[table,xcdraw]{xcolor}
\usepackage{caption}

\newcommand{\addblankpages}[1]{
    \ifnum\value{page}>1\newpage\fi
    \begingroup
    \pagestyle{empty}
    \count0=#1
    \loop
        \ifnum\count0>0
            \null
            \newpage
            \advance\count0 by -1
    \repeat
    \endgroup
}

\theoremstyle{thmstyleone}%
\theoremstyle{thmstyletwo}%

\theoremstyle{thmstylethree}%

\begin{document}
% \linenumbers

\title[]{An Extensive Benchmark for Single-round and Multi-round Instruction-based Image Editing}

%%  \sfx{IV}}
%%=============================================================%%

\author[1]{\fnm{Yiwei} \sur{Ma}}
\equalcont{These authors contributed equally to this work.}

\author[1]{\fnm{Ke} \sur{Ye}}
\equalcont{These authors contributed equally to this work.}

\author[1]{\fnm{Weihuang} \sur{Lin}}
\equalcont{These authors contributed equally to this work.}

\author[12]{\fnm{Jiayi} \sur{Ji}}

\author*[1]{\fnm{Xiaoshuai} \sur{Sun}}

\author[2]{\fnm{Tat-Seng} \sur{Chua}}

\author[1]{\fnm{Rongrong} \sur{Ji}}

\affil[1]{\orgdiv{Key Laboratory of Multimedia Trusted Perception and Efficient Computing}, \orgname{Ministry of Education of China, Xiamen University}, \orgaddress{ \postcode{361005}, \state{Fujian}, \country{P.R. China}}}

\affil[2]{ \orgname{National University of Singapore},\orgaddress{ \postcode{119077}, \state{Singapore}}}

%% Corresponding-author note without email (arxiv version)
\makeatletter
\newcommand{\corrnote}[1]{\gdef\@corrnote{#1}}
\corrnote{Corresponding author.}
\makeatother

% \affil[4]{\orgname{National Tsing Hua University}, \orgaddress{\country{Taiwan}}}

% \affil[3]{\orgdiv{Department}, \orgname{Organization}, \orgaddress{\street{Street}, \city{City}, \postcode{610101}, \state{State}, \country{Country}}}


\abstract{

In recent years, there have been notable advancements in the area of instruction-based image editing (IIE), which focuses on the automatic alteration of input images using a model.
Nevertheless, assessing the effectiveness of these editing models poses a considerable challenge due to the intricate nature of instructions and the wide variety of edits.
To tackle this problem, one urgent task in this domain is the development of a robust evaluation framework that can precisely gauge the quality of editing outcomes and offer valuable benchmarks to guide future improvements.
To address this challenge, we present a comprehensive evaluation benchmark named \textit{\benchname}, designed for single-round and multi-round assessment of IIE models.
\benchname has four key features:
1) \textit{Evaluation Across Single and Multi-rounds}: \benchname simultaneously evaluates both single-round and multi-round instruction-based edits, assessing the precision and consistency of the edits.
2) \textit{Extensive Evaluation Criteria}: \benchname encompasses a broad range of criteria, evaluating both high-level and low-level aspects of each IIE model. Specifically, it incorporates 16 dimensions for single-round evaluations and 7 for multi-round evaluations
3) \textit{Alignment with Human Judgment}: To ensure our benchmark aligns with human evaluation, we conducted a comprehensive user study for each criterion.
4) \textit{Research-driven Insights}: By analyzing the strengths and weaknesses of current IIE models across all 16 single-round and 7 multi-round dimensions, we provide critical insights aimed at directing future research in this area.
We tested eight recently developed IIE models using \benchname and derived academic insights through meticulous comparison and analysis.
The related code, dataset, and images generated by all IIE models are available on GitHub: \textcolor{black}{\url{https://github.com/cocoshe/I2EBench}}.

}




\maketitle

\begin{figure*}
  \centering
  \includegraphics[width=1.5\columnwidth]{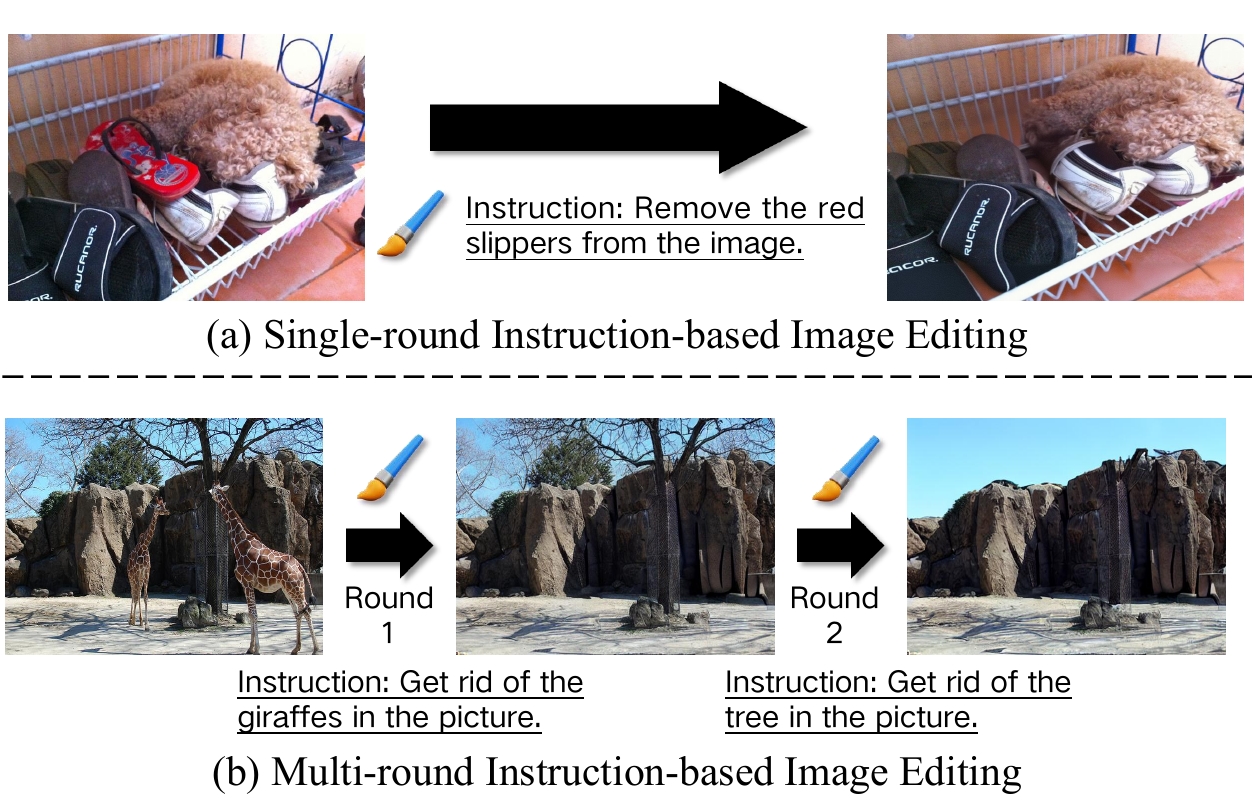}
  % \vspace{-1em}
  \caption{
    \textcolor{black}{A comparison between single-round and multi-round instruction-based image editing. Multi-round editing is crucial as it demonstrates the stability and accuracy of the editing process. Therefore, we have enhanced \benchname to include the multi-round editing evaluation, aimed at testing the accuracy and robustness of the IIE model across multiple rounds of editing.}
    }
  \label{fig:intro}
  % \vspace{-1em}
\end{figure*}

\section{Introduction}
% P1: IIE很重要，评估IIE也很重要
Instruction-based Image Editing  (IIE)\cite{brooks2023instructpix2pix,geng2023instructdiffusion,zhang2024magicbrush,li2023instructany2pix,wang2023instructedit,zhang2023hive,fu2023guiding} represents a transformative approach whereby images are modified based on textual instructions, offering an accessible interface for users within the community.
In recent years, substantial progress has been achieved in the domain of IIE, primarily fueled by the introduction of diffusion model~\cite{ho2020denoising,sohl2015deep,welling2011bayesian,kulikov2023sinddm} as well as large vision-language models (LVLMs)~\cite{liu2023improvedllava,liu2023llava,fei2024enhancing,fei2024video,fei2024vitron,ma2024inf}.
Despite these advancements, it is critical to develop a comprehensive benchmark capable of effectively evaluating the performance of such models.
An optimal evaluation scheme should encompass multifaceted assessments of editing quality while maintaining congruence with human perceptual standards to ensure reliability.
Furthermore, the framework should be adept at addressing both single-round and multi-round editing processes.
Importantly, the evaluation system must elucidate the distinctive strengths and weaknesses inherent in each model, thereby furnishing valuable insights to guide subsequent research endeavors in data selection, training strategy formulation, and architectural design within the field.
However, the challenge of evaluating IIE models is compounded by the diversity of editing tasks and the inherent complexity of measuring the degree of correspondence between modified images and their textual directives.

% P2: 现有的指标不好
Current IIE evaluation methodologies can be divided into three primary categories: 1) \textit{traditional metrics}; 2) \textit{user studies}; and 3) \textit{specialized benchmarks}.
The first category~\cite{brooks2023instructpix2pix,geng2023instructdiffusion,zhang2024magicbrush,li2023instructany2pix,wang2023instructedit,huang2023smartedit} utilizes standard metrics to assess the performance of IIE models. These include CLIP Score~\cite{radford2021learning}, CLIP Text-Image Direction Similarity~\cite{radford2021learning}, PSNR~\cite{korhonen2012peak}, SSIM~\cite{wang2004image}, and LPIPS~\cite{zhang2018unreasonable}. The simplicity of this method is its main advantage. Nonetheless, relying on a single metric is insufficient for evaluating the wide variety of editing tasks. For instance, while the CLIP score is adept at gauging the similarity between images and text, it may not be applicable for low-level visual editing tasks like denoising and low-light enhancement. Conversely, PSNR, which assesses image similarity, proves inadequate for high-level editing tasks such as object removal and replacement.
The second category~\cite{li2023instructany2pix,zhang2023hive,fu2023guiding} engages human participants to rate the effectiveness of different editing methodologies. This approach directly captures human preferences and aligns outcomes with human perceptual standards. However, it incurs high costs and suffers from a lack of reproducibility, as the test sets and participant samples may vary across evaluations.
The third category involves benchmarks~\cite{kawar2023imagic,wang2023imagen,basu2023editval,huang2024diffusion} explicitly designed to assess IIE models. Despite being tailored for this purpose, current benchmarks exhibit certain limitations. For example, TedBench~\cite{kawar2023imagic} evaluates only 100 images featuring common editing types, which may not fully encapsulate the models' capabilities. EditBench~\cite{wang2023imagen} is concentrated on mask-guided editing, rendering it ineffective for evaluating techniques that do not use masks. In the context of EditVal~\cite{basu2023editval}, only a limited range of dimensions concerning the size or location can be automatically assessed, restricting its general applicability. Furthermore, these benchmarks predominantly focus on single-round instruction-based editing, overlooking the iterative, multiple-round editing common in real-world applications.

\begin{figure*}
  \centering
  \includegraphics[width=2.0\columnwidth]{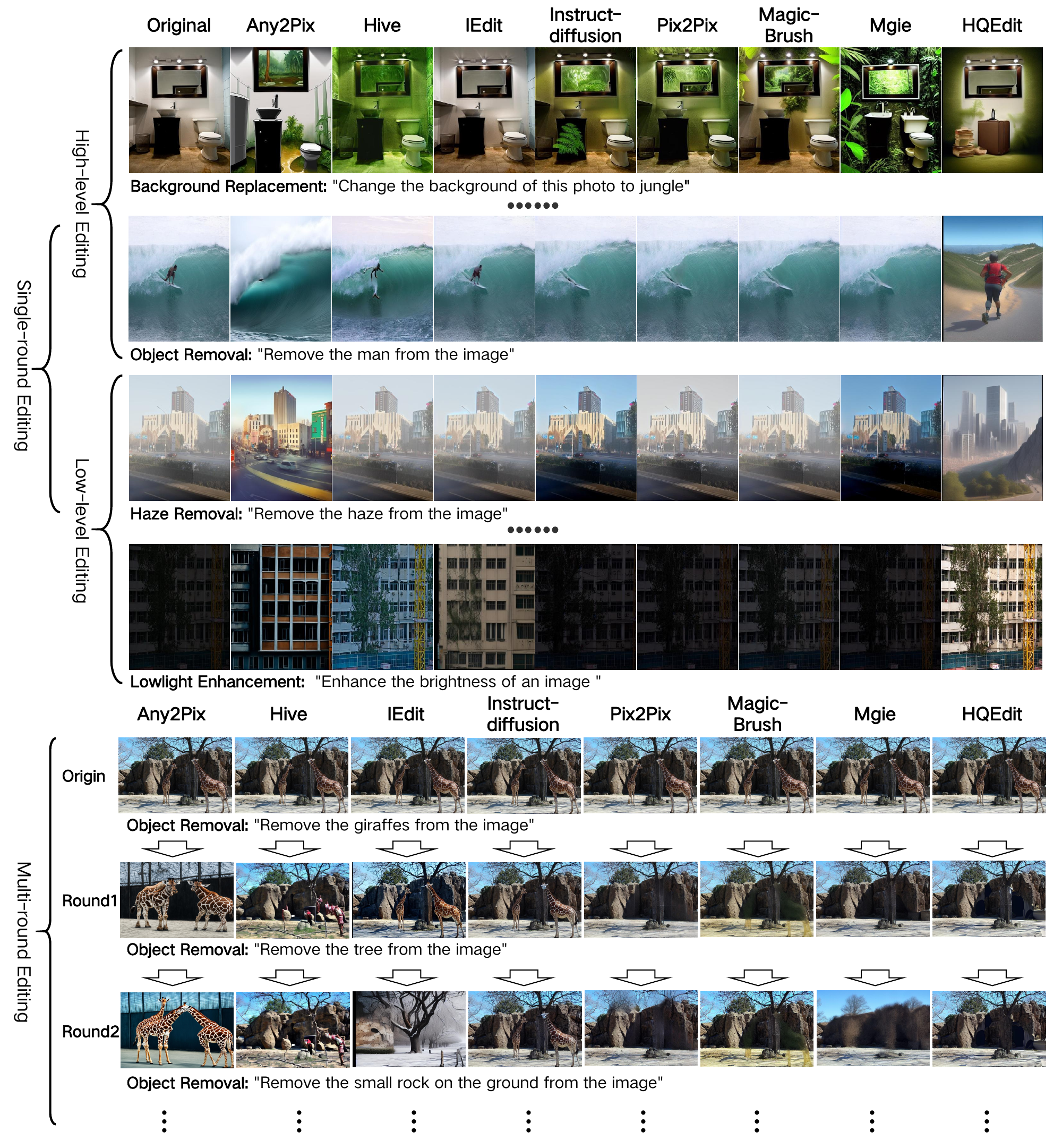}
  % \vspace{-1em}
  \caption{
    Visualization of editing results of single-round and multi-round editing across 16 proposed evaluation dimensions using various instruction-based image editing (IIE) models. The models include InstructAny2Pix~\cite{li2023instructany2pix}, HIVE~\cite{zhang2023hive}, InstructEdit~\cite{wang2023instructedit}, InstructDiffusion~\cite{geng2023instructdiffusion}, InstructPix2Pix~\cite{brooks2023instructpix2pix}, MagicBrush~\cite{zhang2024magicbrush}, MGIE~\cite{fu2023guiding}, and HQEdit~\cite{hui2024hq}.
    }
  \label{fig:res_sample}
  % \vspace{-1em}
\end{figure*}

% P3: 因此，我们提出了I2EBench
In this study, we introduce \textit{\benchname}, a comprehensive benchmark crafted to automatically assess IIE models' efficacy in single-round and multi-round instruction-based image editing. \benchname is distinguished by two key features: \textit{Multi-round Evaluation} and \textit{Comprehensive Dimensionality}.
For the first feature, the aspect of multi-round evaluation is crucial yet frequently overlooked in existing evaluations, as depicted in Fig.~\ref{fig:intro}. Multi-round editing necessitates modifications based on the outcomes of prior rounds, demanding high stability. Current benchmarks~\cite{kawar2023imagic,wang2023imagen,basu2023editval,zhang2024magicbrush,sheynin2023emu,huang2023smartedit} fail to address the unique demands associated with evaluating multiple editing rounds. Our proposed \benchname framework assesses both single and multi-round image editing capabilities of existing IIE models.
For the second feature, the focus on comprehensive dimensions is designed to capture as many editing types as possible, thereby gauging the robustness and transferability of models across varied editing scenarios. Consequently, \benchname encompasses 16 distinct editing dimensions, covering both high-level and low-level tasks. As depicted in Fig.~\ref{fig:res_sample}, high-level editing emphasizes the comprehension of instructions and the modification of specific image regions, while low-level editing targets image detail refinement or modifications affecting the entire image. Both high-level and low-level categories consist of 8 granular dimensions, which collectively assess the model's adeptness in nuanced editing tasks.
Utilizing the proposed \benchname, we assessed the multi-dimensional performance of eight recently published IIE models. Additionally, to ensure that \benchname scores are reflective of human perception, a consistency analysis was conducted comparing benchmark scores with human ratings. Our comprehensive analysis revealed a significant correlation between the two, providing strong evidence that our evaluation methodology is aligned with human judgment. Finally, by facilitating a thorough assessment of existing models across multi-rounds, multi-dimensions, and multi-categories, \benchname uncovers valuable insights into their individual strengths and weaknesses. These findings serve as strategic guidance for improving architectural designs, refining data selection techniques, and ultimately enhancing the quality of editing results.

This paper presents a significantly expanded edition of our previous work presented at the conference~\cite{ma2024i2ebench}. Compared to the conference version, this journal article incorporates several major enhancements:
(1) Most notably, we have developed a comprehensive dataset specifically for evaluating multi-round instruction-based image editing, thereby broadening \benchname's capacity to assess multi-round editing scenarios.
(2) We performed an in-depth aesthetic quality analysis for each dimension, examining how different editing dimensions influence aesthetic outcomes.
(3) Additional experimental analyses and discussions have been included, such as a detailed examination of the results from multi-round editing experiments.

In summary, our primary contributions are as follows:
\begin{itemize}
    
    \item We have developed a benchmark for both single-round and multi-round instruction-driven image editing, encompassing a wide range of evaluation dimensions. This was achieved by creating a dataset containing over 2,000 images and more than 6,700 editing instructions, which address both single and multi-round editing scenarios.
    
    \item We tested eight recently released models on the proposed \benchname to assess their performance. Furthermore, we performed a consistency analysis comparing the benchmark evaluations with human assessments and observed that the benchmark's scores were well-aligned with human preferences.
    
    \item Through an analysis of the strengths and weaknesses of current IIE models for single-round and multi-round editing across the 16 dimensions, we provided valuable research insights to inform future advancements in the field.
    
\end{itemize}

\section{Related Work}

\subsection{Instruction-based Image Editing}

Instruction-based Image Editing (IIE) represents a burgeoning field that seeks to simplify the process of image modification through the use of natural language instructions. This approach leverages advancements in both Generative Adversarial Networks (GANs)~\cite{goodfellow2014generative,karras2019style} and Diffusion models~\cite{song2020denoising,ho2020denoising}, which have enabled more nuanced and contextually relevant image synthesis and transformation.
At the core of IIE is the notion of translating text-based instructions into precise image transformations. Early attempts at text-to-image generation focused heavily on aligning generated visuals with detailed textual descriptions~\cite{saharia2022photorealistic,ramesh2021zero}. However, these models often required comprehensive descriptions, posing challenges in the ease of use and accessibility, especially for casual users seeking to modify images without detailed instruction.
To address these accessibility issues, Prompt-based Image Editing (PIE) models emerged~\cite{avrahami2022blended}. PIE facilitates the editing of images based on more concise target prompts, requiring users to provide both the original image and a brief description of the desired outcome. This method, however, still necessitates users to articulate specific content features, which can be cumbersome in scenarios requiring complex edits.
In response to the limitations of prompt-based systems, Instruction-based Image Editing (IIE) advances the paradigm by accepting simpler directives alongside the original image. The approach reduces the cognitive load on the user, requiring only straightforward modification instructions (e.g., “Add a sunset in the background”).
Prominent implementations such as InstructPix2Pix~\cite{brooks2023instructpix2pix} have further refined this approach. This model introduces large-scale image-text datasets generated using advanced language models like GPT-3~\cite{brown2020language} and employs diffusion processes for enhanced image manipulation. The automatic creation and filtering of datasets, however, raise concerns about the fidelity and noise levels in the resultant data.
To mitigate such issues, alternative strategies such as MagicBrush~\cite{zhang2024magicbrush} focus on manually curated datasets that align editing tasks more closely with human-like instructions. This method aims to balance the precision of machine-generated instructions with the contextual relevance typically observed in human annotations.
Further extending the dimensionality of IIE, InstructAny2Pix~\cite{li2023instructany2pix} accommodates multi-modal instructions, integrating not only textual directives but also inputs from other domains such as audio or visual cues. This multi-modality seeks to broaden the applicability and flexibility of IIE systems across various user preferences and environments.
To enhance specificity and fidelity in instruction-guided editing, MGIE~\cite{fu2023guiding} employs Multimodal Large Language Models (MLLMs). These models assist in providing detailed, context-aware interpretations of user instructions, thereby improving the precision of subsequent image modifications.
SmartEdit~\cite{achiam2023gpt}, designed for complex scene understanding, integrates MLLMs to bolster the interpretative capabilities of IIE models, fostering a better grasp of intricate editing requests. This advancement is accompanied by a specialized dataset tailored for complex image environments, supporting the model’s ability to handle sophisticated editing tasks with higher reliability.
\textcolor{black}{LEDITS++~\cite{brack2024ledits++} is shown to be an intuitive and lightweight framework that provides strong semantic control for image editing, which introduces a perfect inversion technique for more efficient diffusion-based editing, and demonstrate the method’s high efficiency, versatility, and editing precision through both qualitative and quantitative evaluations
}
Despite these strides forward, the evaluation of IIE model performance remains a critical challenge. The complexity and variability of human perception make it difficult to fully automate the assessment of editing quality. As such, systematic evaluation frameworks like \benchname are instrumental in providing a structured analysis of these models. Our contribution includes an extensive evaluation across diverse dimensions and multiple rounds of editing tasks, offering crucial insights into the strengths and limitations of current IIE methodologies and informing future research directions.

\subsection{Instruction-based Image Editing Benchmarks}

The area of instruction-based image editing is at an unique intersection of computer vision and natural language processing, demanding rigorous evaluation metrics to accurately assess model performance. Existing benchmarks~\cite{marino2019ok,hudson2019gqa,bigham2010vizwiz,lu2022learn,li2023evaluating,li2023seed,li2023mvbench,yu2023mm,wu2023q} for vision-language interactions have laid foundational groundwork, yet the specific demands of text-based image editing require tailored evaluation strategies.
Traditionally, the evaluation of text-to-image editing models has relied on widely accepted metrics such as CLIP Score~\cite{radford2021learning}, which assesses the similarity between images and textual descriptions, and metrics like PSNR~\cite{korhonen2012peak} and SSIM~\cite{wang2004image}, which evaluate image fidelity. LPIPS~\cite{zhang2018unreasonable} is also frequently used to measure perceptual similarity. While these metrics provide an initial framework for evaluation, they often fall short in capturing the nuance and intent expressed in complex editing instructions.
Several benchmarks have emerged seeking to address these challenges in specific contexts. TedBench~\cite{kawar2023imagic}, for instance, provides a dataset that, although limited to 100 images, captures a range of common editing actions. However, its scope is constrained, potentially limiting its utility for comprehensive model evaluation.
EditBench~\cite{wang2023imagen} narrows its focus to mask-guided image editing, a technique requiring additional masks that explicitly define editable regions. This specialization provides depth in evaluating mask applications but limits its generalizability across broader editing tasks.
EditVal~\cite{basu2023editval} introduces manual labor into its evaluation processes for certain features, enhancing accuracy at the cost of scalability and reproducibility. Its concentration on modifications like object sizing or positioning further narrows its assessment scope, missing broader interpretative editing capabilities.
Proposals like MagicBrush~\cite{zhang2024magicbrush} and Emu Edit~\cite{sheynin2023emu} augment traditional test sets with standard metrics such as L1, L2, CLIP-I, DINO, and CLIP-T. While providing a more granular assessment, these models still rely heavily on conventional metrics that may overlook the rich subtleties of varied editing tasks such as context-awareness or stylistic consistency.
SmartEdit~\cite{huang2023smartedit} contributes a benchmark specifically devised for complex scenes, yet its focus is not broad enough to cover the spectrum of simpler and more diverse editing scenarios, thus limiting its applicability.
\textcolor{black}{{TedBench++~\cite{brack2024ledits++} further extends the classical TedBench protocol with more comprehensive coverage and evaluation methodology, offering valuable insights into task diversity, compositionality, and evaluation challenges in instruction-driven image editing. TedBench++ also provides analysis on robustness, compositionality, and generalization, and discusses relevant failure cases, thus enriching the field's understanding of benchmarking needs. However, similar to earlier works, it is still limited in the coverage of multi-round and open-ended instruction-based editing tasks, highlighting the necessity of deeper and broader evaluation frameworks.}}
There is a pressing need for a systematic and comprehensive benchmark to evaluate IIE models, particularly one that covers a wide range of editing tasks—from basic to complex modifications. Current benchmarks often overlook the necessity of multi-round image editing, highlighting the importance of developing \benchname. This new benchmark aims to address this evaluative gap by offering a versatile framework that evaluates model performance across various tasks and supports multi-round editing. It aligns closely with human perception and interactive paradigms in image editing.
This advancement marks a significant step toward enhancing the standards for evaluating text-guided image editing. It provides detailed insights into model performance, informs future research directions, and ultimately improves user interaction models in the field.

\section{\benchname}
This part offers a summary of the primary elements of \benchname.
In Section~\ref{sec:eval}, we give a brief overview of the concepts, terminologies, and assessment techniques for high-level, low-level, and multi-round editing.
Section~\ref{sec:annotation} describes the data annotation procedure.
Finally, Section~\ref{sec:human} details the human assessment process used to examine the consistency between the \benchname score and human evaluation scores.

\begin{figure*}
  \centering
  \includegraphics[width=2.0\columnwidth]{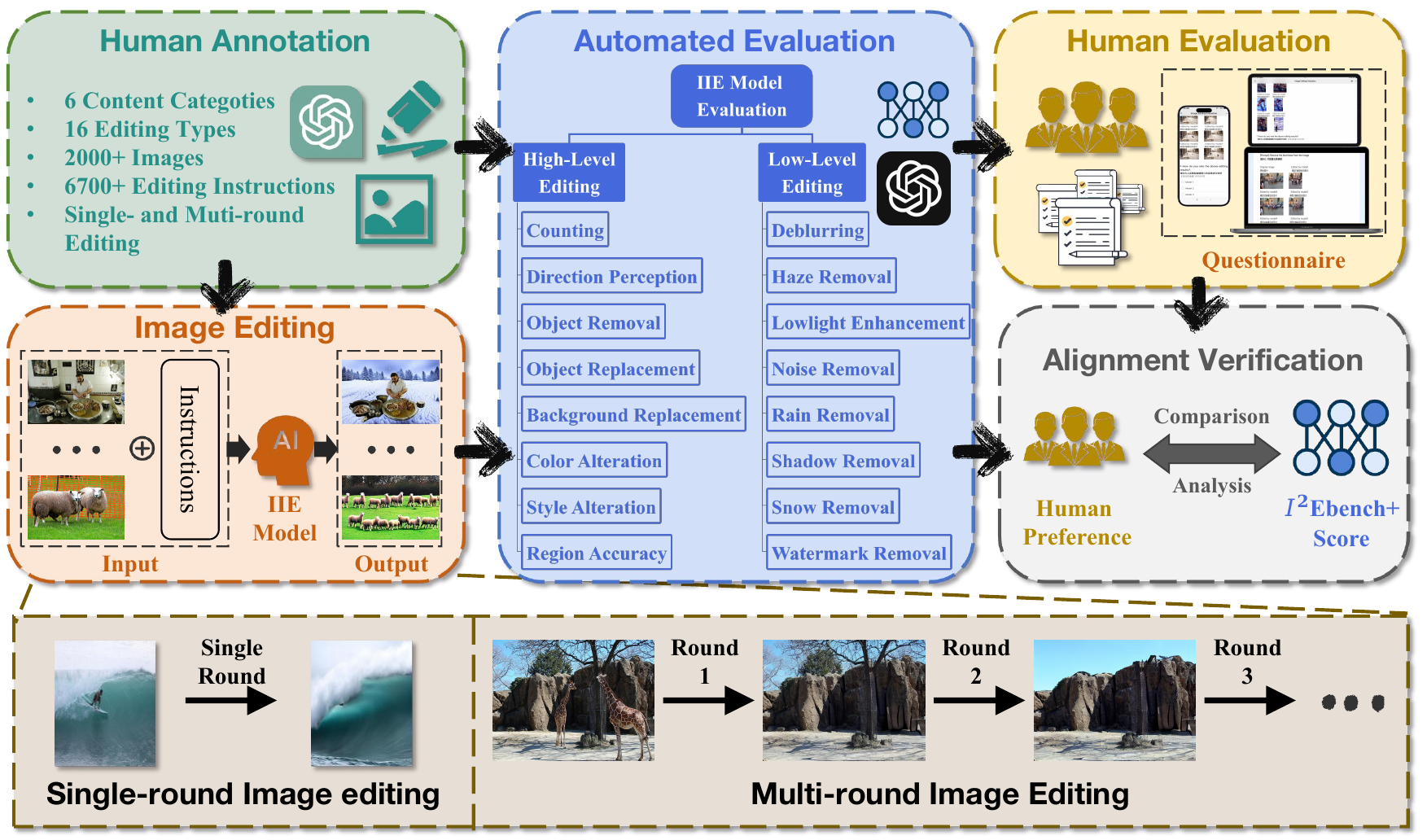}
  % \vspace{-1em}
  \caption{
    Overview of \benchname, an advanced automated system designed to evaluate the quality of edits produced by instruction-based image editing (IIE) models.
    We compiled a dataset comprising over 2000+ images sourced from various public datasets~\cite{lin2014microsoft,guo2023sky,MartinFTM01,chen2021all,ancuti2019dense,liu2021synthetic,Liu_2021_WACV,qu2017deshadownet,Nah_2017_CVPR,shen2019human,wei2018deep}, each annotated with its respective original editing instructions for both single-round and multi-round instruction-based image editing. To enhance instructional diversity, we leveraged ChatGPT~\cite{achiam2023gpt} to generate varied instruction versions.
    Utilizing these images along with both original and diverse instructions for single-round and multi-round editing, existing IIE models were employed to produce edited images.
    We then developed a comprehensive evaluation methodology to automatically measure the degree to which these edited images comply with the specified instructions across multiple dimensions.
    Additionally, we conducted human evaluations to gather subjective preferences regarding the editing results produced by different IIE models.
    Our analysis revealed a positive correlation between automated and human evaluations.
  }
  \label{fig:overview}
    % \vspace{-1em}
\end{figure*}

\subsection{Evaluation Dimension}
\label{sec:eval}

In our assessment of the IIE model's editing performance, we have identified 16 separate dimensions, each evaluating a different aspect of the editing process from a top-down perspective.
Moreover, beyond single-round editing, we also incorporate evaluations of multi-round editing for certain high-level dimensions.
Fig.~\ref{fig:overview} provides an illustration of \benchname.
High-level Editing Evaluation primarily evaluates the model's capability to accurately interpret instructions and execute precise modifications on specific areas of the input image, encompassing 8 distinct dimensions.
Low-level Editing Evaluation, conversely, focuses on global adjustments and intricate image processing. It also includes 8 evaluation dimensions.
Multi-round Editing Evaluation is designed to assess stability and robustness across multiple rounds of instruction editing.
Contrary to previous methodologies~\cite{fu2023guiding, zhang2023hive, geng2023instructdiffusion} that employed a single metric, such as the CLIP score~\cite{radford2021learning}, for evaluating all types of edits, we have devised tailored evaluation methods unique to each of the 16 dimensions.
This methodology is essential to account for the differing objectives of high-level and low-level edits.
For multi-round editing, we thoroughly consider the outcomes of each evaluation round to derive the overall multi-round score.

\subsubsection{High-Level Editing}
Assessing the quality of edits in high-level dimensions is challenging due to the varied objectives, making a single metric insufficient.
The progression of Multimodal Large Language Models (MLLM)~\cite{gao2024sphinx,chu2024mobilevlm,zhu2024vislinginstruct,dong2024internlm,ma2022x,ma2023towards,ji2022knowing}, such as GPT-4V~\cite{achiam2023gpt}, Gemini Pro~\cite{reid2024gemini}, and QWen-VL-Plus~\cite{bai2023qwen}, has greatly advanced automated image comprehension.
Thus, to accurately evaluate the editing quality of IIE models in high-level dimensions, we utilize the exceptional capabilities of the widely recognized GPT-4V model to assess most high-level evaluation dimensions.

\noindent{\textbf{Counting.}}
The Counting dimension involves instructions related to the quantity of objects, such as ``add two apples to the image."
To evaluate this aspect, we use GPT-4V to determine the number of target objects in the picture and compare its answers to human-annotated data.

\noindent{\textbf{Direction Perception.}}
Direction Perception requires the IIE model to interpret directional instructions and accurately modify images based on those directions.
This is assessed by querying GPT-4V about whether the target object is positioned correctly.

\noindent{\textbf{Object Removal.}}
Object removal focuses on removing the specified object in accordance with the provided instruction.
This aspect is evaluated by checking whether GPT-4V identifies the target object's presence in the image.

\noindent{\textbf{Object Replacement.}}
Object replacement pertains to substituting the original object with a new target object as directed.
We assess this by asking GPT-4V if the target object is present in the image.

\noindent{\textbf{Background Replacement.}}
Background replacement involves changing the original background to a specified one according to the instructions.
We evaluate this by verifying with GPT-4V whether the image background corresponds to the textual instruction.

\noindent{\textbf{Color Alteration.}}
Color alteration involves using instructions to change the color of the target object.
This dimension is evaluated by asking GPT-4V to identify the color of the target object in the edited image.

\noindent{\textbf{Style Alteration.}}
Style alteration focuses on modifying the style of the image.
This is assessed by measuring the CLIP similarity~\cite{radford2021learning} between the edited image and the text ``an image with $\cdots$ style".

\noindent{\textbf{Region Accuracy.}}
In editing tasks, region accuracy assesses not just if the target area was correctly edited, but also if unintended areas were altered.
To evaluate this, we sample input images and instructions from object removal, object replacement, and color alteration dimensions.
The area requiring editing is annotated with a mask.
We then fill the mask area in pre- and post-editing images with white and calculate SSIM~\cite{wang2004image} to measure this dimension.
A high SSIM value means that the area that is not expected to remain unchanged has not been modified.

\begin{figure*}
  \centering
  \includegraphics[width=1.8\columnwidth]{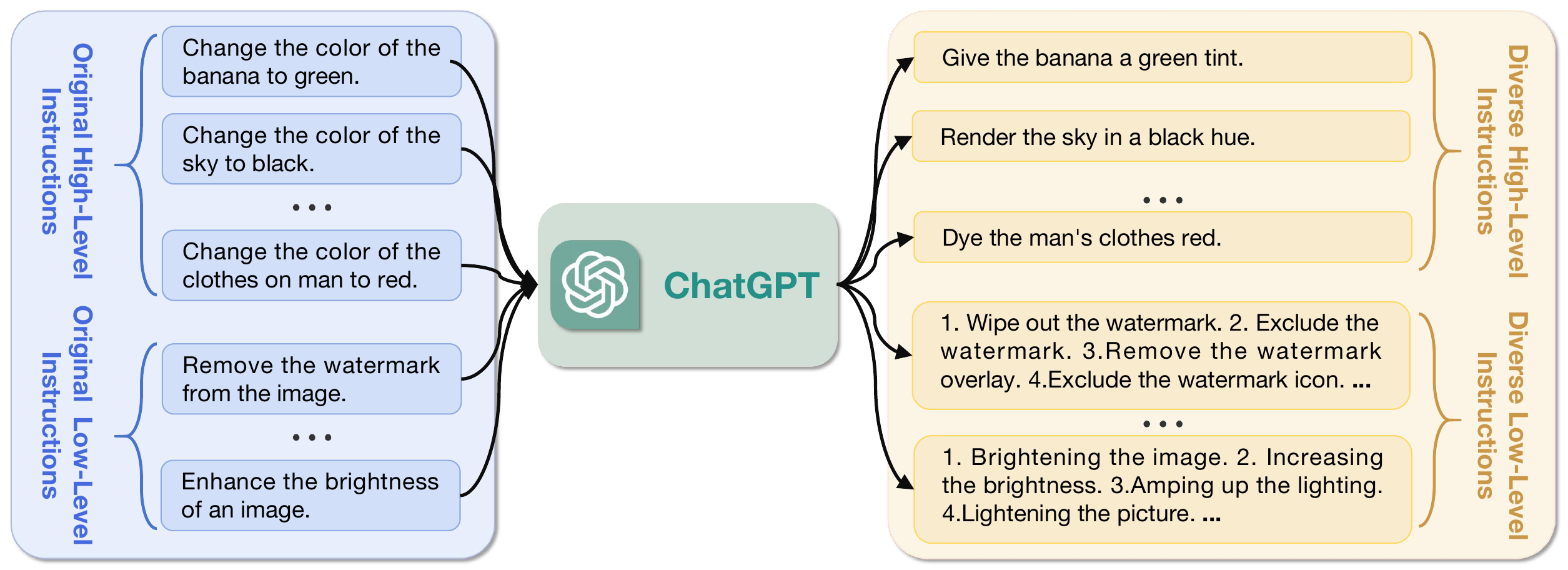}
  \caption{
    Illustration of converting original instructions to diverse instructions using ChatGPT.
  }
  \label{fig:diverse}
\end{figure*}

\begin{figure}
  \centering
  \includegraphics[width=1.0\columnwidth]{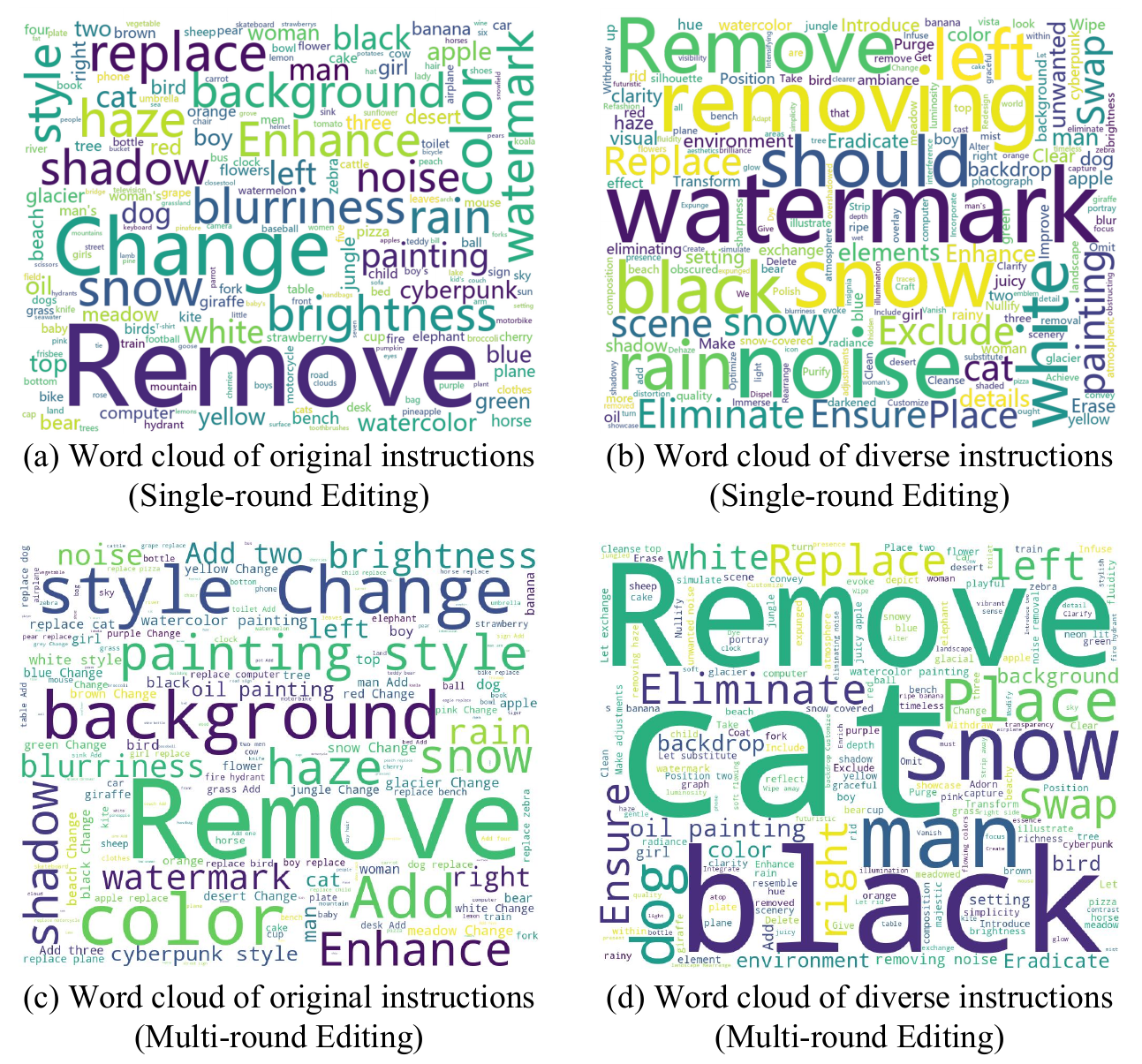}
  % \vspace{-1.5em}
  \caption{
    Word cloud visualization for single-round editing (a,b) and multi-round editing (c,d) of \benchname.
  }
  \label{fig:data}
  % \vspace{-2em}
\end{figure}

\subsubsection{Low-level Editing}
In contrast to high-level editing, instructions for low-level editing are more straightforward and do not specify object attributes like size, orientation, or color.
Over the years, numerous low-level editing tasks~\cite{wang2023omni,chen2023masked,sanghvi2023structured,chen2023learning,wu2023ridcp,guo2023shadowdiffusion,kong2022reflash} have been extensively developed, resulting in a relatively robust evaluation framework. As such, for low-level editing, we utilize the widely accepted metric, SSIM~\cite{wang2004image}, to assess editing quality.

\noindent{\textbf{Deblurring.}}
Deblurring involves the process of reducing or fully removing blurriness from images, thereby enhancing their sharpness and clarity.

\noindent{\textbf{Haze Removal.}}
Haze removal is the procedure of extracting or minimizing atmospheric haze or fog within images, which improves visibility and restores the scene's true colors and fine details.

\noindent{\textbf{Lowlight Enhancement.}}
Lowlight enhancement is the process of refining the quality of photos taken in dim conditions by increasing brightness and decreasing noise.

\noindent{\textbf{Noise Removal.}}
Noise removal is the practice of removing or reducing unwanted artifacts in images, resulting in cleaner and more aesthetically pleasing visuals.

\noindent{\textbf{Rain Removal.}}
Rain removal targets the diminishing or removal of visual disruptions caused by raindrops or rain streaks in photos, thus enhancing clarity and reinstating the image's original look.

\noindent{\textbf{Shadow Removal.}}
Shadow removal involves the reduction or elimination of undesirable shadows from images to boost visibility and improve overall visual quality.

\noindent{\textbf{Snow Removal.}}
Snow Removal concentrates on efficiently reducing or clearing snow from photographs.

\noindent{\textbf{Watermark Removal.}}
Watermark removal focuses on extracting or eradicating embedded watermarks from images, returning the image to its original state without the watermark.

\subsubsection{Multi-round Editing}
Multi-round editing is a widely used technique in practical applications, allowing users to iteratively refine images to achieve desired outcomes. Despite its importance, there is a significant lack of benchmarks specifically designed for multi-round editing, hindering progress in this area. To address this gap and promote development, we have enhanced \benchname to include evaluation capabilities for multi-round editing.

Considering the practical significance and popular usage of conducting multi-round editing on high-level dimensions, our evaluation primarily targets high-level dimensions. For example, repeatedly executing noise removal on an image does not sufficiently demonstrate the model's robustness or other capabilities.

To evaluate multi-round editing, our methodology uses supplementary instruction annotations for high-level dimensions, enabling image editing across 2-5 iterative rounds. Our focus is on dimensions beyond region accuracy. This is because, in multiple rounds of editing for the region accuracy dimension, edited images must first be generated by the IIE model before the editing region for the subsequent round can be annotated. This approach is unfavorable for automated evaluation, as different IIE models may produce significantly different edited images, complicating the annotation process. Additionally, when assessing new IIE models, the need to re-annotate the editing regions poses further challenges to automation.

\subsection{Human Annotation}
\label{sec:annotation}

\subsubsection{Data Annotation}
We carefully select around 140 images from publicly available datasets~\cite{lin2014microsoft, guo2023sky, MartinFTM01, chen2021all, ancuti2019dense, liu2021synthetic, Liu_2021_WACV, qu2017deshadownet, Nah_2017_CVPR, shen2019human, wei2018deep} for each evaluation dimension of \benchname. These images are then thoroughly annotated with textual editing instructions by human contributors, referred to as original instructions. However, instructions from human annotators typically follow a similar sentence structure. For example, in the object removal dimension, the common sentence pattern is ``remove $\cdots$ from the image." To enhance diversity, we employ ChatGPT~\cite{achiam2023gpt} to effectively rewrite the original instructions, as shown in Fig.~\ref{fig:diverse}.

Fig.~\ref{fig:data} (a) and (b) present the word cloud visualizations of the original and diverse instructions for single-round editing, respectively.
For the evaluation of multi-round editing, we further annotate high-level dimensions, except for the region accuracy dimension, with 2-5 rounds of textual editing instructions.
Fig.~\ref{fig:data} (c) and (d) show the word cloud visualizations of the original and diverse instructions for multi-round editing, respectively.
Moreover, each instruction is further categorized into types such as animal, object, scenery, plant, human, and global.

\begin{figure*}
  \centering
  \includegraphics[width=2.0\columnwidth]{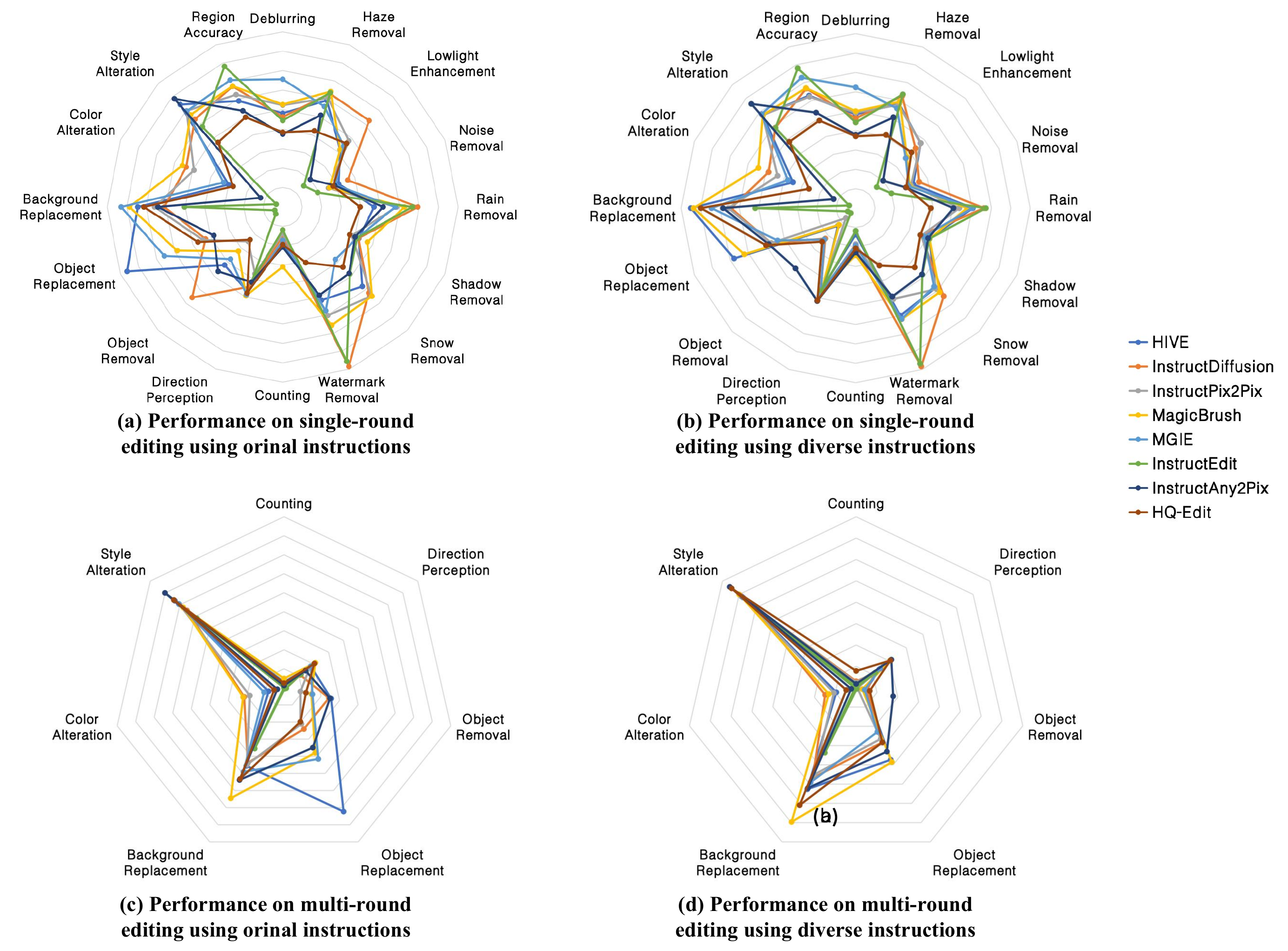}
  \caption{
    Comparison of radar charts for \benchname scores in different dimensions using (a,c) original instructions and (b,d) diverse instructions for both single- and multi-round editing.
  }
  \label{fig:score_radar}
\end{figure*}

\subsubsection{Evaluation Annotation}
The evaluation of \benchname involves two distinct categories.
The first category employs conventional metrics to assess various dimensions. For the style alteration dimension, we use the CLIP score as a standard measure, which does not require any additional evaluation annotations. For the region accuracy and low-level editing dimensions, we utilize the SSIM metric.
In the second category, we leverage GPT-4V to evaluate the editing quality. To support this evaluation, we engaged human annotators to create questions specifically for GPT-4V, accompanied by appropriate standard answers.
For example, in the counting dimension with the instruction ``Add a cat to the shoe rack," the human annotators provided the question ``How many cats are there on the shoe rack?" with the corresponding answer ``One."

For multi-round editing of an input image, each round must be successful for the sample to be considered successfully edited and to receive a score. If any round of editing fails, the sample is deemed to have failed in multi-round editing.

\subsection{Human Evaluation}
\label{sec:human}

The primary aim of the human evaluation is to determine the relationship between human perceptions and the \benchname\ scores.
To accomplish this, human evaluators are provided with a textual instruction \(T\), an initial image \(V_I\), and a series of edited images \(\{V_1, V_2, \ldots, V_M\}\) produced by \(M\) different IIE models. Evaluators are then responsible for ranking these results based on their own judgment.
Specifically, we select \(N\) samples for each evaluation dimension, resulting in a comprehensive set of \(N \times 16 \times 2\) edited image comparisons. In each comparison, evaluators examine \(M\) edited images and rank them relative to one another.
Each model is then assigned a human score according to its ranking among the \(M\) models.
In detail, the model ranked first among the \(M\) models receives a human score of \(M\), while the model ranked last receives a human score of 1. A model ranked \(k^{\text{th}}\) is assigned a human score of \(M - k + 1\).
The human score for each dimension is calculated by averaging the human scores across all samples within that dimension.
Therefore, the human score for each model ranges from 1 to \(M\).

\begin{table*}[!t]
\caption{
\textcolor{black}{
\benchname\ evaluation results per dimension using original instructions. 
Abbreviations: Deb. = Deblurring; Haze Rem. = Haze Removal; Lowl. Enh. = Lowlight Enhancement; Noise Rem. = Noise Removal; Rain Rem. = Rain Removal; Shadow Rem. = Shadow Removal; Snow Rem. = Snow Removal; Wtrmk Rem. = Watermark Removal; Cnt. = Counting; Dir. Perc. = Direction Perception; Obj. Rem. = Object Removal; Obj. Repl. = Object Replacement; Bkg. Repl. = Background Replacement; Col. Alt. = Color Alteration; Sty. Alt. = Style Alteration; Reg. Acc. = Region Accuracy.
}
}
\setlength{\tabcolsep}{3pt}
\renewcommand{\arraystretch}{1.0}
\centering
\resizebox{\textwidth}{!}{%
\begin{tabular}{l|c|c|c|c|c|c|c|c}
\midrule
\multicolumn{9}{c}{\textcolor{black}{\textbf{Low-level Editing}}} \\ 
\midrule
\textcolor{black}{\textbf{Model}} & \textcolor{black}{\textbf{Deb.}} & \textcolor{black}{\textbf{Haze Rem.}} & \textcolor{black}{\textbf{Lowl. Enh.}} & \textcolor{black}{\textbf{Noise Rem.}} & \textcolor{black}{\textbf{Rain Rem.}} & \textcolor{black}{\textbf{Shadow Rem.}} & \textcolor{black}{\textbf{Snow Rem.}} & \textcolor{black}{\textbf{Wtrmk Rem.}} \\
\midrule
\textcolor{black}{\textbf{HIVE~\cite{zhang2023hive}}} & \textcolor{black}{44.25} & \textcolor{black}{54.89} & \textcolor{black}{37.61} & \textcolor{black}{24.59} & \textcolor{black}{45.47} & \textcolor{black}{37.61} & \textcolor{black}{51.49} & \textcolor{black}{49.99} \\
\textcolor{black}{\textbf{InstructDiffusion~\cite{geng2023instructdiffusion}}} & \textcolor{black}{42.48} & \textcolor{black}{58.45} & \textcolor{black}{56.61} & \textcolor{black}{28.60} & \textcolor{black}{67.20} & \textcolor{black}{37.43} & \textcolor{black}{55.65} & \textcolor{black}{85.49} \\
\textcolor{black}{\textbf{InstructPix2Pix~\cite{brooks2023instructpix2pix}}} & \textcolor{black}{48.03} & \textcolor{black}{56.15} & \textcolor{black}{43.32} & \textcolor{black}{20.11} & \textcolor{black}{56.64} & \textcolor{black}{34.19} & \textcolor{black}{57.59} & \textcolor{black}{58.12} \\
\textcolor{black}{\textbf{MagicBrush~\cite{zhang2024magicbrush}}} & \textcolor{black}{48.38} & \textcolor{black}{59.46} & \textcolor{black}{37.71} & \textcolor{black}{20.59} & \textcolor{black}{60.60} & \textcolor{black}{\textbf{41.91}} & \textcolor{black}{\textbf{57.81}} & \textcolor{black}{63.33} \\
\textcolor{black}{\textbf{MGIE~\cite{fu2023guiding}}} & \textcolor{black}{\textbf{60.30}} & \textcolor{black}{51.75} & \textcolor{black}{39.99} & \textcolor{black}{23.25} & \textcolor{black}{56.00} & \textcolor{black}{36.91} & \textcolor{black}{34.04} & \textcolor{black}{55.53} \\
\textcolor{black}{\textbf{InstructEdit~\cite{wang2023instructedit}}} & \textcolor{black}{40.77} & \textcolor{black}{58.85} & \textcolor{black}{13.83} & \textcolor{black}{15.40} & \textcolor{black}{64.44} & \textcolor{black}{36.88} & \textcolor{black}{43.45} & \textcolor{black}{82.68} \\
\textcolor{black}{\textbf{InstructAny2Pix~\cite{li2023instructany2pix}}} & \textcolor{black}{34.34} & \textcolor{black}{47.27} & \textcolor{black}{18.03} & \textcolor{black}{22.89} & \textcolor{black}{49.94} & \textcolor{black}{35.84} & \textcolor{black}{42.97} & \textcolor{black}{47.28} \\
\textcolor{black}{\textbf{HQ-Edit~\cite{hui2024hq}}} & \textcolor{black}{35.27} & \textcolor{black}{39.25} & \textcolor{black}{41.71} & \textcolor{black}{22.13} & \textcolor{black}{38.52} & \textcolor{black}{33.13} & \textcolor{black}{38.97} & \textcolor{black}{29.80} \\
\textcolor{black}{\textbf{FLUX Kontext~\cite{batifol2025flux}}} & \textcolor{black}{33.35} & \textcolor{black}{\textbf{66.30}} & \textcolor{black}{34.42} & \textcolor{black}{28.23} & \textcolor{black}{\textbf{79.71}} & \textcolor{black}{31.69} & \textcolor{black}{45.17} & \textcolor{black}{\textbf{92.24}} \\
\textcolor{black}{\textbf{Qwen-image~\cite{wu2025qwen}}} & \textcolor{black}{43.35} & \textcolor{black}{47.19} & \textcolor{black}{\textbf{73.73}} & \textcolor{black}{\textbf{52.75}} & \textcolor{black}{53.38} & \textcolor{black}{38.77} & \textcolor{black}{57.68} & \textcolor{black}{47.10} \\
\hline
\rowcolor[HTML]{F2F2F2}
\textcolor{black}{\textbf{Exp Min}} & \textcolor{black}{13.79} & \textcolor{black}{12.66} & \textcolor{black}{0.09} & \textcolor{black}{0.79} & \textcolor{black}{7.38} & \textcolor{black}{1.05} & \textcolor{black}{2.18} & \textcolor{black}{1.34} \\
\rowcolor[HTML]{F2F2F2}
\textcolor{black}{\textbf{Exp Max}} & \textcolor{black}{91.94} & \textcolor{black}{92.70} & \textcolor{black}{89.60} & \textcolor{black}{77.00} & \textcolor{black}{96.11} & \textcolor{black}{89.19} & \textcolor{black}{89.26} & \textcolor{black}{96.42} \\
\midrule

\multicolumn{9}{c}{\textcolor{black}{\textbf{High-level Editing}}} \\
\midrule
\textcolor{black}{\textbf{Model}} & \textcolor{black}{\textbf{Cnt.}} & \textcolor{black}{\textbf{Dir. Perc.}} & \textcolor{black}{\textbf{Obj. Rem.}} & \textcolor{black}{\textbf{Obj. Repl.}} & \textcolor{black}{\textbf{Bkg. Repl.}} & \textcolor{black}{\textbf{Col. Alt.}} & \textcolor{black}{\textbf{Sty. Alt.}} & \textcolor{black}{\textbf{Reg. Acc.}} \\
\midrule
\textcolor{black}{\textbf{HIVE~\cite{zhang2023hive}}} & \textcolor{black}{18.57} & \textcolor{black}{47.14} & \textcolor{black}{42.14} & \textcolor{black}{86.43} & \textcolor{black}{74.29} & \textcolor{black}{30.00} & \textcolor{black}{25.32} & \textcolor{black}{58.15} \\
\textcolor{black}{\textbf{InstructDiffusion~\cite{geng2023instructdiffusion}}} & \textcolor{black}{15.00} & \textcolor{black}{44.29} & \textcolor{black}{65.71} & \textcolor{black}{42.86} & \textcolor{black}{60.71} & \textcolor{black}{53.57} & \textcolor{black}{21.69} & \textcolor{black}{66.18} \\
\textcolor{black}{\textbf{InstructPix2Pix~\cite{brooks2023instructpix2pix}}} & \textcolor{black}{13.57} & \textcolor{black}{37.14} & \textcolor{black}{25.00} & \textcolor{black}{44.29} & \textcolor{black}{65.71} & \textcolor{black}{49.29} & \textcolor{black}{23.76} & \textcolor{black}{61.63} \\
\textcolor{black}{\textbf{MagicBrush~\cite{zhang2024magicbrush}}} & \textcolor{black}{30.71} & \textcolor{black}{49.29} & \textcolor{black}{32.14} & \textcolor{black}{58.57} & \textcolor{black}{78.57} & \textcolor{black}{55.71} & \textcolor{black}{22.78} & \textcolor{black}{66.34} \\
\textcolor{black}{\textbf{MGIE~\cite{fu2023guiding}}} & \textcolor{black}{17.14} & \textcolor{black}{48.57} & \textcolor{black}{37.86} & \textcolor{black}{65.71} & \textcolor{black}{82.86} & \textcolor{black}{32.86} & \textcolor{black}{23.68} & \textcolor{black}{69.60} \\
\textcolor{black}{\textbf{InstructEdit~\cite{wang2023instructedit}}} & \textcolor{black}{11.76} & \textcolor{black}{41.73} & \textcolor{black}{5.04} & \textcolor{black}{4.41} & \textcolor{black}{50.36} & \textcolor{black}{3.62} & \textcolor{black}{19.83} & \textcolor{black}{\textbf{77.08}} \\
\textcolor{black}{\textbf{InstructAny2Pix~\cite{li2023instructany2pix}}} & \textcolor{black}{20.59} & \textcolor{black}{41.73} & \textcolor{black}{46.76} & \textcolor{black}{38.24} & \textcolor{black}{64.03} & \textcolor{black}{12.32} & \textcolor{black}{26.76} & \textcolor{black}{52.75} \\
\textcolor{black}{\textbf{HQ-Edit~\cite{hui2024hq}}} & \textcolor{black}{19.26} & \textcolor{black}{47.79} & \textcolor{black}{23.74} & \textcolor{black}{47.06} & \textcolor{black}{71.22} & \textcolor{black}{27.54} & \textcolor{black}{15.96} & \textcolor{black}{49.21} \\
\textcolor{black}{\textbf{FLUX Kontext~\cite{batifol2025flux}}} & \textcolor{black}{49.63} & \textcolor{black}{78.68} & \textcolor{black}{64.75} & \textcolor{black}{62.50} & \textcolor{black}{85.61} & \textcolor{black}{78.26} & \textcolor{black}{26.33} & \textcolor{black}{53.14} \\
\textcolor{black}{\textbf{Qwen-image~\cite{wu2025qwen}}} & \textcolor{black}{\textbf{65.19}} & \textcolor{black}{\textbf{83.09}} & \textcolor{black}{\textbf{78.42}} & \textcolor{black}{\textbf{88.24}} & \textcolor{black}{\textbf{96.40}} & \textcolor{black}{\textbf{83.33}} & \textcolor{black}{\textbf{27.18}} & \textcolor{black}{71.07} \\
\hline
\rowcolor[HTML]{F2F2F2}
\textcolor{black}{\textbf{Exp Min}} & \textcolor{black}{0.00} & \textcolor{black}{0.00} & \textcolor{black}{0.00} & \textcolor{black}{0.00} & \textcolor{black}{0.00} & \textcolor{black}{0.00} & \textcolor{black}{12.55} & \textcolor{black}{6.41} \\
\rowcolor[HTML]{F2F2F2}
\textcolor{black}{\textbf{Exp Max}} & \textcolor{black}{100.00} & \textcolor{black}{100.00} & \textcolor{black}{100.00} & \textcolor{black}{100.00} & \textcolor{black}{100.00} & \textcolor{black}{100.00} & \textcolor{black}{33.91} & \textcolor{black}{98.70} \\
\midrule

\multicolumn{9}{c}{\textcolor{black}{\textbf{Multi-round Editing}}} \\
\midrule
\textcolor{black}{\textbf{Model}} & \textcolor{black}{\textbf{Cnt.}} & \textcolor{black}{\textbf{Dir. Perc.}} & \textcolor{black}{\textbf{Obj. Rem.}} & \textcolor{black}{\textbf{Obj. Repl.}} & \textcolor{black}{\textbf{Bkg. Repl.}} & \textcolor{black}{\textbf{Col. Alt.}} & \textcolor{black}{\textbf{Sty. Alt.}} & \textcolor{black}{\textbf{Reg. Acc.}} \\
\midrule
\textcolor{black}{\textbf{HIVE~\cite{zhang2023hive}}} & \textcolor{black}{3.03} & \textcolor{black}{18.07} & \textcolor{black}{25.44} & \textcolor{black}{\textbf{72.29}} & \textcolor{black}{45.56} & \textcolor{black}{8.33} & \textcolor{black}{24.84} & \textcolor{black}{-} \\
\textcolor{black}{\textbf{InstructDiffusion~\cite{geng2023instructdiffusion}}} & \textcolor{black}{1.21} & \textcolor{black}{10.84} & \textcolor{black}{24.26} & \textcolor{black}{24.10} & \textcolor{black}{43.79} & \textcolor{black}{21.43} & \textcolor{black}{22.09} & \textcolor{black}{-} \\
\textcolor{black}{\textbf{InstructPix2Pix~\cite{brooks2023instructpix2pix}}} & \textcolor{black}{1.82} & \textcolor{black}{17.47} & \textcolor{black}{8.88} & \textcolor{black}{21.08} & \textcolor{black}{44.97} & \textcolor{black}{18.45} & \textcolor{black}{23.84} & \textcolor{black}{-} \\
\textcolor{black}{\textbf{MagicBrush~\cite{zhang2024magicbrush}}} & \textcolor{black}{4.85} & \textcolor{black}{21.08} & \textcolor{black}{14.79} & \textcolor{black}{37.95} & \textcolor{black}{64.50} & \textcolor{black}{22.02} & \textcolor{black}{22.96} & \textcolor{black}{-} \\
\textcolor{black}{\textbf{MGIE~\cite{fu2023guiding}}} & \textcolor{black}{1.21} & \textcolor{black}{18.07} & \textcolor{black}{15.38} & \textcolor{black}{41.57} & \textcolor{black}{49.11} & \textcolor{black}{10.71} & \textcolor{black}{23.95} & \textcolor{black}{-} \\
\textcolor{black}{\textbf{InstructEdit~\cite{wang2023instructedit}}} & \textcolor{black}{0.61} & \textcolor{black}{11.45} & \textcolor{black}{1.18} & \textcolor{black}{0.60} & \textcolor{black}{35.50} & \textcolor{black}{0.00} & \textcolor{black}{19.88} & \textcolor{black}{-} \\
\textcolor{black}{\textbf{InstructAny2Pix~\cite{li2023instructany2pix}}} & \textcolor{black}{1.21} & \textcolor{black}{14.46} & \textcolor{black}{24.85} & \textcolor{black}{34.94} & \textcolor{black}{53.85} & \textcolor{black}{3.57} & \textcolor{black}{27.09} & \textcolor{black}{-} \\
\textcolor{black}{\textbf{HQ-Edit~\cite{hui2024hq}}} & \textcolor{black}{2.42} & \textcolor{black}{20.48} & \textcolor{black}{11.83} & \textcolor{black}{19.88} & \textcolor{black}{53.25} & \textcolor{black}{5.36} & \textcolor{black}{25.07} & \textcolor{black}{-} \\
\textcolor{black}{\textbf{FLUX Kontext~\cite{batifol2025flux}}} & \textcolor{black}{2.42} & \textcolor{black}{45.18} & \textcolor{black}{36.69} & \textcolor{black}{36.75} & \textcolor{black}{71.01} & \textcolor{black}{54.17} & \textcolor{black}{26.84} & \textcolor{black}{-} \\
\textcolor{black}{\textbf{Qwen-image~\cite{wu2025qwen}}} & \textcolor{black}{\textbf{11.52}} & \textcolor{black}{\textbf{57.23}} & \textcolor{black}{\textbf{43.20}} & \textcolor{black}{63.86} & \textcolor{black}{\textbf{81.66}} & \textcolor{black}{\textbf{66.67}} & \textcolor{black}{\textbf{27.70}} & \textcolor{black}{-} \\
\hline
\rowcolor[HTML]{F2F2F2}
\textcolor{black}{\textbf{Exp Min}} & \textcolor{black}{0.00} & \textcolor{black}{0.00} & \textcolor{black}{0.00} & \textcolor{black}{0.00} & \textcolor{black}{0.00} & \textcolor{black}{0.00} & \textcolor{black}{13.12} & \textcolor{black}{-} \\
\rowcolor[HTML]{F2F2F2}
\textcolor{black}{\textbf{Exp Max}} & \textcolor{black}{100.00} & \textcolor{black}{100.00} & \textcolor{black}{100.00} & \textcolor{black}{100.00} & \textcolor{black}{100.00} & \textcolor{black}{100.00} & \textcolor{black}{33.84} & \textcolor{black}{-} \\
\midrule
\end{tabular}
}% end resizebox
\label{tab:original}
\end{table*}

% % ------------------------

% % ------------------------

\begin{table*}[!t]
\caption{
\textcolor{black}{
\benchname\ evaluation results per dimension using diverse instructions. {Abbreviations:} Debl. = Deblurring; Haze Rem. = Haze Removal; Lowl. Enh. = Lowlight Enhancement; Noise Rem. = Noise Removal; Rain Rem. = Rain Removal; Shadow Rem. = Shadow Removal; Snow Rem. = Snow Removal; Wtrmk Rem. = Watermark Removal; Cnt. = Counting; Dir. Perc. = Direction Perception; Obj. Rem. = Object Removal; Obj. Repl. = Object Replacement; Bkg. Repl. = Background Replacement; Col. Alt. = Color Alteration; Sty. Alt. = Style Alteration; Reg. Acc. = Region Accuracy.
}
}
\setlength{\tabcolsep}{4pt}
\renewcommand{\arraystretch}{0.9}
\centering
\resizebox{\textwidth}{!}{%
\begin{tabular}{l|c|c|c|c|c|c|c|c}
\midrule
\multicolumn{9}{c}{\textcolor{black}{\textbf{Low-level Editing}}} \\ 
\midrule
\textcolor{black}{\textbf{Model}}             
& \textcolor{black}{\textbf{Debl.}} 
& \textcolor{black}{\textbf{Haze Rem.}} 
& \textcolor{black}{\textbf{Lowl. Enh.}} 
& \textcolor{black}{\textbf{Noise Rem.}}     
& \textcolor{black}{\textbf{Rain Rem.}}     
& \textcolor{black}{\textbf{Shadow Rem.}}   
& \textcolor{black}{\textbf{Snow Rem.}}     
& \textcolor{black}{\textbf{Wtrmk Rem.}} \\
\midrule
\textcolor{black}{\textbf{HIVE~\cite{zhang2023hive}}}              & \textcolor{black}{44.41} & \textcolor{black}{54.09} & \textcolor{black}{42.78} & \textcolor{black}{25.51} & \textcolor{black}{58.59} & \textcolor{black}{36.69} & \textcolor{black}{51.92} & \textcolor{black}{57.88} \\
\textcolor{black}{\textbf{InstructDiffusion~\cite{geng2023instructdiffusion}}} & \textcolor{black}{42.62} & \textcolor{black}{58.01} & \textcolor{black}{39.47} & \textcolor{black}{28.06} & \textcolor{black}{64.18} & \textcolor{black}{32.54} & \textcolor{black}{\textbf{57.30}} & \textcolor{black}{85.14} \\
\textcolor{black}{\textbf{InstructPix2Pix~\cite{brooks2023instructpix2pix}}}   & \textcolor{black}{45.24} & \textcolor{black}{53.52} & \textcolor{black}{42.88} & \textcolor{black}{24.49} & \textcolor{black}{51.86} & \textcolor{black}{32.79} & \textcolor{black}{52.67} & \textcolor{black}{48.91} \\
\textcolor{black}{\textbf{MagicBrush~\cite{zhang2024magicbrush}}}        & \textcolor{black}{45.96} & \textcolor{black}{55.11} & \textcolor{black}{33.74} & \textcolor{black}{23.91} & \textcolor{black}{55.77} & \textcolor{black}{36.73} & \textcolor{black}{54.68} & \textcolor{black}{59.76} \\
\textcolor{black}{\textbf{MGIE~\cite{fu2023guiding}}}              & \textcolor{black}{57.33} & \textcolor{black}{51.61} & \textcolor{black}{32.96} & \textcolor{black}{23.49} & \textcolor{black}{58.27} & \textcolor{black}{34.07} & \textcolor{black}{51.02} & \textcolor{black}{59.64} \\
\textcolor{black}{\textbf{InstructEdit~\cite{wang2023instructedit}}}      & \textcolor{black}{40.66} & \textcolor{black}{58.89} & \textcolor{black}{13.92} & \textcolor{black}{15.81} & \textcolor{black}{65.08} & \textcolor{black}{36.66} & \textcolor{black}{43.34} & \textcolor{black}{83.68} \\
\textcolor{black}{\textbf{InstructAny2Pix~\cite{li2023instructany2pix}}}   & \textcolor{black}{34.77} & \textcolor{black}{47.00} & \textcolor{black}{18.09} & \textcolor{black}{22.18} & \textcolor{black}{48.92} & \textcolor{black}{36.04} & \textcolor{black}{43.13} & \textcolor{black}{47.58} \\
\textcolor{black}{\textbf{HQ-Edit~\cite{hui2024hq}}} & \textcolor{black}{34.11} & \textcolor{black}{37.95} & \textcolor{black}{36.76} & \textcolor{black}{22.38} & \textcolor{black}{37.60} & \textcolor{black}{32.17} & \textcolor{black}{38.45} & \textcolor{black}{30.83} \\
\textcolor{black}{\textbf{FLUX Kontext~\cite{batifol2025flux}}} & \textcolor{black}{32.11} & \textcolor{black}{\textbf{66.00}} & \textcolor{black}{27.14} & \textcolor{black}{27.11} & \textcolor{black}{\textbf{75.07}} & \textcolor{black}{32.04} & \textcolor{black}{43.67} & \textcolor{black}{\textbf{93.18}} \\
\textcolor{black}{\textbf{Qwen-image~\cite{wu2025qwen}}} & \textcolor{black}{\textbf{59.69}} & \textcolor{black}{49.12} & \textcolor{black}{\textbf{72.78}} & \textcolor{black}{\textbf{52.91}} & \textcolor{black}{54.70} & \textcolor{black}{\textbf{47.78}} & \textcolor{black}{{57.07}} & \textcolor{black}{65.10} \\
\hline
\rowcolor[HTML]{F2F2F2} 
\textcolor{black}{\textbf{Exp Min}}           & \textcolor{black}{6.32} & \textcolor{black}{3.67} & \textcolor{black}{0.60} & \textcolor{black}{0.03} & \textcolor{black}{7.22} & \textcolor{black}{1.46} & \textcolor{black}{3.78} & \textcolor{black}{2.58} \\
\rowcolor[HTML]{F2F2F2} 
\textcolor{black}{\textbf{Exp Max}}           & \textcolor{black}{91.10} & \textcolor{black}{96.59} & \textcolor{black}{90.66} & \textcolor{black}{92.61} & \textcolor{black}{98.31} & \textcolor{black}{93.23} & \textcolor{black}{90.83} & \textcolor{black}{98.60} \\
\midrule

\multicolumn{9}{c}{\textcolor{black}{\textbf{High-level Editing}}} \\
\midrule
\textcolor{black}{\textbf{Model}}             
& \textcolor{black}{\textbf{Cnt.}}   
& \textcolor{black}{\textbf{Dir. Perc.}} 
& \textcolor{black}{\textbf{Obj. Rem.}}       
& \textcolor{black}{\textbf{Obj. Repl.}} 
& \textcolor{black}{\textbf{Bkg. Repl.}} 
& \textcolor{black}{\textbf{Col. Alt.}} 
& \textcolor{black}{\textbf{Sty. Alt.}} 
& \textcolor{black}{\textbf{Reg. Acc.}}   \\
\midrule
\textcolor{black}{\textbf{HIVE~\cite{zhang2023hive}}}              & \textcolor{black}{13.57} & \textcolor{black}{43.57} & \textcolor{black}{12.86} & \textcolor{black}{67.86} & \textcolor{black}{85.00} & \textcolor{black}{35.00} & \textcolor{black}{23.08} & \textcolor{black}{61.97} \\
\textcolor{black}{\textbf{InstructDiffusion~\cite{geng2023instructdiffusion}}} & \textcolor{black}{21.43} & \textcolor{black}{47.86} & \textcolor{black}{22.14} & \textcolor{black}{47.14} & \textcolor{black}{64.29} & \textcolor{black}{48.57} & \textcolor{black}{19.96} & \textcolor{black}{65.92} \\
\textcolor{black}{\textbf{InstructPix2Pix~\cite{brooks2023instructpix2pix}}}   & \textcolor{black}{18.57} & \textcolor{black}{47.86} & \textcolor{black}{7.14} & \textcolor{black}{47.14} & \textcolor{black}{65.71} & \textcolor{black}{43.57} & \textcolor{black}{23.13} & \textcolor{black}{61.32} \\
\textcolor{black}{\textbf{MagicBrush~\cite{zhang2024magicbrush}}}        & \textcolor{black}{24.29} & \textcolor{black}{45.71} & \textcolor{black}{12.14} & \textcolor{black}{62.14} & \textcolor{black}{83.57} & \textcolor{black}{54.29} & \textcolor{black}{23.08} & \textcolor{black}{66.21} \\
\textcolor{black}{\textbf{MGIE~\cite{fu2023guiding}}}              & \textcolor{black}{19.29} & \textcolor{black}{47.14} & \textcolor{black}{22.86} & \textcolor{black}{43.57} & \textcolor{black}{74.29} & \textcolor{black}{37.86} & \textcolor{black}{23.36} & \textcolor{black}{71.89} \\
\textcolor{black}{\textbf{InstructEdit~\cite{wang2023instructedit}}}      & \textcolor{black}{11.76} & \textcolor{black}{46.04} & \textcolor{black}{3.60} & \textcolor{black}{4.41} & \textcolor{black}{51.80} & \textcolor{black}{3.62} & \textcolor{black}{19.91} & \textcolor{black}{\textbf{77.08}} \\
\textcolor{black}{\textbf{InstructAny2Pix~\cite{li2023instructany2pix}}}   & \textcolor{black}{22.79} & \textcolor{black}{51.80} & \textcolor{black}{43.88} & \textcolor{black}{48.53} & \textcolor{black}{68.35} & \textcolor{black}{12.32} & \textcolor{black}{25.93} & \textcolor{black}{52.61} \\
\textcolor{black}{\textbf{HQ-Edit~\cite{hui2024hq}}} & \textcolor{black}{20.74} & \textcolor{black}{51.47} & \textcolor{black}{24.46} & \textcolor{black}{50.00} & \textcolor{black}{79.86} & \textcolor{black}{26.09} & \textcolor{black}{16.48} & \textcolor{black}{48.29} \\
\textcolor{black}{\textbf{FLUX Kontext~\cite{batifol2025flux}}} & \textcolor{black}{45.19} & \textcolor{black}{76.47} & \textcolor{black}{50.36} & \textcolor{black}{83.82} & \textcolor{black}{84.89} & \textcolor{black}{73.91} & \textcolor{black}{26.27} & \textcolor{black}{52.43} \\
\textcolor{black}{\textbf{Qwen-image~\cite{wu2025qwen}}} & \textcolor{black}{\textbf{59.26}} & \textcolor{black}{\textbf{82.35}} & \textcolor{black}{\textbf{74.10}} & \textcolor{black}{\textbf{97.06}} & \textcolor{black}{\textbf{93.53}} & \textcolor{black}{\textbf{79.71}} & \textcolor{black}{\textbf{27.42}} & \textcolor{black}{{71.35}} \\
\hline
\rowcolor[HTML]{F2F2F2} 
\textcolor{black}{\textbf{Exp Min}}           & \textcolor{black}{0.00} & \textcolor{black}{0.00} & \textcolor{black}{0.00} & \textcolor{black}{0.00} & \textcolor{black}{0.00} & \textcolor{black}{0.00} & \textcolor{black}{10.62} & \textcolor{black}{9.79} \\
\rowcolor[HTML]{F2F2F2} 
\textcolor{black}{\textbf{Exp Max}}           & \textcolor{black}{100.00} & \textcolor{black}{100.00} & \textcolor{black}{100.00} & \textcolor{black}{100.00} & \textcolor{black}{100.00} & \textcolor{black}{100.00} & \textcolor{black}{34.06} & \textcolor{black}{98.68} \\
\midrule

\multicolumn{9}{c}{\textcolor{black}{\textbf{Multi-round Editing}}} \\ 
\midrule
\textcolor{black}{\textbf{Model}}             
& \textcolor{black}{\textbf{Cnt.}}   
& \textcolor{black}{\textbf{Dir. Perc.}} 
& \textcolor{black}{\textbf{Obj. Rem.}}       
& \textcolor{black}{\textbf{Obj. Repl.}} 
& \textcolor{black}{\textbf{Bkg. Repl.}} 
& \textcolor{black}{\textbf{Col. Alt.}} 
& \textcolor{black}{\textbf{Sty. Alt.}} 
& \textcolor{black}{\textbf{Reg. Acc.}}   \\
\midrule
\textcolor{black}{\textbf{HIVE~\cite{zhang2023hive}}}              & \textcolor{black}{1.21} & \textcolor{black}{13.86} & \textcolor{black}{4.73} & \textcolor{black}{37.35} & \textcolor{black}{52.66} & \textcolor{black}{9.52} & \textcolor{black}{22.85} & \textcolor{black}{-} \\
\textcolor{black}{\textbf{InstructDiffusion~\cite{geng2023instructdiffusion}}} & \textcolor{black}{3.03} & \textcolor{black}{10.84} & \textcolor{black}{5.33} & \textcolor{black}{28.31} & \textcolor{black}{46.75} & \textcolor{black}{14.88} & \textcolor{black}{20.78} & \textcolor{black}{-} \\
\textcolor{black}{\textbf{InstructPix2Pix~\cite{brooks2023instructpix2pix}}}   & \textcolor{black}{2.42} & \textcolor{black}{16.27} & \textcolor{black}{0.59} & \textcolor{black}{26.51} & \textcolor{black}{44.97} & \textcolor{black}{10.71} & \textcolor{black}{23.15} & \textcolor{black}{-} \\
\textcolor{black}{\textbf{MagicBrush~\cite{zhang2024magicbrush}}}        & \textcolor{black}{1.82} & \textcolor{black}{20.48} & \textcolor{black}{2.96} & \textcolor{black}{38.55} & \textcolor{black}{69.64} & \textcolor{black}{13.10} & \textcolor{black}{23.34} & \textcolor{black}{-} \\
\textcolor{black}{\textbf{MGIE~\cite{fu2023guiding}}}              & \textcolor{black}{1.21} & \textcolor{black}{14.46} & \textcolor{black}{4.14} & \textcolor{black}{22.89} & \textcolor{black}{49.11} & \textcolor{black}{4.76} & \textcolor{black}{22.57} & \textcolor{black}{-} \\
\textcolor{black}{\textbf{InstructEdit~\cite{wang2023instructedit}}}      & \textcolor{black}{0.61} & \textcolor{black}{11.45} & \textcolor{black}{0.00} & \textcolor{black}{0.60} & \textcolor{black}{33.73} & \textcolor{black}{0.60} & \textcolor{black}{19.81} & \textcolor{black}{-} \\
\textcolor{black}{\textbf{InstructAny2Pix~\cite{li2023instructany2pix}}}   & \textcolor{black}{1.82} & \textcolor{black}{21.08} & \textcolor{black}{17.75} & \textcolor{black}{33.13} & \textcolor{black}{52.07} & \textcolor{black}{2.38} & \textcolor{black}{25.23} & \textcolor{black}{-} \\
\textcolor{black}{\textbf{HQ-Edit~\cite{hui2024hq}}} & \textcolor{black}{\textbf{7.88}} & \textcolor{black}{20.48} & \textcolor{black}{6.51} & \textcolor{black}{28.31} & \textcolor{black}{60.95} & \textcolor{black}{4.76} & \textcolor{black}{24.85} & \textcolor{black}{-} \\
\textcolor{black}{\textbf{FLUX Kontext~\cite{batifol2025flux}}} & \textcolor{black}{0.61} & \textcolor{black}{37.95} & \textcolor{black}{21.89} & \textcolor{black}{58.43} & \textcolor{black}{62.72} & \textcolor{black}{60.12} & \textcolor{black}{25.31} & \textcolor{black}{-} \\
\textcolor{black}{\textbf{Qwen-image~\cite{wu2025qwen}}} & \textcolor{black}{4.85} & \textcolor{black}{\textbf{53.01}} & \textcolor{black}{\textbf{39.05}} & \textcolor{black}{\textbf{68.07}} & \textcolor{black}{\textbf{71.60}} & \textcolor{black}{\textbf{67.86}} & \textcolor{black}{\textbf{26.29}} & \textcolor{black}{-} \\
\hline
\rowcolor[HTML]{F2F2F2} 
\textcolor{black}{\textbf{Exp Min}}           & \textcolor{black}{0.00} & \textcolor{black}{0.00} & \textcolor{black}{0.00} & \textcolor{black}{0.00} & \textcolor{black}{0.00} & \textcolor{black}{0.00} & \textcolor{black}{11.49} & \textcolor{black}{-} \\
\rowcolor[HTML]{F2F2F2} 
\textcolor{black}{\textbf{Exp Max}}           & \textcolor{black}{100.00} & \textcolor{black}{100.00} & \textcolor{black}{100.00} & \textcolor{black}{100.00} & \textcolor{black}{100.00} & \textcolor{black}{100.00} & \textcolor{black}{33.34} & \textcolor{black}{-} \\
\midrule
\end{tabular}
 }
\label{tab:diverse}
\end{table*}

% %%%%%%%%%%%%%qwen3vl 8b%%%%%%%%%%%%
\begin{table*}[!t]
\caption{
\textcolor{black}{
\benchname\ evaluation results per dimension using original instructions, where Qwen3VL-8B~\cite{bai2025qwen3} is used as the judge model.  
Abbreviations: Deb. = Deblurring; Haze Rem. = Haze Removal; Lowl. Enh. = Lowlight Enhancement; Noise Rem. = Noise Removal; Rain Rem. = Rain Removal; Shadow Rem. = Shadow Removal; Snow Rem. = Snow Removal; Wtrmk Rem. = Watermark Removal; Cnt. = Counting; Dir. Perc. = Direction Perception; Obj. Rem. = Object Removal; Obj. Repl. = Object Replacement; Bkg. Repl. = Background Replacement; Col. Alt. = Color Alteration; Sty. Alt. = Style Alteration; Reg. Acc. = Region Accuracy.
}
}
\setlength{\tabcolsep}{3pt}
\renewcommand{\arraystretch}{1.0}
\centering
\resizebox{\textwidth}{!}{
\begin{tabular}{l|c|c|c|c|c|c|c|c}
\midrule
\multicolumn{9}{c}{\textcolor{black}{\textbf{Low-level Editing}}} \\
\midrule
\textcolor{black}{\textbf{Model}} & \textcolor{black}{\textbf{Deb.}} & \textcolor{black}{\textbf{Haze Rem.}} & \textcolor{black}{\textbf{Lowl. Enh.}} & \textcolor{black}{\textbf{Noise Rem.}} & \textcolor{black}{\textbf{Rain Rem.}} & \textcolor{black}{\textbf{Shadow Rem.}} & \textcolor{black}{\textbf{Snow Rem.}} & \textcolor{black}{\textbf{Wtrmk Rem.}} \\
\midrule
\textcolor{black}{\textbf{HIVE~\cite{zhang2023hive}}} & \textcolor{black}{44.25} & \textcolor{black}{54.89} & \textcolor{black}{37.61} & \textcolor{black}{24.59} & \textcolor{black}{45.47} & \textcolor{black}{37.61} & \textcolor{black}{51.49} & \textcolor{black}{49.99} \\
\textcolor{black}{\textbf{InstructDiffusion~\cite{geng2023instructdiffusion}}} & \textcolor{black}{42.48} & \textcolor{black}{58.45} & \textcolor{black}{56.61} & \textcolor{black}{28.60} & \textcolor{black}{67.20} & \textcolor{black}{37.43} & \textcolor{black}{55.65} & \textcolor{black}{85.49} \\
\textcolor{black}{\textbf{InstructPix2Pix~\cite{brooks2023instructpix2pix}}} & \textcolor{black}{48.03} & \textcolor{black}{56.15} & \textcolor{black}{43.32} & \textcolor{black}{20.11} & \textcolor{black}{56.64} & \textcolor{black}{34.19} & \textcolor{black}{57.59} & \textcolor{black}{58.12} \\
\textcolor{black}{\textbf{MagicBrush~\cite{zhang2024magicbrush}}} & \textcolor{black}{48.38} & \textcolor{black}{59.46} & \textcolor{black}{37.71} & \textcolor{black}{20.59} & \textcolor{black}{60.60} & \textcolor{black}{\textbf{41.91}} & \textcolor{black}{\textbf{57.81}} & \textcolor{black}{63.33} \\
\textcolor{black}{\textbf{MGIE~\cite{fu2023guiding}}} & \textcolor{black}{\textbf{60.30}} & \textcolor{black}{51.75} & \textcolor{black}{39.99} & \textcolor{black}{23.25} & \textcolor{black}{56.00} & \textcolor{black}{36.91} & \textcolor{black}{34.04} & \textcolor{black}{55.53} \\
\textcolor{black}{\textbf{InstructEdit~\cite{wang2023instructedit}}} & \textcolor{black}{40.77} & \textcolor{black}{58.85} & \textcolor{black}{13.83} & \textcolor{black}{15.40} & \textcolor{black}{64.44} & \textcolor{black}{36.88} & \textcolor{black}{43.45} & \textcolor{black}{82.68} \\
\textcolor{black}{\textbf{InstructAny2Pix~\cite{li2023instructany2pix}}} & \textcolor{black}{34.34} & \textcolor{black}{47.27} & \textcolor{black}{18.03} & \textcolor{black}{22.89} & \textcolor{black}{49.94} & \textcolor{black}{35.84} & \textcolor{black}{42.97} & \textcolor{black}{47.28} \\
\textcolor{black}{\textbf{HQ-Edit~\cite{hui2024hq}}} & \textcolor{black}{35.27} & \textcolor{black}{39.25} & \textcolor{black}{41.71} & \textcolor{black}{22.13} & \textcolor{black}{38.52} & \textcolor{black}{33.13} & \textcolor{black}{38.97} & \textcolor{black}{29.80} \\
\textcolor{black}{\textbf{FLUX Kontext~\cite{batifol2025flux}}} & \textcolor{black}{33.35} & \textcolor{black}{\textbf{66.30}} & \textcolor{black}{34.42} & \textcolor{black}{28.23} & \textcolor{black}{\textbf{79.71}} & \textcolor{black}{31.69} & \textcolor{black}{45.17} & \textcolor{black}{\textbf{92.24}} \\
\textcolor{black}{\textbf{Qwen-image~\cite{wu2025qwen}}} & \textcolor{black}{43.35} & \textcolor{black}{47.19} & \textcolor{black}{\textbf{73.73}} & \textcolor{black}{\textbf{52.75}} & \textcolor{black}{53.38} & \textcolor{black}{38.77} & \textcolor{black}{57.68} & \textcolor{black}{47.10} \\
\hline
\rowcolor[HTML]{F2F2F2}
\textcolor{black}{\textbf{Exp Min}} & \textcolor{black}{9.97} & \textcolor{black}{12.66} & \textcolor{black}{0.09} & \textcolor{black}{0.79} & \textcolor{black}{7.38} & \textcolor{black}{1.05} & \textcolor{black}{2.18} & \textcolor{black}{1.34} \\
\rowcolor[HTML]{F2F2F2}
\textcolor{black}{\textbf{Exp Max}} & \textcolor{black}{91.94} & \textcolor{black}{96.43} & \textcolor{black}{92.05} & \textcolor{black}{92.03} & \textcolor{black}{98.41} & \textcolor{black}{93.5} & \textcolor{black}{89.26} & \textcolor{black}{98.54} \\
\midrule

\multicolumn{9}{c}{\textcolor{black}{\textbf{High-level Editing}}} \\
\midrule
\textcolor{black}{\textbf{Model}} & \textcolor{black}{\textbf{Cnt.}} & \textcolor{black}{\textbf{Dir. Perc.}} & \textcolor{black}{\textbf{Obj. Rem.}} & \textcolor{black}{\textbf{Obj. Repl.}} & \textcolor{black}{\textbf{Bkg. Repl.}} & \textcolor{black}{\textbf{Col. Alt.}} & \textcolor{black}{\textbf{Sty. Alt.}} & \textcolor{black}{\textbf{Reg. Acc.}} \\
\midrule
\textcolor{black}{\textbf{HIVE~\cite{zhang2023hive}}} & \textcolor{black}{15.56} & \textcolor{black}{53.68} & \textcolor{black}{45.32} & \textcolor{black}{91.18} & \textcolor{black}{76.26} & \textcolor{black}{38.41} & \textcolor{black}{25.31} & \textcolor{black}{58.15} \\
\textcolor{black}{\textbf{InstructDiffusion~\cite{geng2023instructdiffusion}}} & \textcolor{black}{15.56} & \textcolor{black}{47.06} & \textcolor{black}{69.06} & \textcolor{black}{50.00} & \textcolor{black}{61.15} & \textcolor{black}{63.77} & \textcolor{black}{21.68} & \textcolor{black}{66.18} \\
\textcolor{black}{\textbf{InstructPix2Pix~\cite{brooks2023instructpix2pix}}} & \textcolor{black}{18.52} & \textcolor{black}{47.79} & \textcolor{black}{23.74} & \textcolor{black}{45.59} & \textcolor{black}{67.63} & \textcolor{black}{52.17} & \textcolor{black}{23.76} & \textcolor{black}{61.63} \\
\textcolor{black}{\textbf{MagicBrush~\cite{zhang2024magicbrush}}} & \textcolor{black}{31.85} & \textcolor{black}{50.74} & \textcolor{black}{30.22} & \textcolor{black}{67.65} & \textcolor{black}{88.49} & \textcolor{black}{56.52} & \textcolor{black}{22.78} & \textcolor{black}{66.34} \\
\textcolor{black}{\textbf{MGIE~\cite{fu2023guiding}}} & \textcolor{black}{16.30} & \textcolor{black}{50.00} & \textcolor{black}{36.69} & \textcolor{black}{76.47} & \textcolor{black}{89.21} & \textcolor{black}{41.30} & \textcolor{black}{23.67} & \textcolor{black}{69.60} \\
\textcolor{black}{\textbf{InstructEdit~\cite{wang2023instructedit}}} & \textcolor{black}{10.37} & \textcolor{black}{47.79} & \textcolor{black}{2.88} & \textcolor{black}{2.21} & \textcolor{black}{49.64} & \textcolor{black}{5.07} & \textcolor{black}{19.82} & \textcolor{black}{\textbf{77.08}} \\
\textcolor{black}{\textbf{InstructAny2Pix~\cite{li2023instructany2pix}}} & \textcolor{black}{26.67} & \textcolor{black}{45.59} & \textcolor{black}{48.92} & \textcolor{black}{39.71} & \textcolor{black}{64.75} & \textcolor{black}{9.42} & \textcolor{black}{26.76} & \textcolor{black}{52.75} \\
\textcolor{black}{\textbf{HQ-Edit~\cite{hui2024hq}}} & \textcolor{black}{20.74} & \textcolor{black}{52.94} & \textcolor{black}{27.34} & \textcolor{black}{47.79} & \textcolor{black}{70.50} & \textcolor{black}{27.54} & \textcolor{black}{25.02} & \textcolor{black}{49.21} \\
\textcolor{black}{\textbf{FLUX Kontext~\cite{batifol2025flux}}} & \textcolor{black}{50.37} & \textcolor{black}{72.79} & \textcolor{black}{64.03} & \textcolor{black}{61.03} & \textcolor{black}{91.37} & \textcolor{black}{75.36} & \textcolor{black}{26.33} & \textcolor{black}{53.14} \\
\textcolor{black}{\textbf{Qwen-image~\cite{wu2025qwen}}} & \textcolor{black}{\textbf{66.67}} & \textcolor{black}{\textbf{77.21}} & \textcolor{black}{\textbf{75.54}} & \textcolor{black}{\textbf{88.97}} & \textcolor{black}{\textbf{94.96}} & \textcolor{black}{\textbf{78.26}} & \textcolor{black}{\textbf{27.18}} & \textcolor{black}{71.07} \\
\hline
\rowcolor[HTML]{F2F2F2}
\textcolor{black}{\textbf{Exp Min}} & \textcolor{black}{0.00} & \textcolor{black}{0.00} & \textcolor{black}{0.00} & \textcolor{black}{0.00} & \textcolor{black}{0.00} & \textcolor{black}{0.00} & \textcolor{black}{12.55} & \textcolor{black}{6.41} \\
\rowcolor[HTML]{F2F2F2}
\textcolor{black}{\textbf{Exp Max}} & \textcolor{black}{100.00} & \textcolor{black}{100.00} & \textcolor{black}{100.00} & \textcolor{black}{100.00} & \textcolor{black}{100.00} & \textcolor{black}{100.00} & \textcolor{black}{33.91} & \textcolor{black}{98.70} \\
\midrule

\multicolumn{9}{c}{\textcolor{black}{\textbf{Multi-round Editing}}} \\
\midrule
\textcolor{black}{\textbf{Model}} & \textcolor{black}{\textbf{Cnt.}} & \textcolor{black}{\textbf{Dir. Perc.}} & \textcolor{black}{\textbf{Obj. Rem.}} & \textcolor{black}{\textbf{Obj. Repl.}} & \textcolor{black}{\textbf{Bkg. Repl.}} & \textcolor{black}{\textbf{Col. Alt.}} & \textcolor{black}{\textbf{Sty. Alt.}} & \textcolor{black}{\textbf{Reg. Acc.}} \\
\midrule
\textcolor{black}{\textbf{HIVE~\cite{zhang2023hive}}} & \textcolor{black}{1.21} & \textcolor{black}{27.71} & \textcolor{black}{30.77} & \textcolor{black}{\textbf{79.52}} & \textcolor{black}{49.70} & \textcolor{black}{8.93} & \textcolor{black}{24.84} & \textcolor{black}{-} \\
\textcolor{black}{\textbf{InstructDiffusion~\cite{geng2023instructdiffusion}}} & \textcolor{black}{0.61} & \textcolor{black}{22.89} & \textcolor{black}{38.46} & \textcolor{black}{37.35} & \textcolor{black}{42.60} & \textcolor{black}{41.67} & \textcolor{black}{22.09} & \textcolor{black}{-} \\
\textcolor{black}{\textbf{InstructPix2Pix~\cite{brooks2023instructpix2pix}}} & \textcolor{black}{1.82} & \textcolor{black}{22.29} & \textcolor{black}{11.83} & \textcolor{black}{32.53} & \textcolor{black}{50.89} & \textcolor{black}{27.98} & \textcolor{black}{23.84} & \textcolor{black}{-} \\
\textcolor{black}{\textbf{MagicBrush~\cite{zhang2024magicbrush}}} & \textcolor{black}{4.85} & \textcolor{black}{23.49} & \textcolor{black}{13.02} & \textcolor{black}{50.00} & \textcolor{black}{73.96} & \textcolor{black}{33.33} & \textcolor{black}{22.96} & \textcolor{black}{-} \\
\textcolor{black}{\textbf{MGIE~\cite{fu2023guiding}}} & \textcolor{black}{3.64} & \textcolor{black}{30.72} & \textcolor{black}{20.12} & \textcolor{black}{56.63} & \textcolor{black}{53.85} & \textcolor{black}{14.29} & \textcolor{black}{23.95} & \textcolor{black}{-} \\
\textcolor{black}{\textbf{InstructEdit~\cite{wang2023instructedit}}} & \textcolor{black}{0.00} & \textcolor{black}{24.10} & \textcolor{black}{0.00} & \textcolor{black}{1.20} & \textcolor{black}{34.32} & \textcolor{black}{0.00} & \textcolor{black}{19.88} & \textcolor{black}{-} \\
\textcolor{black}{\textbf{InstructAny2Pix~\cite{li2023instructany2pix}}} & \textcolor{black}{2.42} & \textcolor{black}{23.49} & \textcolor{black}{24.26} & \textcolor{black}{36.75} & \textcolor{black}{56.80} & \textcolor{black}{1.79} & \textcolor{black}{27.09} & \textcolor{black}{-} \\
\textcolor{black}{\textbf{HQ-Edit~\cite{hui2024hq}}} & \textcolor{black}{1.82} & \textcolor{black}{27.71} & \textcolor{black}{13.02} & \textcolor{black}{24.10} & \textcolor{black}{55.62} & \textcolor{black}{4.76} & \textcolor{black}{25.08} & \textcolor{black}{-} \\
\textcolor{black}{\textbf{FLUX Kontext~\cite{batifol2025flux}}} & \textcolor{black}{1.82} & \textcolor{black}{42.77} & \textcolor{black}{32.54} & \textcolor{black}{36.14} & \textcolor{black}{74.56} & \textcolor{black}{49.40} & \textcolor{black}{26.84} & \textcolor{black}{-} \\
\textcolor{black}{\textbf{Qwen-image~\cite{wu2025qwen}}} & \textcolor{black}{\textbf{12.73}} & \textcolor{black}{\textbf{54.82}} & \textcolor{black}{\textbf{40.24}} & \textcolor{black}{63.86} & \textcolor{black}{\textbf{82.25}} & \textcolor{black}{\textbf{61.31}} & \textcolor{black}{\textbf{27.70}} & \textcolor{black}{-} \\
\hline
\rowcolor[HTML]{F2F2F2}
\textcolor{black}{\textbf{Exp Min}} & \textcolor{black}{0.00} & \textcolor{black}{0.00} & \textcolor{black}{0.00} & \textcolor{black}{0.00} & \textcolor{black}{0.00} & \textcolor{black}{0.00} & \textcolor{black}{13.12} & \textcolor{black}{-} \\
\rowcolor[HTML]{F2F2F2}
\textcolor{black}{\textbf{Exp Max}} & \textcolor{black}{100.00} & \textcolor{black}{100.00} & \textcolor{black}{100.00} & \textcolor{black}{100.00} & \textcolor{black}{100.00} & \textcolor{black}{100.00} & \textcolor{black}{33.84} & \textcolor{black}{-} \\
\midrule
\end{tabular}
}
\label{tab:original_qwen3vl}
\end{table*}

\begin{table*}[!t]
\caption{
\textcolor{black}{
\benchname\ evaluation results per dimension using diverse instructions, where Qwen3VL-8B~\cite{bai2025qwen3} is used as the judge model.  
Abbreviations: Deb. = Deblurring; Haze Rem. = Haze Removal; Lowl. Enh. = Lowlight Enhancement; Noise Rem. = Noise Removal; Rain Rem. = Rain Removal; Shadow Rem. = Shadow Removal; Snow Rem. = Snow Removal; Wtrmk Rem. = Watermark Removal; Cnt. = Counting; Dir. Perc. = Direction Perception; Obj. Rem. = Object Removal; Obj. Repl. = Object Replacement; Bkg. Repl. = Background Replacement; Col. Alt. = Color Alteration; Sty. Alt. = Style Alteration; Reg. Acc. = Region Accuracy.
}
}
\setlength{\tabcolsep}{3pt}
\renewcommand{\arraystretch}{1.0}
\centering
\resizebox{\textwidth}{!}{
\begin{tabular}{l|c|c|c|c|c|c|c|c}
\midrule
\multicolumn{9}{c}{\textcolor{black}{\textbf{Low-level Editing}}} \\
\midrule
\textcolor{black}{\textbf{Model}} & \textcolor{black}{\textbf{Deb.}} & \textcolor{black}{\textbf{Haze Rem.}} & \textcolor{black}{\textbf{Lowl. Enh.}} & \textcolor{black}{\textbf{Noise Rem.}} & \textcolor{black}{\textbf{Rain Rem.}} & \textcolor{black}{\textbf{Shadow Rem.}} & \textcolor{black}{\textbf{Snow Rem.}} & \textcolor{black}{\textbf{Wtrmk Rem.}} \\
\midrule
\textcolor{black}{\textbf{HIVE~\cite{zhang2023hive}}} & \textcolor{black}{44.41} & \textcolor{black}{54.09} & \textcolor{black}{42.78} & \textcolor{black}{25.51} & \textcolor{black}{58.59} & \textcolor{black}{36.69} & \textcolor{black}{51.92} & \textcolor{black}{57.88} \\
\textcolor{black}{\textbf{InstructDiffusion~\cite{geng2023instructdiffusion}}} & \textcolor{black}{42.62} & \textcolor{black}{58.01} & \textcolor{black}{39.47} & \textcolor{black}{28.06} & \textcolor{black}{64.18} & \textcolor{black}{32.54} & \textcolor{black}{\textbf{57.30}} & \textcolor{black}{85.14} \\
\textcolor{black}{\textbf{InstructPix2Pix~\cite{brooks2023instructpix2pix}}} & \textcolor{black}{45.24} & \textcolor{black}{53.52} & \textcolor{black}{42.88} & \textcolor{black}{24.49} & \textcolor{black}{51.86} & \textcolor{black}{32.79} & \textcolor{black}{52.67} & \textcolor{black}{48.91} \\
\textcolor{black}{\textbf{MagicBrush~\cite{zhang2024magicbrush}}} & \textcolor{black}{45.96} & \textcolor{black}{55.11} & \textcolor{black}{33.74} & \textcolor{black}{23.91} & \textcolor{black}{55.77} & \textcolor{black}{36.73} & \textcolor{black}{54.68} & \textcolor{black}{59.76} \\
\textcolor{black}{\textbf{MGIE~\cite{fu2023guiding}}} & \textcolor{black}{57.33} & \textcolor{black}{51.61} & \textcolor{black}{32.96} & \textcolor{black}{23.49} & \textcolor{black}{58.27} & \textcolor{black}{34.07} & \textcolor{black}{51.02} & \textcolor{black}{59.64} \\
\textcolor{black}{\textbf{InstructEdit~\cite{wang2023instructedit}}} & \textcolor{black}{40.66} & \textcolor{black}{58.89} & \textcolor{black}{13.92} & \textcolor{black}{15.81} & \textcolor{black}{65.08} & \textcolor{black}{36.66} & \textcolor{black}{43.34} & \textcolor{black}{83.68} \\
\textcolor{black}{\textbf{InstructAny2Pix~\cite{li2023instructany2pix}}} & \textcolor{black}{34.77} & \textcolor{black}{47.00} & \textcolor{black}{18.09} & \textcolor{black}{22.18} & \textcolor{black}{48.92} & \textcolor{black}{36.04} & \textcolor{black}{43.13} & \textcolor{black}{47.58} \\
\textcolor{black}{\textbf{HQ-Edit~\cite{hui2024hq}}} & \textcolor{black}{34.11} & \textcolor{black}{37.95} & \textcolor{black}{36.76} & \textcolor{black}{22.38} & \textcolor{black}{37.60} & \textcolor{black}{32.17} & \textcolor{black}{38.45} & \textcolor{black}{30.83} \\
\textcolor{black}{\textbf{FLUX Kontext~\cite{batifol2025flux}}} & \textcolor{black}{32.11} & \textcolor{black}{\textbf{66.00}} & \textcolor{black}{27.14} & \textcolor{black}{27.11} & \textcolor{black}{\textbf{75.07}} & \textcolor{black}{32.04} & \textcolor{black}{43.67} & \textcolor{black}{\textbf{93.18}} \\
\textcolor{black}{\textbf{Qwen-image~\cite{wu2025qwen}}} & \textcolor{black}{\textbf{59.69}} & \textcolor{black}{49.12} & \textcolor{black}{\textbf{72.78}} & \textcolor{black}{\textbf{52.91}} & \textcolor{black}{54.70} & \textcolor{black}{\textbf{47.78}} & \textcolor{black}{{57.07}} & \textcolor{black}{65.10} \\
\hline
\rowcolor[HTML]{F2F2F2}
\textcolor{black}{\textbf{Exp Min}} & \textcolor{black}{6.32} & \textcolor{black}{3.67} & \textcolor{black}{0.60} & \textcolor{black}{0.03} & \textcolor{black}{7.22} & \textcolor{black}{1.46} & \textcolor{black}{3.78} & \textcolor{black}{2.58} \\
\rowcolor[HTML]{F2F2F2}
\textcolor{black}{\textbf{Exp Max}} & \textcolor{black}{91.10} & \textcolor{black}{96.59} & \textcolor{black}{90.66} & \textcolor{black}{92.61} & \textcolor{black}{98.31} & \textcolor{black}{93.23} & \textcolor{black}{90.83} & \textcolor{black}{98.60} \\
\midrule

\multicolumn{9}{c}{\textcolor{black}{\textbf{High-level Editing}}} \\
\midrule
\textcolor{black}{\textbf{Model}} & \textcolor{black}{\textbf{Cnt.}} & \textcolor{black}{\textbf{Dir. Perc.}} & \textcolor{black}{\textbf{Obj. Rem.}} & \textcolor{black}{\textbf{Obj. Repl.}} & \textcolor{black}{\textbf{Bkg. Repl.}} & \textcolor{black}{\textbf{Col. Alt.}} & \textcolor{black}{\textbf{Sty. Alt.}} & \textcolor{black}{\textbf{Reg. Acc.}} \\
\midrule
\textcolor{black}{\textbf{HIVE~\cite{zhang2023hive}}} & \textcolor{black}{11.85} & \textcolor{black}{50.00} & \textcolor{black}{10.07} & \textcolor{black}{70.59} & \textcolor{black}{84.17} & \textcolor{black}{38.41} & \textcolor{black}{23.08} & \textcolor{black}{61.97} \\
\textcolor{black}{\textbf{InstructDiffusion~\cite{geng2023instructdiffusion}}} & \textcolor{black}{21.48} & \textcolor{black}{44.12} & \textcolor{black}{21.58} & \textcolor{black}{58.09} & \textcolor{black}{65.47} & \textcolor{black}{{56.52}} & \textcolor{black}{19.96} & \textcolor{black}{65.92} \\
\textcolor{black}{\textbf{InstructPix2Pix~\cite{brooks2023instructpix2pix}}} & \textcolor{black}{20.00} & \textcolor{black}{52.21} & \textcolor{black}{6.47} & \textcolor{black}{53.68} & \textcolor{black}{69.78} & \textcolor{black}{51.45} & \textcolor{black}{23.13} & \textcolor{black}{61.32} \\
\textcolor{black}{\textbf{MagicBrush~\cite{zhang2024magicbrush}}} & \textcolor{black}{31.11} & \textcolor{black}{50.00} & \textcolor{black}{7.91} & \textcolor{black}{72.79} & \textcolor{black}{\textbf{93.53}} & \textcolor{black}{53.62} & \textcolor{black}{23.07} & \textcolor{black}{66.21} \\
\textcolor{black}{\textbf{MGIE~\cite{fu2023guiding}}} & \textcolor{black}{18.52} & \textcolor{black}{41.91} & \textcolor{black}{16.55} & \textcolor{black}{54.41} & \textcolor{black}{80.58} & \textcolor{black}{39.13} & \textcolor{black}{23.36} & \textcolor{black}{71.89} \\
\textcolor{black}{\textbf{InstructEdit~\cite{wang2023instructedit}}} & \textcolor{black}{7.41} & \textcolor{black}{46.32} & \textcolor{black}{3.60} & \textcolor{black}{3.68} & \textcolor{black}{51.08} & \textcolor{black}{6.52} & \textcolor{black}{19.90} & \textcolor{black}{\textbf{77.08}} \\
\textcolor{black}{\textbf{InstructAny2Pix~\cite{li2023instructany2pix}}} & \textcolor{black}{23.70} & \textcolor{black}{50.74} & \textcolor{black}{39.57} & \textcolor{black}{44.85} & \textcolor{black}{71.22} & \textcolor{black}{11.59} & \textcolor{black}{25.93} & \textcolor{black}{52.61} \\
\textcolor{black}{\textbf{HQ-Edit~\cite{hui2024hq}}} & \textcolor{black}{14.81} & \textcolor{black}{47.79} & \textcolor{black}{19.42} & \textcolor{black}{50.00} & \textcolor{black}{79.14} & \textcolor{black}{26.81} & \textcolor{black}{25.81} & \textcolor{black}{48.29} \\
\textcolor{black}{\textbf{FLUX Kontext~\cite{batifol2025flux}}} & \textcolor{black}{43.70} & \textcolor{black}{67.65} & \textcolor{black}{52.52} & \textcolor{black}{80.88} & \textcolor{black}{89.93} & \textcolor{black}{75.36} & \textcolor{black}{26.27} & \textcolor{black}{52.43} \\
\textcolor{black}{\textbf{Qwen-image~\cite{wu2025qwen}}} & \textcolor{black}{\textbf{57.78}} & \textcolor{black}{\textbf{80.15}} & \textcolor{black}{\textbf{69.78}} & \textcolor{black}{\textbf{96.32}} & \textcolor{black}{92.09} & \textcolor{black}{\textbf{76.09}} & \textcolor{black}{\textbf{27.42}} & \textcolor{black}{71.35} \\
\hline
\rowcolor[HTML]{F2F2F2}
\textcolor{black}{\textbf{Exp Min}} & \textcolor{black}{0.00} & \textcolor{black}{0.00} & \textcolor{black}{0.00} & \textcolor{black}{0.00} & \textcolor{black}{0.00} & \textcolor{black}{0.00} & \textcolor{black}{10.62} & \textcolor{black}{9.79} \\
\rowcolor[HTML]{F2F2F2}
\textcolor{black}{\textbf{Exp Max}} & \textcolor{black}{100.00} & \textcolor{black}{100.00} & \textcolor{black}{100.00} & \textcolor{black}{100.00} & \textcolor{black}{100.00} & \textcolor{black}{100.00} & \textcolor{black}{34.06} & \textcolor{black}{98.68} \\
\midrule

\multicolumn{9}{c}{\textcolor{black}{\textbf{Multi-round Editing}}} \\
\midrule
\textcolor{black}{\textbf{Model}} & \textcolor{black}{\textbf{Cnt.}} & \textcolor{black}{\textbf{Dir. Perc.}} & \textcolor{black}{\textbf{Obj. Rem.}} & \textcolor{black}{\textbf{Obj. Repl.}} & \textcolor{black}{\textbf{Bkg. Repl.}} & \textcolor{black}{\textbf{Col. Alt.}} & \textcolor{black}{\textbf{Sty. Alt.}} & \textcolor{black}{\textbf{Reg. Acc.}} \\
\midrule
\textcolor{black}{\textbf{HIVE~\cite{zhang2023hive}}} & \textcolor{black}{1.21} & \textcolor{black}{24.10} & \textcolor{black}{3.55} & \textcolor{black}{46.99} & \textcolor{black}{50.30} & \textcolor{black}{12.50} & \textcolor{black}{22.85} & \textcolor{black}{-} \\
\textcolor{black}{\textbf{InstructDiffusion~\cite{geng2023instructdiffusion}}} & \textcolor{black}{2.42} & \textcolor{black}{20.48} & \textcolor{black}{4.14} & \textcolor{black}{38.55} & \textcolor{black}{49.70} & \textcolor{black}{27.38} & \textcolor{black}{20.78} & \textcolor{black}{-} \\
\textcolor{black}{\textbf{InstructPix2Pix~\cite{brooks2023instructpix2pix}}} & \textcolor{black}{0.61} & \textcolor{black}{{27.11}} & \textcolor{black}{1.78} & \textcolor{black}{36.75} & \textcolor{black}{49.11} & \textcolor{black}{22.62} & \textcolor{black}{23.15} & \textcolor{black}{-} \\
\textcolor{black}{\textbf{MagicBrush~\cite{zhang2024magicbrush}}} & \textcolor{black}{2.42} & \textcolor{black}{27.71} & \textcolor{black}{0.59} & \textcolor{black}{50.00} & \textcolor{black}{\textbf{78.70}} & \textcolor{black}{28.57} & \textcolor{black}{23.34} & \textcolor{black}{-} \\
\textcolor{black}{\textbf{MGIE~\cite{fu2023guiding}}} & \textcolor{black}{3.03} & \textcolor{black}{19.88} & \textcolor{black}{4.73} & \textcolor{black}{34.94} & \textcolor{black}{53.25} & \textcolor{black}{10.12} & \textcolor{black}{22.57} & \textcolor{black}{-} \\
\textcolor{black}{\textbf{InstructEdit~\cite{wang2023instructedit}}} & \textcolor{black}{0.61} & \textcolor{black}{22.89} & \textcolor{black}{0.00} & \textcolor{black}{1.20} & \textcolor{black}{34.32} & \textcolor{black}{0.00} & \textcolor{black}{19.81} & \textcolor{black}{-} \\
\textcolor{black}{\textbf{InstructAny2Pix~\cite{li2023instructany2pix}}} & \textcolor{black}{1.21} & \textcolor{black}{24.70} & \textcolor{black}{15.98} & \textcolor{black}{28.92} & \textcolor{black}{60.36} & \textcolor{black}{2.98} & \textcolor{black}{25.24} & \textcolor{black}{-} \\
\textcolor{black}{\textbf{HQ-Edit~\cite{hui2024hq}}} & \textcolor{black}{2.42} & \textcolor{black}{27.11} & \textcolor{black}{8.28} & \textcolor{black}{30.72} & \textcolor{black}{63.31} & \textcolor{black}{6.55} & \textcolor{black}{24.85} & \textcolor{black}{-} \\
\textcolor{black}{\textbf{FLUX Kontext~\cite{batifol2025flux}}} & \textcolor{black}{0.61} & \textcolor{black}{34.34} & \textcolor{black}{21.30} & \textcolor{black}{56.63} & \textcolor{black}{62.72} & \textcolor{black}{{53.57}} & \textcolor{black}{{25.31}} & \textcolor{black}{-} \\
\textcolor{black}{\textbf{Qwen-image~\cite{wu2025qwen}}} & \textcolor{black}{\textbf{4.85}} & \textcolor{black}{\textbf{53.01}} & \textcolor{black}{\textbf{30.18}} & \textcolor{black}{\textbf{66.27}} & \textcolor{black}{71.01} & \textcolor{black}{\textbf{61.31}} & \textcolor{black}{\textbf{26.29}} & \textcolor{black}{-} \\
\hline
\rowcolor[HTML]{F2F2F2}
\textcolor{black}{\textbf{Exp Min}} & \textcolor{black}{0.00} & \textcolor{black}{0.00} & \textcolor{black}{0.00} & \textcolor{black}{0.00} & \textcolor{black}{0.00} & \textcolor{black}{0.00} & \textcolor{black}{11.49} & \textcolor{black}{-} \\
\rowcolor[HTML]{F2F2F2}
\textcolor{black}{\textbf{Exp Max}} & \textcolor{black}{100.00} & \textcolor{black}{100.00} & \textcolor{black}{100.00} & \textcolor{black}{100.00} & \textcolor{black}{100.00} & \textcolor{black}{100.00} & \textcolor{black}{33.34} & \textcolor{black}{-} \\
\midrule
\end{tabular}
}
\label{tab:diverse_qwen3vl}
\end{table*}

% %%%%%%%%%%%%%llava 1.5 7b%%%%%%%%%%%%

\begin{table*}[!t]
\caption{
\textcolor{black}{
\benchname\ evaluation results per dimension using original instructions, where LLaVA-1.5-7B~\cite{liu2023improvedllava} is used as the judge model.
Abbreviations: Deb. = Deblurring; Haze Rem. = Haze Removal; Lowl. Enh. = Lowlight Enhancement; Noise Rem. = Noise Removal; Rain Rem. = Rain Removal; Shadow Rem. = Shadow Removal; Snow Rem. = Snow Removal; Wtrmk Rem. = Watermark Removal; Cnt. = Counting; Dir. Perc. = Direction Perception; Obj. Rem. = Object Removal; Obj. Repl. = Object Replacement; Bkg. Repl. = Background Replacement; Col. Alt. = Color Alteration; Sty. Alt. = Style Alteration; Reg. Acc. = Region Accuracy.
}
}
\setlength{\tabcolsep}{3pt}
\renewcommand{\arraystretch}{1.0}
\centering
\resizebox{\textwidth}{!}{
\begin{tabular}{l|c|c|c|c|c|c|c|c}
\midrule
\multicolumn{9}{c}{\textcolor{black}{\textbf{Low-level Editing}}} \\
\midrule
\textcolor{black}{\textbf{Model}} & \textcolor{black}{\textbf{Deb.}} & \textcolor{black}{\textbf{Haze Rem.}} & \textcolor{black}{\textbf{Lowl. Enh.}} & \textcolor{black}{\textbf{Noise Rem.}} & \textcolor{black}{\textbf{Rain Rem.}} & \textcolor{black}{\textbf{Shadow Rem.}} & \textcolor{black}{\textbf{Snow Rem.}} & \textcolor{black}{\textbf{Wtrmk Rem.}} \\
\midrule
\textcolor{black}{\textbf{HIVE~\cite{zhang2023hive}}} & \textcolor{black}{44.25} & \textcolor{black}{54.89} & \textcolor{black}{37.61} & \textcolor{black}{24.59} & \textcolor{black}{45.47} & \textcolor{black}{37.61} & \textcolor{black}{51.49} & \textcolor{black}{49.99} \\
\textcolor{black}{\textbf{InstructDiffusion~\cite{geng2023instructdiffusion}}} & \textcolor{black}{42.48} & \textcolor{black}{58.45} & \textcolor{black}{56.61} & \textcolor{black}{28.60} & \textcolor{black}{67.20} & \textcolor{black}{37.43} & \textcolor{black}{55.65} & \textcolor{black}{85.49} \\
\textcolor{black}{\textbf{InstructPix2Pix~\cite{brooks2023instructpix2pix}}} & \textcolor{black}{48.03} & \textcolor{black}{56.15} & \textcolor{black}{43.32} & \textcolor{black}{20.11} & \textcolor{black}{56.64} & \textcolor{black}{34.19} & \textcolor{black}{57.59} & \textcolor{black}{58.12} \\
\textcolor{black}{\textbf{MagicBrush~\cite{zhang2024magicbrush}}} & \textcolor{black}{48.38} & \textcolor{black}{59.46} & \textcolor{black}{37.71} & \textcolor{black}{20.59} & \textcolor{black}{60.60} & \textcolor{black}{\textbf{41.91}} & \textcolor{black}{\textbf{57.81}} & \textcolor{black}{63.33} \\
\textcolor{black}{\textbf{MGIE~\cite{fu2023guiding}}} & \textcolor{black}{\textbf{60.30}} & \textcolor{black}{51.75} & \textcolor{black}{39.99} & \textcolor{black}{23.25} & \textcolor{black}{56.00} & \textcolor{black}{36.91} & \textcolor{black}{34.04} & \textcolor{black}{55.53} \\
\textcolor{black}{\textbf{InstructEdit~\cite{wang2023instructedit}}} & \textcolor{black}{40.77} & \textcolor{black}{58.85} & \textcolor{black}{13.83} & \textcolor{black}{15.40} & \textcolor{black}{64.44} & \textcolor{black}{36.88} & \textcolor{black}{43.45} & \textcolor{black}{82.68} \\
\textcolor{black}{\textbf{InstructAny2Pix~\cite{li2023instructany2pix}}} & \textcolor{black}{34.34} & \textcolor{black}{47.27} & \textcolor{black}{18.03} & \textcolor{black}{22.89} & \textcolor{black}{49.94} & \textcolor{black}{35.84} & \textcolor{black}{42.97} & \textcolor{black}{47.28} \\
\textcolor{black}{\textbf{HQ-Edit~\cite{hui2024hq}}} & \textcolor{black}{35.27} & \textcolor{black}{39.25} & \textcolor{black}{41.71} & \textcolor{black}{22.13} & \textcolor{black}{38.52} & \textcolor{black}{33.13} & \textcolor{black}{38.97} & \textcolor{black}{29.80} \\
\textcolor{black}{\textbf{FLUX Kontext~\cite{batifol2025flux}}} & \textcolor{black}{33.35} & \textcolor{black}{\textbf{66.30}} & \textcolor{black}{34.42} & \textcolor{black}{28.23} & \textcolor{black}{\textbf{79.71}} & \textcolor{black}{31.69} & \textcolor{black}{45.17} & \textcolor{black}{\textbf{92.24}} \\
\textcolor{black}{\textbf{Qwen-image~\cite{wu2025qwen}}} & \textcolor{black}{43.35} & \textcolor{black}{47.19} & \textcolor{black}{\textbf{73.73}} & \textcolor{black}{\textbf{52.75}} & \textcolor{black}{53.38} & \textcolor{black}{38.77} & \textcolor{black}{57.68} & \textcolor{black}{47.10} \\
\hline
\rowcolor[HTML]{F2F2F2}
\textcolor{black}{\textbf{Exp Min}} & \textcolor{black}{9.97} & \textcolor{black}{12.66} & \textcolor{black}{0.09} & \textcolor{black}{0.79} & \textcolor{black}{7.38} & \textcolor{black}{1.05} & \textcolor{black}{2.18} & \textcolor{black}{1.34} \\
\rowcolor[HTML]{F2F2F2}
\textcolor{black}{\textbf{Exp Max}} & \textcolor{black}{91.94} & \textcolor{black}{96.43} & \textcolor{black}{92.05} & \textcolor{black}{92.03} & \textcolor{black}{98.41} & \textcolor{black}{93.5} & \textcolor{black}{89.26} & \textcolor{black}{98.54} \\
\midrule

\multicolumn{9}{c}{\textcolor{black}{\textbf{High-level Editing}}} \\
\midrule
\textcolor{black}{\textbf{Model}} & \textcolor{black}{\textbf{Cnt.}} & \textcolor{black}{\textbf{Dir. Perc.}} & \textcolor{black}{\textbf{Obj. Rem.}} & \textcolor{black}{\textbf{Obj. Repl.}} & \textcolor{black}{\textbf{Bkg. Repl.}} & \textcolor{black}{\textbf{Col. Alt.}} & \textcolor{black}{\textbf{Sty. Alt.}} & \textcolor{black}{\textbf{Reg. Acc.}} \\
\midrule
\textcolor{black}{\textbf{HIVE~\cite{zhang2023hive}}} & \textcolor{black}{17.78} & \textcolor{black}{50.74} & \textcolor{black}{49.64} & \textcolor{black}{80.88} & \textcolor{black}{75.54} & \textcolor{black}{34.78} & \textcolor{black}{25.31} & \textcolor{black}{58.15} \\
\textcolor{black}{\textbf{InstructDiffusion~\cite{geng2023instructdiffusion}}} & \textcolor{black}{20.00} & \textcolor{black}{47.06} & \textcolor{black}{55.40} & \textcolor{black}{51.47} & \textcolor{black}{62.59} & \textcolor{black}{52.90} & \textcolor{black}{21.68} & \textcolor{black}{66.18} \\
\textcolor{black}{\textbf{InstructPix2Pix~\cite{brooks2023instructpix2pix}}} & \textcolor{black}{15.56} & \textcolor{black}{47.06} & \textcolor{black}{25.18} & \textcolor{black}{47.06} & \textcolor{black}{60.43} & \textcolor{black}{47.10} & \textcolor{black}{23.76} & \textcolor{black}{61.63} \\
\textcolor{black}{\textbf{MagicBrush~\cite{zhang2024magicbrush}}} & \textcolor{black}{27.41} & \textcolor{black}{46.32} & \textcolor{black}{35.25} & \textcolor{black}{63.24} & \textcolor{black}{81.29} & \textcolor{black}{48.55} & \textcolor{black}{22.78} & \textcolor{black}{66.34} \\
\textcolor{black}{\textbf{MGIE~\cite{fu2023guiding}}} & \textcolor{black}{17.78} & \textcolor{black}{47.06} & \textcolor{black}{34.53} & \textcolor{black}{70.59} & \textcolor{black}{86.33} & \textcolor{black}{40.58} & \textcolor{black}{23.67} & \textcolor{black}{69.60} \\
\textcolor{black}{\textbf{InstructEdit~\cite{wang2023instructedit}}} & \textcolor{black}{12.59} & \textcolor{black}{52.21} & \textcolor{black}{7.19} & \textcolor{black}{13.24} & \textcolor{black}{48.20} & \textcolor{black}{7.25} & \textcolor{black}{19.82} & \textcolor{black}{\textbf{77.08}} \\
\textcolor{black}{\textbf{InstructAny2Pix~\cite{li2023instructany2pix}}} & \textcolor{black}{25.19} & \textcolor{black}{48.53} & \textcolor{black}{49.64} & \textcolor{black}{41.91} & \textcolor{black}{66.19} & \textcolor{black}{15.22} & \textcolor{black}{26.76} & \textcolor{black}{52.75} \\
\textcolor{black}{\textbf{HQ-Edit~\cite{hui2024hq}}} & \textcolor{black}{16.30} & \textcolor{black}{45.59} & \textcolor{black}{28.78} & \textcolor{black}{46.32} & \textcolor{black}{71.22} & \textcolor{black}{30.43} & \textcolor{black}{25.02} & \textcolor{black}{49.21} \\
\textcolor{black}{\textbf{FLUX Kontext~\cite{batifol2025flux}}} & \textcolor{black}{40.74} & \textcolor{black}{65.44} & \textcolor{black}{61.87} & \textcolor{black}{60.29} & \textcolor{black}{86.33} & \textcolor{black}{62.32} & \textcolor{black}{26.33} & \textcolor{black}{53.14} \\
\textcolor{black}{\textbf{Qwen-image~\cite{wu2025qwen}}} & \textcolor{black}{\textbf{51.11}} & \textcolor{black}{\textbf{67.65}} & \textcolor{black}{\textbf{67.63}} & \textcolor{black}{\textbf{80.88}} & \textcolor{black}{\textbf{87.77}} & \textcolor{black}{\textbf{69.57}} & \textcolor{black}{\textbf{27.18}} & \textcolor{black}{71.07} \\
\hline
\rowcolor[HTML]{F2F2F2}
\textcolor{black}{\textbf{Exp Min}} & \textcolor{black}{0.00} & \textcolor{black}{0.00} & \textcolor{black}{0.00} & \textcolor{black}{0.00} & \textcolor{black}{0.00} & \textcolor{black}{0.00} & \textcolor{black}{12.55} & \textcolor{black}{6.41} \\
\rowcolor[HTML]{F2F2F2}
\textcolor{black}{\textbf{Exp Max}} & \textcolor{black}{100.00} & \textcolor{black}{100.00} & \textcolor{black}{100.00} & \textcolor{black}{100.00} & \textcolor{black}{100.00} & \textcolor{black}{100.00} & \textcolor{black}{33.91} & \textcolor{black}{98.70} \\
\midrule

\multicolumn{9}{c}{\textcolor{black}{\textbf{Multi-round Editing}}} \\
\midrule
\textcolor{black}{\textbf{Model}} & \textcolor{black}{\textbf{Cnt.}} & \textcolor{black}{\textbf{Dir. Perc.}} & \textcolor{black}{\textbf{Obj. Rem.}} & \textcolor{black}{\textbf{Obj. Repl.}} & \textcolor{black}{\textbf{Bkg. Repl.}} & \textcolor{black}{\textbf{Col. Alt.}} & \textcolor{black}{\textbf{Sty. Alt.}} & \textcolor{black}{\textbf{Reg. Acc.}} \\
\midrule
\textcolor{black}{\textbf{HIVE~\cite{zhang2023hive}}} & \textcolor{black}{0.61} & \textcolor{black}{22.89} & \textcolor{black}{33.14} & \textcolor{black}{\textbf{56.63}} & \textcolor{black}{43.79} & \textcolor{black}{7.74} & \textcolor{black}{24.84} & \textcolor{black}{-} \\
\textcolor{black}{\textbf{InstructDiffusion~\cite{geng2023instructdiffusion}}} & \textcolor{black}{4.24} & \textcolor{black}{21.08} & \textcolor{black}{27.22} & \textcolor{black}{36.14} & \textcolor{black}{44.38} & \textcolor{black}{26.79} & \textcolor{black}{22.09} & \textcolor{black}{-} \\
\textcolor{black}{\textbf{InstructPix2Pix~\cite{brooks2023instructpix2pix}}} & \textcolor{black}{1.82} & \textcolor{black}{20.48} & \textcolor{black}{11.83} & \textcolor{black}{21.69} & \textcolor{black}{42.60} & \textcolor{black}{17.86} & \textcolor{black}{23.84} & \textcolor{black}{-} \\
\textcolor{black}{\textbf{MagicBrush~\cite{zhang2024magicbrush}}} & \textcolor{black}{4.85} & \textcolor{black}{23.49} & \textcolor{black}{12.43} & \textcolor{black}{53.61} & \textcolor{black}{55.03} & \textcolor{black}{20.24} & \textcolor{black}{22.96} & \textcolor{black}{-} \\
\textcolor{black}{\textbf{MGIE~\cite{fu2023guiding}}} & \textcolor{black}{1.82} & \textcolor{black}{24.70} & \textcolor{black}{17.75} & \textcolor{black}{36.14} & \textcolor{black}{47.34} & \textcolor{black}{13.10} & \textcolor{black}{23.95} & \textcolor{black}{-} \\
\textcolor{black}{\textbf{InstructEdit~\cite{wang2023instructedit}}} & \textcolor{black}{0.00} & \textcolor{black}{28.92} & \textcolor{black}{2.37} & \textcolor{black}{9.04} & \textcolor{black}{33.73} & \textcolor{black}{0.60} & \textcolor{black}{19.88} & \textcolor{black}{-} \\
\textcolor{black}{\textbf{InstructAny2Pix~\cite{li2023instructany2pix}}} & \textcolor{black}{3.64} & \textcolor{black}{29.52} & \textcolor{black}{23.67} & \textcolor{black}{25.90} & \textcolor{black}{47.93} & \textcolor{black}{3.57} & \textcolor{black}{{27.09}} & \textcolor{black}{-} \\
\textcolor{black}{\textbf{HQ-Edit~\cite{hui2024hq}}} & \textcolor{black}{1.82} & \textcolor{black}{22.89} & \textcolor{black}{11.83} & \textcolor{black}{24.10} & \textcolor{black}{50.30} & \textcolor{black}{5.95} & \textcolor{black}{25.08} & \textcolor{black}{-} \\
\textcolor{black}{\textbf{FLUX Kontext~\cite{batifol2025flux}}} & \textcolor{black}{\textbf{7.88}} & \textcolor{black}{28.92} & \textcolor{black}{26.63} & \textcolor{black}{31.93} & \textcolor{black}{57.40} & \textcolor{black}{32.14} & \textcolor{black}{26.84} & \textcolor{black}{-} \\
\textcolor{black}{\textbf{Qwen-image~\cite{wu2025qwen}}} & \textcolor{black}{4.85} & \textcolor{black}{\textbf{35.54}} & \textcolor{black}{\textbf{35.50}} & \textcolor{black}{38.55} & \textcolor{black}{\textbf{72.19}} & \textcolor{black}{\textbf{46.43}} & \textcolor{black}{\textbf{27.70}} & \textcolor{black}{-} \\
\hline
\rowcolor[HTML]{F2F2F2}
\textcolor{black}{\textbf{Exp Min}} & \textcolor{black}{0.00} & \textcolor{black}{0.00} & \textcolor{black}{0.00} & \textcolor{black}{0.00} & \textcolor{black}{0.00} & \textcolor{black}{0.00} & \textcolor{black}{13.12} & \textcolor{black}{-} \\
\rowcolor[HTML]{F2F2F2}
\textcolor{black}{\textbf{Exp Max}} & \textcolor{black}{100.00} & \textcolor{black}{100.00} & \textcolor{black}{100.00} & \textcolor{black}{100.00} & \textcolor{black}{100.00} & \textcolor{black}{100.00} & \textcolor{black}{33.84} & \textcolor{black}{-} \\
\midrule
\end{tabular}
}
\label{tab:original_llava}
\end{table*}

\begin{table*}[!t]
\caption{
\textcolor{black}{
\benchname\ evaluation results per dimension using diverse instructions, where LLaVA-1.5-7B~\cite{liu2023improvedllava} is used as the judge model.
Abbreviations: Deb. = Deblurring; Haze Rem. = Haze Removal; Lowl. Enh. = Lowlight Enhancement; Noise Rem. = Noise Removal; Rain Rem. = Rain Removal; Shadow Rem. = Shadow Removal; Snow Rem. = Snow Removal; Wtrmk Rem. = Watermark Removal; Cnt. = Counting; Dir. Perc. = Direction Perception; Obj. Rem. = Object Removal; Obj. Repl. = Object Replacement; Bkg. Repl. = Background Replacement; Col. Alt. = Color Alteration; Sty. Alt. = Style Alteration; Reg. Acc. = Region Accuracy.
}
}
\setlength{\tabcolsep}{3pt}
\renewcommand{\arraystretch}{1.0}
\centering
\resizebox{\textwidth}{!}{
\begin{tabular}{l|c|c|c|c|c|c|c|c}
\midrule
\multicolumn{9}{c}{\textcolor{black}{\textbf{Low-level Editing}}} \\
\midrule
\textcolor{black}{\textbf{Model}} & \textcolor{black}{\textbf{Deb.}} & \textcolor{black}{\textbf{Haze Rem.}} & \textcolor{black}{\textbf{Lowl. Enh.}} & \textcolor{black}{\textbf{Noise Rem.}} & \textcolor{black}{\textbf{Rain Rem.}} & \textcolor{black}{\textbf{Shadow Rem.}} & \textcolor{black}{\textbf{Snow Rem.}} & \textcolor{black}{\textbf{Wtrmk Rem.}} \\
\midrule
\textcolor{black}{\textbf{HIVE~\cite{zhang2023hive}}} & \textcolor{black}{44.41} & \textcolor{black}{54.09} & \textcolor{black}{42.78} & \textcolor{black}{25.51} & \textcolor{black}{58.59} & \textcolor{black}{36.69} & \textcolor{black}{51.92} & \textcolor{black}{57.88} \\
\textcolor{black}{\textbf{InstructDiffusion~\cite{geng2023instructdiffusion}}} & \textcolor{black}{42.62} & \textcolor{black}{58.01} & \textcolor{black}{39.47} & \textcolor{black}{28.06} & \textcolor{black}{64.18} & \textcolor{black}{32.54} & \textcolor{black}{\textbf{57.30}} & \textcolor{black}{85.14} \\
\textcolor{black}{\textbf{InstructPix2Pix~\cite{brooks2023instructpix2pix}}} & \textcolor{black}{45.24} & \textcolor{black}{53.52} & \textcolor{black}{42.88} & \textcolor{black}{24.49} & \textcolor{black}{51.86} & \textcolor{black}{32.79} & \textcolor{black}{52.67} & \textcolor{black}{48.91} \\
\textcolor{black}{\textbf{MagicBrush~\cite{zhang2024magicbrush}}} & \textcolor{black}{45.96} & \textcolor{black}{55.11} & \textcolor{black}{33.74} & \textcolor{black}{23.91} & \textcolor{black}{55.77} & \textcolor{black}{36.73} & \textcolor{black}{54.68} & \textcolor{black}{59.76} \\
\textcolor{black}{\textbf{MGIE~\cite{fu2023guiding}}} & \textcolor{black}{57.33} & \textcolor{black}{51.61} & \textcolor{black}{32.96} & \textcolor{black}{23.49} & \textcolor{black}{58.27} & \textcolor{black}{34.07} & \textcolor{black}{51.02} & \textcolor{black}{59.64} \\
\textcolor{black}{\textbf{InstructEdit~\cite{wang2023instructedit}}} & \textcolor{black}{40.66} & \textcolor{black}{58.89} & \textcolor{black}{13.92} & \textcolor{black}{15.81} & \textcolor{black}{65.08} & \textcolor{black}{36.66} & \textcolor{black}{43.34} & \textcolor{black}{83.68} \\
\textcolor{black}{\textbf{InstructAny2Pix~\cite{li2023instructany2pix}}} & \textcolor{black}{34.77} & \textcolor{black}{47.00} & \textcolor{black}{18.09} & \textcolor{black}{22.18} & \textcolor{black}{48.92} & \textcolor{black}{36.04} & \textcolor{black}{43.13} & \textcolor{black}{47.58} \\
\textcolor{black}{\textbf{HQ-Edit~\cite{hui2024hq}}} & \textcolor{black}{34.11} & \textcolor{black}{37.95} & \textcolor{black}{36.76} & \textcolor{black}{22.38} & \textcolor{black}{37.60} & \textcolor{black}{32.17} & \textcolor{black}{38.45} & \textcolor{black}{30.83} \\
\textcolor{black}{\textbf{FLUX Kontext~\cite{batifol2025flux}}} & \textcolor{black}{32.11} & \textcolor{black}{\textbf{66.00}} & \textcolor{black}{27.14} & \textcolor{black}{27.11} & \textcolor{black}{\textbf{75.07}} & \textcolor{black}{32.04} & \textcolor{black}{43.67} & \textcolor{black}{\textbf{93.18}} \\
\textcolor{black}{\textbf{Qwen-image~\cite{wu2025qwen}}} & \textcolor{black}{\textbf{59.69}} & \textcolor{black}{49.12} & \textcolor{black}{\textbf{72.78}} & \textcolor{black}{\textbf{52.91}} & \textcolor{black}{54.70} & \textcolor{black}{\textbf{47.78}} & \textcolor{black}{57.07} & \textcolor{black}{65.10} \\
\hline
\rowcolor[HTML]{F2F2F2}
\textcolor{black}{\textbf{Exp Min}} & \textcolor{black}{6.32} & \textcolor{black}{3.67} & \textcolor{black}{0.60} & \textcolor{black}{0.03} & \textcolor{black}{7.22} & \textcolor{black}{1.46} & \textcolor{black}{3.78} & \textcolor{black}{2.58} \\
\rowcolor[HTML]{F2F2F2}
\textcolor{black}{\textbf{Exp Max}} & \textcolor{black}{91.10} & \textcolor{black}{96.59} & \textcolor{black}{90.66} & \textcolor{black}{92.61} & \textcolor{black}{98.31} & \textcolor{black}{93.23} & \textcolor{black}{90.83} & \textcolor{black}{98.60} \\
\midrule

\multicolumn{9}{c}{\textcolor{black}{\textbf{High-level Editing}}} \\
\midrule
\textcolor{black}{\textbf{Model}} & \textcolor{black}{\textbf{Cnt.}} & \textcolor{black}{\textbf{Dir. Perc.}} & \textcolor{black}{\textbf{Obj. Rem.}} & \textcolor{black}{\textbf{Obj. Repl.}} & \textcolor{black}{\textbf{Bkg. Repl.}} & \textcolor{black}{\textbf{Col. Alt.}} & \textcolor{black}{\textbf{Sty. Alt.}} & \textcolor{black}{\textbf{Reg. Acc.}} \\
\midrule
\textcolor{black}{\textbf{HIVE~\cite{zhang2023hive}}} & \textcolor{black}{13.33} & \textcolor{black}{45.59} & \textcolor{black}{17.27} & \textcolor{black}{64.71} & \textcolor{black}{79.86} & \textcolor{black}{34.78} & \textcolor{black}{23.08} & \textcolor{black}{61.97} \\
\textcolor{black}{\textbf{InstructDiffusion~\cite{geng2023instructdiffusion}}} & \textcolor{black}{19.26} & \textcolor{black}{40.44} & \textcolor{black}{21.58} & \textcolor{black}{56.62} & \textcolor{black}{60.43} & \textcolor{black}{50.72} & \textcolor{black}{19.96} & \textcolor{black}{65.92} \\
\textcolor{black}{\textbf{InstructPix2Pix~\cite{brooks2023instructpix2pix}}} & \textcolor{black}{16.30} & \textcolor{black}{41.91} & \textcolor{black}{12.95} & \textcolor{black}{46.32} & \textcolor{black}{65.47} & \textcolor{black}{42.75} & \textcolor{black}{23.13} & \textcolor{black}{61.32} \\
\textcolor{black}{\textbf{MagicBrush~\cite{zhang2024magicbrush}}} & \textcolor{black}{32.59} & \textcolor{black}{44.12} & \textcolor{black}{18.71} & \textcolor{black}{64.71} & \textcolor{black}{81.29} & \textcolor{black}{47.83} & \textcolor{black}{23.07} & \textcolor{black}{66.21} \\
\textcolor{black}{\textbf{MGIE~\cite{fu2023guiding}}} & \textcolor{black}{21.48} & \textcolor{black}{36.76} & \textcolor{black}{23.02} & \textcolor{black}{54.41} & \textcolor{black}{72.66} & \textcolor{black}{38.41} & \textcolor{black}{23.36} & \textcolor{black}{71.89} \\
\textcolor{black}{\textbf{InstructEdit~\cite{wang2023instructedit}}} & \textcolor{black}{11.11} & \textcolor{black}{44.12} & \textcolor{black}{13.67} & \textcolor{black}{15.44} & \textcolor{black}{47.48} & \textcolor{black}{8.70} & \textcolor{black}{19.90} & \textcolor{black}{\textbf{77.08}} \\
\textcolor{black}{\textbf{InstructAny2Pix~\cite{li2023instructany2pix}}} & \textcolor{black}{28.15} & \textcolor{black}{48.53} & \textcolor{black}{39.57} & \textcolor{black}{52.94} & \textcolor{black}{69.06} & \textcolor{black}{15.94} & \textcolor{black}{25.93} & \textcolor{black}{52.61} \\
\textcolor{black}{\textbf{HQ-Edit~\cite{hui2024hq}}} & \textcolor{black}{17.78} & \textcolor{black}{49.26} & \textcolor{black}{23.02} & \textcolor{black}{47.79} & \textcolor{black}{78.42} & \textcolor{black}{26.81} & \textcolor{black}{25.81} & \textcolor{black}{48.29} \\
\textcolor{black}{\textbf{FLUX Kontext~\cite{batifol2025flux}}} & \textcolor{black}{33.33} & \textcolor{black}{58.09} & \textcolor{black}{43.17} & \textcolor{black}{80.15} & \textcolor{black}{82.73} & \textcolor{black}{\textbf{68.12}} & \textcolor{black}{26.27} & \textcolor{black}{52.43} \\
\textcolor{black}{\textbf{Qwen-image~\cite{wu2025qwen}}} & \textcolor{black}{\textbf{47.41}} & \textcolor{black}{\textbf{65.44}} & \textcolor{black}{\textbf{65.47}} & \textcolor{black}{\textbf{91.91}} & \textcolor{black}{\textbf{86.33}} & \textcolor{black}{66.67} & \textcolor{black}{\textbf{27.42}} & \textcolor{black}{71.35} \\
\hline
\rowcolor[HTML]{F2F2F2}
\textcolor{black}{\textbf{Exp Min}} & \textcolor{black}{0.00} & \textcolor{black}{0.00} & \textcolor{black}{0.00} & \textcolor{black}{0.00} & \textcolor{black}{0.00} & \textcolor{black}{0.00} & \textcolor{black}{10.62} & \textcolor{black}{9.79} \\
\rowcolor[HTML]{F2F2F2}
\textcolor{black}{\textbf{Exp Max}} & \textcolor{black}{100.00} & \textcolor{black}{100.00} & \textcolor{black}{100.00} & \textcolor{black}{100.00} & \textcolor{black}{100.00} & \textcolor{black}{100.00} & \textcolor{black}{34.06} & \textcolor{black}{98.68} \\
\midrule

\multicolumn{9}{c}{\textcolor{black}{\textbf{Multi-round Editing}}} \\
\midrule
\textcolor{black}{\textbf{Model}} & \textcolor{black}{\textbf{Cnt.}} & \textcolor{black}{\textbf{Dir. Perc.}} & \textcolor{black}{\textbf{Obj. Rem.}} & \textcolor{black}{\textbf{Obj. Repl.}} & \textcolor{black}{\textbf{Bkg. Repl.}} & \textcolor{black}{\textbf{Col. Alt.}} & \textcolor{black}{\textbf{Sty. Alt.}} & \textcolor{black}{\textbf{Reg. Acc.}} \\
\midrule
\textcolor{black}{\textbf{HIVE~\cite{zhang2023hive}}} & \textcolor{black}{1.82} & \textcolor{black}{21.69} & \textcolor{black}{5.33} & \textcolor{black}{33.13} & \textcolor{black}{47.34} & \textcolor{black}{11.31} & \textcolor{black}{22.85} & \textcolor{black}{-} \\
\textcolor{black}{\textbf{InstructDiffusion~\cite{geng2023instructdiffusion}}} & \textcolor{black}{3.03} & \textcolor{black}{24.70} & \textcolor{black}{8.88} & \textcolor{black}{31.33} & \textcolor{black}{33.73} & \textcolor{black}{20.24} & \textcolor{black}{20.78} & \textcolor{black}{-} \\
\textcolor{black}{\textbf{InstructPix2Pix~\cite{brooks2023instructpix2pix}}} & \textcolor{black}{3.03} & \textcolor{black}{24.10} & \textcolor{black}{4.14} & \textcolor{black}{33.13} & \textcolor{black}{42.60} & \textcolor{black}{16.67} & \textcolor{black}{23.15} & \textcolor{black}{-} \\
\textcolor{black}{\textbf{MagicBrush~\cite{zhang2024magicbrush}}} & \textcolor{black}{4.24} & \textcolor{black}{22.29} & \textcolor{black}{4.14} & \textcolor{black}{\textbf{46.39}} & \textcolor{black}{\textbf{56.21}} & \textcolor{black}{21.43} & \textcolor{black}{23.34} & \textcolor{black}{-} \\
\textcolor{black}{\textbf{MGIE~\cite{fu2023guiding}}} & \textcolor{black}{2.42} & \textcolor{black}{21.08} & \textcolor{black}{7.10} & \textcolor{black}{27.11} & \textcolor{black}{44.97} & \textcolor{black}{8.33} & \textcolor{black}{22.57} & \textcolor{black}{-} \\
\textcolor{black}{\textbf{InstructEdit~\cite{wang2023instructedit}}} & \textcolor{black}{0.61} & \textcolor{black}{24.70} & \textcolor{black}{1.18} & \textcolor{black}{7.23} & \textcolor{black}{32.54} & \textcolor{black}{0.60} & \textcolor{black}{19.81} & \textcolor{black}{-} \\
\textcolor{black}{\textbf{InstructAny2Pix~\cite{li2023instructany2pix}}} & \textcolor{black}{4.85} & \textcolor{black}{24.10} & \textcolor{black}{18.93} & \textcolor{black}{30.12} & \textcolor{black}{{55.62}} & \textcolor{black}{3.57} & \textcolor{black}{25.24} & \textcolor{black}{-} \\
\textcolor{black}{\textbf{HQ-Edit~\cite{hui2024hq}}} & \textcolor{black}{4.85} & \textcolor{black}{25.90} & \textcolor{black}{9.47} & \textcolor{black}{29.52} & \textcolor{black}{53.85} & \textcolor{black}{5.95} & \textcolor{black}{24.85} & \textcolor{black}{-} \\
\textcolor{black}{\textbf{FLUX Kontext~\cite{batifol2025flux}}} & \textcolor{black}{3.03} & \textcolor{black}{28.31} & \textcolor{black}{16.57} & \textcolor{black}{41.57} & \textcolor{black}{54.44} & \textcolor{black}{37.50} & \textcolor{black}{25.31} & \textcolor{black}{-} \\
\textcolor{black}{\textbf{Qwen-image~\cite{wu2025qwen}}} & \textcolor{black}{\textbf{7.27}} & \textcolor{black}{\textbf{35.54}} & \textcolor{black}{\textbf{30.77}} & \textcolor{black}{40.96} & \textcolor{black}{55.03} & \textcolor{black}{\textbf{45.24}} & \textcolor{black}{\textbf{26.29}} & \textcolor{black}{-} \\
\hline
\rowcolor[HTML]{F2F2F2}
\textcolor{black}{\textbf{Exp Min}} & \textcolor{black}{0.00} & \textcolor{black}{0.00} & \textcolor{black}{0.00} & \textcolor{black}{0.00} & \textcolor{black}{0.00} & \textcolor{black}{0.00} & \textcolor{black}{11.49} & \textcolor{black}{-} \\
\rowcolor[HTML]{F2F2F2}
\textcolor{black}{\textbf{Exp Max}} & \textcolor{black}{100.00} & \textcolor{black}{100.00} & \textcolor{black}{100.00} & \textcolor{black}{100.00} & \textcolor{black}{100.00} & \textcolor{black}{100.00} & \textcolor{black}{33.34} & \textcolor{black}{-} \\
\midrule
\end{tabular}
}
\label{tab:diverse_llava}
\end{table*}

% %%%%%%%%%%%% Aesthetic %%%%%%%%%%%%%

\begin{table*}[!t]
\caption{
\textcolor{black}{
Evaluation Results for Aesthetic Quality Assessment in \benchname\ Across Different Dimensions Using Original Instructions. {Abbreviations:} Debl. = Deblurring; Haze Rem. = Haze Removal; Lowl. Enh. = Lowlight Enhancement; Noise Rem. = Noise Removal; Rain Rem. = Rain Removal; Shadow Rem. = Shadow Removal; Snow Rem. = Snow Removal; Wtrmk Rem. = Watermark Removal; Cnt. = Counting; Dir. Perc. = Direction Perception; Obj. Rem. = Object Removal; Obj. Repl. = Object Replacement; Bkg. Repl. = Background Replacement; Col. Alt. = Color Alteration; Sty. Alt. = Style Alteration; Reg. Acc. = Region Accuracy.
}
}
\setlength{\tabcolsep}{4pt}

\centering

\resizebox{\textwidth}{!}{%
\begin{tabular}{l|c|c|c|c|c|c|c|c}

\midrule

\multicolumn{9}{c}{\textcolor{black}{\textbf{Low-level Editing}}} \\ 

\midrule

\textcolor{black}{\textbf{Model}}             
& \textcolor{black}{\textbf{Debl.}} 
& \textcolor{black}{\textbf{Haze Rem.}} 
& \textcolor{black}{\textbf{Lowl. Enh.}} 
& \textcolor{black}{\textbf{Noise Rem.}}     
& \textcolor{black}{\textbf{Rain Rem.}}     
& \textcolor{black}{\textbf{Shadow Rem.}}   
& \textcolor{black}{\textbf{Snow Rem.}}     
& \textcolor{black}{\textbf{Wtrmk Rem.}} \\

\midrule

\textcolor{black}{\textbf{HIVE~\cite{zhang2023hive}}}                      & \textcolor{black}{4.34}  & \textcolor{black}{4.80}  & \textcolor{black}{4.18}  & \textcolor{black}{5.05}  & \textcolor{black}{4.70}  & \textcolor{black}{4.71}  & \textcolor{black}{5.81}  & \textcolor{black}{4.11}     \\
\textcolor{black}{\textbf{InstructDiffusion~\cite{geng2023instructdiffusion}}}         & \textcolor{black}{3.67}  & \textcolor{black}{4.72}  & \textcolor{black}{4.70}  & \textcolor{black}{5.16}  & \textcolor{black}{4.40}  & \textcolor{black}{4.19}  & \textcolor{black}{5.67}  & \textcolor{black}{5.05}     \\
\textcolor{black}{\textbf{InstructPix2Pix~\cite{brooks2023instructpix2pix}}}           & \textcolor{black}{3.82}  & \textcolor{black}{4.66}  & \textcolor{black}{3.61}  & \textcolor{black}{4.50}  & \textcolor{black}{4.85}  & \textcolor{black}{4.09}  & \textcolor{black}{6.03}  & \textcolor{black}{4.50}     \\
\textcolor{black}{\textbf{MagicBrush~\cite{zhang2024magicbrush}}}                & \textcolor{black}{3.74}  & \textcolor{black}{4.56}  & \textcolor{black}{3.80}  & \textcolor{black}{4.23}  & \textcolor{black}{4.75}  & \textcolor{black}{3.93}  & \textcolor{black}{5.80}  & \textcolor{black}{4.42}     \\
\textcolor{black}{\textbf{MGIE~\cite{fu2023guiding}}}                      & \textcolor{black}{3.86}  & \textcolor{black}{4.72}  & \textcolor{black}{4.83}  & \textcolor{black}{4.92}  & \textcolor{black}{4.73}  & \textcolor{black}{4.12}  & \textcolor{black}{5.59}  & \textcolor{black}{4.68}     \\
\textcolor{black}{\textbf{InstructEdit~\cite{wang2023instructedit}}}              & \textcolor{black}{3.83}  & \textcolor{black}{4.61}  & \textcolor{black}{3.90}  & \textcolor{black}{4.79}  & \textcolor{black}{4.94}  & \textcolor{black}{4.11}  & \textcolor{black}{6.41}  & \textcolor{black}{4.85}     \\
\textcolor{black}{\textbf{InstructAny2Pix~\cite{li2023instructany2pix}}}           & \textcolor{black}{5.74}  & \textcolor{black}{5.87}  & \textcolor{black}{5.84}  & \textcolor{black}{6.22}  & \textcolor{black}{6.13}  & \textcolor{black}{5.56}  & \textcolor{black}{6.47}  & \textcolor{black}{5.95}     \\
\textcolor{black}{\textbf{HQ-Edit~\cite{hui2024hq}}}                   & \textcolor{black}{4.71}  & \textcolor{black}{5.25}  & \textcolor{black}{4.82}  & \textcolor{black}{5.82}  & \textcolor{black}{5.23}  & \textcolor{black}{4.49}  & \textcolor{black}{6.00}  & \textcolor{black}{5.23}     \\

\midrule

\multicolumn{9}{c}{\textcolor{black}{\textbf{High-level Editing}}} \\

\midrule

\textcolor{black}{\textbf{Model}}             
& \textcolor{black}{\textbf{Cnt.}}   
& \textcolor{black}{\textbf{Dir. Perc.}} 
& \textcolor{black}{\textbf{Obj. Rem.}}       
& \textcolor{black}{\textbf{Obj. Repl.}} 
& \textcolor{black}{\textbf{Bkg. Repl.}} 
& \textcolor{black}{\textbf{Col. Alt.}} 
& \textcolor{black}{\textbf{Sty. Alt.}} 
& \textcolor{black}{\textbf{Reg. Acc.}}   \\

\midrule

\textcolor{black}{\textbf{HIVE~\cite{zhang2023hive}}}                      & \textcolor{black}{5.80}  & \textcolor{black}{5.72}  & \textcolor{black}{5.29}  & \textcolor{black}{5.61}  & \textcolor{black}{5.78}  & \textcolor{black}{5.86}  & \textcolor{black}{5.89}  & \textcolor{black}{5.71}     \\
\textcolor{black}{\textbf{InstructDiffusion~\cite{geng2023instructdiffusion}}}         & \textcolor{black}{5.51}  & \textcolor{black}{5.44}  & \textcolor{black}{4.82}  & \textcolor{black}{5.46}  & \textcolor{black}{5.60}  & \textcolor{black}{5.82}  & \textcolor{black}{5.57}  & \textcolor{black}{5.47}     \\
\textcolor{black}{\textbf{InstructPix2Pix~\cite{brooks2023instructpix2pix}}}           & \textcolor{black}{5.69}  & \textcolor{black}{5.74}  & \textcolor{black}{5.26}  & \textcolor{black}{5.68}  & \textcolor{black}{5.75}  & \textcolor{black}{5.86}  & \textcolor{black}{5.70}  & \textcolor{black}{5.68}     \\
\textcolor{black}{\textbf{MagicBrush~\cite{zhang2024magicbrush}}}                & \textcolor{black}{5.76}  & \textcolor{black}{5.77}  & \textcolor{black}{5.47}  & \textcolor{black}{5.58}  & \textcolor{black}{6.01}  & \textcolor{black}{5.96}  & \textcolor{black}{5.82}  & \textcolor{black}{5.73}     \\
\textcolor{black}{\textbf{MGIE~\cite{fu2023guiding}}}                      & \textcolor{black}{5.86}  & \textcolor{black}{5.73}  & \textcolor{black}{5.31}  & \textcolor{black}{5.45}  & \textcolor{black}{5.55}  & \textcolor{black}{5.90}  & \textcolor{black}{5.77}  & \textcolor{black}{5.63}     \\
\textcolor{black}{\textbf{InstructEdit~\cite{wang2023instructedit}}}              & \textcolor{black}{5.75}  & \textcolor{black}{5.65}  & \textcolor{black}{5.93}  & \textcolor{black}{5.74}  & \textcolor{black}{5.65}  & \textcolor{black}{5.95}  & \textcolor{black}{5.66}  & \textcolor{black}{5.85}     \\
\textcolor{black}{\textbf{InstructAny2Pix~\cite{li2023instructany2pix}}}           & \textcolor{black}{6.29}  & \textcolor{black}{6.26}  & \textcolor{black}{6.24}  & \textcolor{black}{6.21}  & \textcolor{black}{6.49}  & \textcolor{black}{6.36}  & \textcolor{black}{6.54}  & \textcolor{black}{6.30}     \\
\textcolor{black}{\textbf{HQ-Edit~\cite{hui2024hq}}}                   & \textcolor{black}{5.85}  & \textcolor{black}{5.84}  & \textcolor{black}{5.92}  & \textcolor{black}{6.00}  & \textcolor{black}{5.89}  & \textcolor{black}{6.12}  & \textcolor{black}{5.72}  & \textcolor{black}{5.83}     \\

\midrule

\multicolumn{9}{c}{\textcolor{black}{\textbf{Multi-round Editing}}} \\ 

\midrule

\textcolor{black}{\textbf{Model}}             
& \textcolor{black}{\textbf{Cnt.}}   
& \textcolor{black}{\textbf{Dir. Perc.}} 
& \textcolor{black}{\textbf{Obj. Rem.}}       
& \textcolor{black}{\textbf{Obj. Repl.}} 
& \textcolor{black}{\textbf{Bkg. Repl.}} 
& \textcolor{black}{\textbf{Col. Alt.}} 
& \textcolor{black}{\textbf{Sty. Alt.}} 
& \textcolor{black}{\textbf{Reg. Acc.}}   \\

\midrule

\textcolor{black}{\textbf{HIVE~\cite{zhang2023hive}}}                      & \textcolor{black}{5.23}  & \textcolor{black}{5.15}  & \textcolor{black}{4.43}  & \textcolor{black}{4.86}  & \textcolor{black}{5.06}  & \textcolor{black}{5.51}  & \textcolor{black}{5.15}  & \textcolor{black}{-}        \\
\textcolor{black}{\textbf{InstructDiffusion~\cite{geng2023instructdiffusion}}}         & \textcolor{black}{4.96}  & \textcolor{black}{4.74}  & \textcolor{black}{3.71}  & \textcolor{black}{4.95}  & \textcolor{black}{5.03}  & \textcolor{black}{5.46}  & \textcolor{black}{4.97}  & \textcolor{black}{-}        \\
\textcolor{black}{\textbf{InstructPix2Pix~\cite{brooks2023instructpix2pix}}}           & \textcolor{black}{5.29}  & \textcolor{black}{5.34}  & \textcolor{black}{4.38}  & \textcolor{black}{5.20}  & \textcolor{black}{5.10}  & \textcolor{black}{5.44}  & \textcolor{black}{5.01}  & \textcolor{black}{-}        \\
\textcolor{black}{\textbf{MagicBrush~\cite{zhang2024magicbrush}}}                & \textcolor{black}{5.46}  & \textcolor{black}{5.42}  & \textcolor{black}{4.64}  & \textcolor{black}{5.17}  & \textcolor{black}{5.74}  & \textcolor{black}{5.70}  & \textcolor{black}{5.63}  & \textcolor{black}{-}        \\
\textcolor{black}{\textbf{MGIE~\cite{fu2023guiding}}}                      & \textcolor{black}{5.55}  & \textcolor{black}{5.43}  & \textcolor{black}{4.56}  & \textcolor{black}{5.01}  & \textcolor{black}{4.63}  & \textcolor{black}{5.55}  & \textcolor{black}{4.94}  & \textcolor{black}{-}        \\
\textcolor{black}{\textbf{InstructEdit~\cite{wang2023instructedit}}}              & \textcolor{black}{5.43}  & \textcolor{black}{5.43}  & \textcolor{black}{5.75}  & \textcolor{black}{5.29}  & \textcolor{black}{5.29}  & \textcolor{black}{5.83}  & \textcolor{black}{5.27}  & \textcolor{black}{-}        \\
\textcolor{black}{\textbf{InstructAny2Pix~\cite{li2023instructany2pix}}}           & \textcolor{black}{6.38}  & \textcolor{black}{6.30}  & \textcolor{black}{6.28}  & \textcolor{black}{6.30}  & \textcolor{black}{6.55}  & \textcolor{black}{6.55}  & \textcolor{black}{6.80}  & \textcolor{black}{-}        \\
\textcolor{black}{\textbf{HQ-Edit~\cite{hui2024hq}}}                   & \textcolor{black}{5.50}  & \textcolor{black}{5.60}  & \textcolor{black}{5.52}  & \textcolor{black}{5.73}  & \textcolor{black}{5.64}  & \textcolor{black}{5.86}  & \textcolor{black}{5.47}  & \textcolor{black}{-}        \\

\midrule

\end{tabular}
}
\label{tab:original_aes}
\end{table*}

\section{Experiments}
\label{sec:exp}

\subsection{Editing Evaluation}
For each image and instruction, we utilize official codes and checkpoints from various models for image editing. We calculate the \benchname scores following the methodology described in Section~\ref{sec:eval}. The \benchname scores for original and diverse instructions are presented in Fig.~\ref{fig:score_radar}, Tab.~\ref{tab:original}, and Tab.~\ref{tab:diverse}, respectively.

\subsubsection{Evaluation Across Dimensions}
Our analysis indicates that none of the models excels across all the evaluation dimensions.
In the context of low-level editing tasks, InstructDiffusion~\cite{geng2023instructdiffusion} stands out with remarkable performance. It secures the highest scores in four out of seven low-level dimensions when assessed with the original instructions, and excels in three out of seven when evaluated with diverse instructions.
When it comes to high-level editing tasks, both MagicBrush~\cite{zhang2024magicbrush} and InstructAny2Pix~\cite{li2023instructany2pix} show commendable performance. MagicBrush leads in three metric dimensions using the original instructions, while InstructAny2Pix also takes the top spot in three dimensions with diverse instructions.
\textcolor{black}{%
Meanwhile, Qwen-image~\cite{wu2025qwen} consistently dominates most high-level and multi-round editing dimensions under both instruction settings, particularly in counting, direction perception, object-centric operations, and color alteration. This highlights its strong capability in following complex semantic instructions and maintaining edit fidelity across more challenging reasoning-intensive tasks.%
}
\textcolor{black}{%
In contrast, FLUX Kontext~\cite{batifol2025flux} exhibits competitive performance in watermark removal, rain removal, and several high-level semantics-related dimensions (e.g., direction perception and color alteration), suggesting that its strengths lie more in instruction-following and compositional edits than in fine-grained low-level restoration.%
}

\subsubsection{Evaluation Across Instructions}
In our evaluation, we utilize both original and diverse instructions to thoroughly assess the performance of models in various dimensions.
By examining the results presented in Tab.~\ref{tab:original} and Tab.~\ref{tab:diverse}, it becomes evident that many models exhibit consistent performance patterns across most dimensions. This suggests a general robustness to instructional variations among these models.
However, a noticeable exception is observed in the dimension of object removal. Here, specific models, such as InstructPix2Pix\cite{brooks2023instructpix2pix}, HIVE~\cite{zhang2023hive}, InstructionDiffusion~\cite{geng2023instructdiffusion}, and MagicBrush~\cite{zhang2024magicbrush}, show marked performance declines when shifting from original to diverse instructions. This indicates a sensitivity to changes in instructional phrasing or complexity for these models.
Conversely, other models maintain stable performance across both instruction types. A critical factor contributing to this stability appears to be their use of advanced techniques like LLM~\cite{achiam2023gpt,touvron2023llama} or MLLM~\cite{liu2023llava,liu2023improvedllava} to process and understand instructions. These models demonstrate an enhanced capacity to handle variations in instructions, thereby increasing their overall robustness and adaptability.
\textcolor{black}{%
For example, Qwen-image~\cite{wu2025qwen} preserves its leading performance in most high-level and multi-round dimensions when transitioning from original to diverse instructions, and even improves in several low-level tasks such as shadow removal.%
}
\textcolor{black}{%
Similarly, FLUX Kontext~\cite{batifol2025flux} shows stable or improved results in watermark removal and multi-round color alteration under diverse instructions, indicating that models equipped with stronger language understanding can better generalize to paraphrased or more complex instruction forms.%
}

\subsubsection{\textcolor{black}{Cross-Model Judge Study with Qwen3VL-8B}}

\textcolor{black}{ To investigate whether our benchmark is biased toward GPT-style linguistic patterns or a specific GPT-family evaluator, we conduct an additional ablation study in which we replace GPT-4V with Qwen3VL-8B~\cite{bai2025qwen3} as the judge model. Qwen3VL-8B is a strong open-source multimodal LLM that is architecturally and parametrically distinct from GPT-4V, making it a suitable alternative evaluator for assessing the robustness of our evaluation protocol. In this study, we keep all components of the evaluation pipeline fixed except for the judge model: we use exactly the same image sets, the same original and diverse instructions, and the same scoring prompts and rubric as in the GPT-4V setting, and Qwen3VL-8B is only asked to output scores in the same format as GPT-4V, without any task-specific adaptation. We re-evaluate all editing models over low-level, high-level, and multi-round editing, and report the detailed per-dimension results under original and diverse instructions in Tab.~\ref{tab:original_qwen3vl} and Tab.~\ref{tab:diverse_qwen3vl}, respectively. The results show that Qwen3VL-8B produces highly consistent rankings and relative performance trends compared to GPT-4V: across both instruction settings, Qwen-image~\cite{wu2025qwen} and FLUX Kontext~\cite{batifol2025flux} remain the strongest models on most high-level and multi-round dimensions (especially counting, direction perception, object-centric edits, and color alteration), while methods such as InstructDiffusion~\cite{geng2023instructdiffusion} and MagicBrush~\cite{zhang2024magicbrush} still perform competitively on several low-level dimensions, and InstructPix2Pix~\cite{brooks2023instructpix2pix} and InstructEdit~\cite{wang2023instructedit} consistently lag behind on more challenging semantic and multi-round tasks. The dimension-level difficulty patterns are also preserved: counting and object removal remain among the hardest dimensions across models, whereas background replacement and color alteration are relatively easier; moreover, multi-round editing still shows a clear performance drop compared to single-round high-level editing, confirming that maintaining consistency over multiple sequential edits is intrinsically challenging regardless of the underlying judge. Overall, the strong agreement between GPT-4V and Qwen3VL-8B in terms of model ranking, relative gaps, and dimension-wise trends suggests that our evaluation framework is not overly biased toward GPT-family linguistic patterns or a specific GPT-based evaluator, and that the benchmark conclusions are robust across heterogeneous multimodal judges. }

\subsubsection{\textcolor{black}{Cross-Model Judge Study with LLaVA-1.5-7B}}

\textcolor{black}{
We further replace GPT-4V with LLaVA-1.5-7B~\cite{liu2023improvedllava} as the judge model while keeping the same image sets, original/diverse instructions, and the identical scoring prompts and rubric. The per-dimension results under original and diverse instructions are reported in Tab.~\ref{tab:original_llava} and Tab.~\ref{tab:diverse_llava}. Overall, LLaVA-1.5-7B yields similar ranking patterns and dimension-wise trends to GPT-4V and Qwen3VL-8B: Qwen-image~\cite{wu2025qwen} and FLUX Kontext~\cite{batifol2025flux} remain among the strongest models on most high-level and multi-round dimensions, and multi-round editing consistently shows a noticeable performance drop compared to single-round high-level editing. These results further support that our benchmark conclusions are robust to the choice of heterogeneous multimodal judges.
}

\begin{table*}[!t]
\caption{
\textcolor{black}{
Average rank summary across judge models.
Ranks are computed per-dimension (higher score $\Rightarrow$ better rank; ties use average rank), then averaged over all valid dimensions.}
}
\setlength{\tabcolsep}{6pt}
\renewcommand{\arraystretch}{1.05}
\centering
\resizebox{\textwidth}{!}{%
\begin{tabular}{l|c|c|c|c|c|c}
\midrule
\textcolor{black}{\textbf{Model}} &
\textcolor{black}{\textbf{Overall Rank}} &
\textcolor{black}{\textbf{GPT Judge}} &
\textcolor{black}{\textbf{Qwen3VL-8B Judge}} &
\textcolor{black}{\textbf{LLaVA-1.5-7B Judge}} &
\textcolor{black}{\textbf{Across-judge Avg Rank}} &
\textcolor{black}{\textbf{Max Diff}} \\
\midrule
\textcolor{black}{\textbf{Qwen-image~\cite{wu2025qwen}}} & \textcolor{black}{1} & \textcolor{black}{2.30} & \textcolor{black}{2.35} & \textcolor{black}{2.43} & \textcolor{black}{2.36} & \textcolor{black}{0.13} \\
\textcolor{black}{\textbf{FLUX Kontext~\cite{batifol2025flux}}} & \textcolor{black}{2} & \textcolor{black}{3.76} & \textcolor{black}{4.00} & \textcolor{black}{3.83} & \textcolor{black}{3.86} & \textcolor{black}{0.24} \\
\textcolor{black}{\textbf{MagicBrush~\cite{zhang2024magicbrush}}} & \textcolor{black}{3} & \textcolor{black}{4.09} & \textcolor{black}{4.48} & \textcolor{black}{4.46} & \textcolor{black}{4.34} & \textcolor{black}{0.39} \\
\textcolor{black}{\textbf{InstructDiffusion~\cite{geng2023instructdiffusion}}} & \textcolor{black}{4} & \textcolor{black}{5.61} & \textcolor{black}{5.37} & \textcolor{black}{4.85} & \textcolor{black}{5.28} & \textcolor{black}{0.76} \\
\textcolor{black}{\textbf{HIVE~\cite{zhang2023hive}}} & \textcolor{black}{5} & \textcolor{black}{5.15} & \textcolor{black}{5.35} & \textcolor{black}{5.39} & \textcolor{black}{5.30} & \textcolor{black}{0.24} \\
\textcolor{black}{\textbf{MGIE~\cite{fu2023guiding}}} & \textcolor{black}{6} & \textcolor{black}{5.37} & \textcolor{black}{5.17} & \textcolor{black}{5.54} & \textcolor{black}{5.36} & \textcolor{black}{0.37} \\
\textcolor{black}{\textbf{InstructPix2Pix~\cite{brooks2023instructpix2pix}}} & \textcolor{black}{7} & \textcolor{black}{6.43} & \textcolor{black}{6.41} & \textcolor{black}{6.72} & \textcolor{black}{6.52} & \textcolor{black}{0.30} \\
\textcolor{black}{\textbf{InstructAny2Pix~\cite{li2023instructany2pix}}} & \textcolor{black}{8} & \textcolor{black}{6.85} & \textcolor{black}{6.76} & \textcolor{black}{6.41} & \textcolor{black}{6.67} & \textcolor{black}{0.43} \\
\textcolor{black}{\textbf{HQ-Edit~\cite{hui2024hq}}} & \textcolor{black}{9} & \textcolor{black}{7.33} & \textcolor{black}{7.13} & \textcolor{black}{7.70} & \textcolor{black}{7.38} & \textcolor{black}{0.57} \\
\textcolor{black}{\textbf{InstructEdit~\cite{wang2023instructedit}}} & \textcolor{black}{10} & \textcolor{black}{8.11} & \textcolor{black}{7.98} & \textcolor{black}{7.67} & \textcolor{black}{7.92} & \textcolor{black}{0.43} \\
\midrule
\end{tabular}
}
\label{tab:avg_rank_summary}
\end{table*}

\begin{figure*}[!t]
  \centering
  \includegraphics[width=2.0\columnwidth]{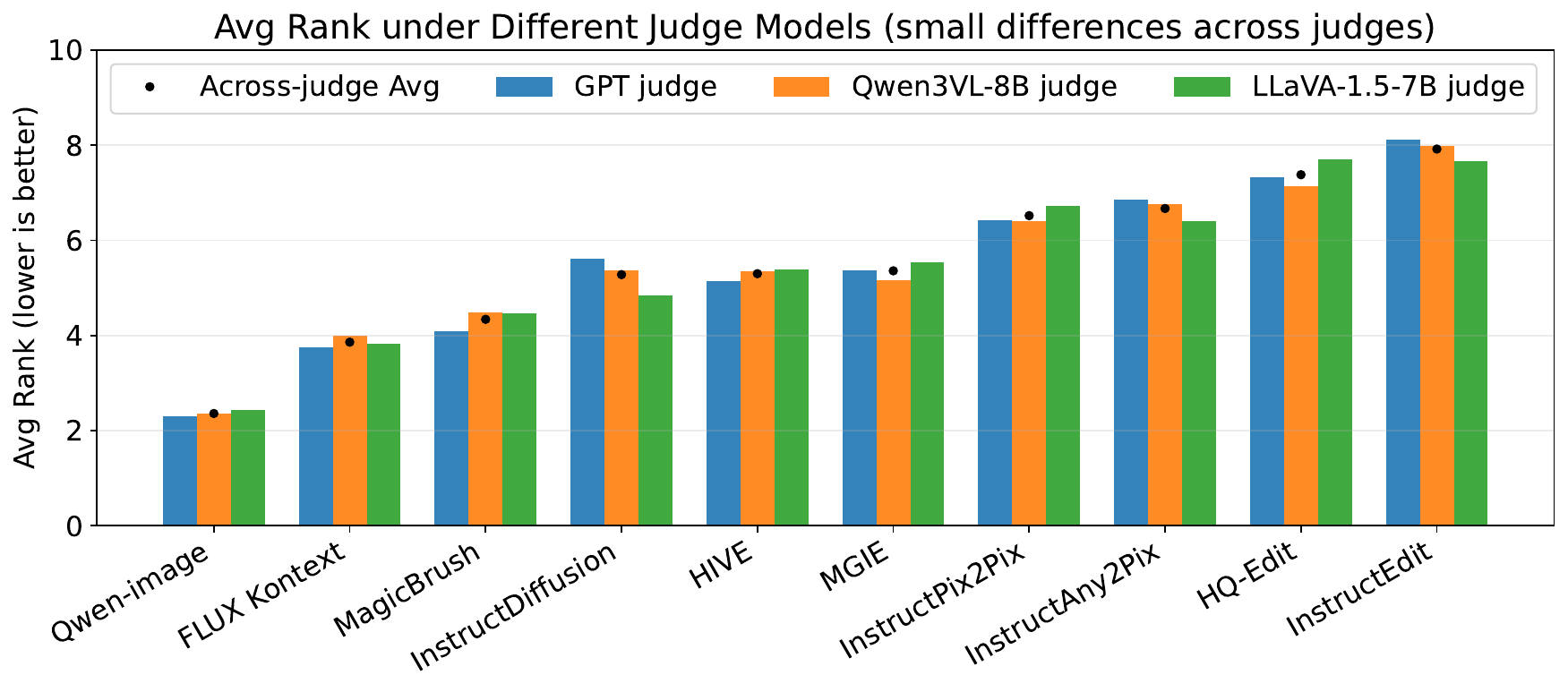}
  \caption{
    \textcolor{black}{
    Avg Rank comparison across different judge models (GPT, Qwen3VL-8B, and LLaVA-1.5-7B). 
    The small gaps between bars indicate that the evaluation results are consistent across judges.
    }
  }
  \label{fig:avg_rank_barplot}
\end{figure*}

\subsubsection{\textcolor{black}{Judge Consistency Analysis}}
\label{sec:judge_consistency}
\textcolor{black}{
To validate that our conclusions are robust to the choice of judge model, we evaluate all methods with three different judges (GPT, Qwen3VL-8B~\cite{bai2025qwen3}, and LLaVA-1.5-7B~\cite{liu2023improvedllava}). For each judge, we first compute per-dimension rankings across methods (higher score indicates better performance) and then average the ranks over all valid dimensions to obtain an Avg Rank for each method under that judge. We further summarize cross-judge consistency using two metrics: (i) \emph{Across-judge Avg Rank}, i.e., the mean Avg Rank over the three judges, and (ii) \emph{Max Diff}, i.e., the maximum difference among the three Avg Ranks for the same method. Tab.~\ref{tab:avg_rank_summary} reports the Avg Rank under each judge and the aggregated statistics, showing that the overall ordering is stable across judges and that rank variations are small (most methods have low Max Diff). Fig.~\ref{fig:avg_rank_barplot} provides a complementary visualization: for each method, the bars corresponding to different judges are closely aligned, indicating only minor changes in Avg Rank when switching the judge model. Overall, these results demonstrate that our benchmark yields consistent evaluations across different judge models.
}

\subsubsection{Evaluation Across Single- and Multi-round Editing}
In this study, we extended our evaluation in \benchname\ by incorporating multi-round editing dimensions, expanding upon the single-round evaluation presented in the conference version.
As shown in Tab.~\ref{tab:original}, a noteworthy decrease in performance is observed when transitioning from single-round to multi-round editing, specifically in tasks such as counting, direction perception, object removal, object replacement, and color alteration. This may be because the outcome of a prior editing round heavily influences subsequent rounds. For instance, if the instruction in the initial round is ``Replace the teddy bear with a puppy," followed by a subsequent instruction to ``Replace the puppy with a kitten," the accuracy of the first edit crucially affects the success of the second. Failure in the first round (\emph{i.e.,} not correctly replacing the teddy bear with a puppy) complicates achieving the desired outcome in the subsequent round.
In contrast, the reduction in performance for global editing tasks, such as background replacement and style alteration, is relatively minor. These tasks generally show independence between rounds; thus, the outcome of one round has little effect on the next. For instance, if one instruction is to ``change the background to a snow mountain" and the subsequent command is to ``change the background to a volcano," the success of the previous round has minimal impact on the effectiveness of the following round.
Notably, in the Style Alteration dimension, we utilized the CLIP score for evaluation. As a result, multiple rounds of editing could potentially yield superior outcomes compared to a single round.

\subsection{Aesthetic Quality Assessment.}

Aesthetic quality is a crucial criterion in image editing, and our study aims to evaluate this aspect systematically using the Aesthetic Predictor’s Score (AP). This approach is similar to that employed by InstructDiffusion~\cite{geng2023instructdiffusion}, leveraging a methodology parallel to LAION-5B~\cite{schuhmann2022laion}, which uses the CLIP+MLP Aesthetic Score Predictor. A higher AP score reflects superior perceptual quality.
In our analysis, we calculated the AP scores for images edited by different methods, averaging these scores across each evaluation dimension.
The results are summarized in Tab.~\ref{tab:original_aes} and Tab.~\ref{tab:diverse_aes}, leading to the following insights:
\begin{itemize}
    \item Across most models and evaluation dimensions, images edited using original instructions maintain comparable AP scores to those edited with diverse instructions. This consistency suggests a robustness in perceptual quality, regardless of the variability in instruction types.
    
    \item AP scores exhibit significant variation across different dimensions. Notably, whether employing original or diverse instructions, the scores for the snow removal dimension consistently surpass those of the deblurring dimension, highlighting the disparity in how these editing tasks affect aesthetic quality.
    
    \item In multi-round editing scenarios, most models and dimensions show a decrease in AP scores relative to single-round editing. This trend indicates that the cumulative nature of edits can lead to a degradation of aesthetic quality, highlighting the challenges involved in sustaining visual integrity through multiple iterations.
    
\end{itemize}

\begin{table*}[!t]
\caption{
\textcolor{black}{
Evaluation Results for Aesthetic Quality Assessment in \benchname\ Across Different Dimensions Using Diverse Instructions. 
Abbreviations: Debl. = Deblurring; Haze Rem. = Haze Removal; Lowl. Enh. = Lowlight Enhancement; Noise Rem. = Noise Removal; Rain Rem. = Rain Removal; Shadow Rem. = Shadow Removal; Snow Rem. = Snow Removal; Wtrmk Rem. = Watermark Removal; Cnt. = Counting; Dir. Perc. = Direction Perception; Obj. Rem. = Object Removal; Obj. Repl. = Object Replacement; Bkg. Repl. = Background Replacement; Col. Alt. = Color Alteration; Sty. Alt. = Style Alteration; Reg. Acc. = Region Accuracy.
}
}
\setlength{\tabcolsep}{4pt}
\centering
\resizebox{\textwidth}{!}{%
\begin{tabular}{l|c|c|c|c|c|c|c|c}

\midrule

\multicolumn{9}{c}{\textcolor{black}{\textbf{Low-level Editing}}} \\

\midrule

\textcolor{black}{\textbf{Model}}           
& \textcolor{black}{\textbf{Debl.}}
& \textcolor{black}{\textbf{Haze Rem.}}
& \textcolor{black}{\textbf{Lowl. Enh.}}
& \textcolor{black}{\textbf{Noise Rem.}}
& \textcolor{black}{\textbf{Rain Rem.}}
& \textcolor{black}{\textbf{Shadow Rem.}}
& \textcolor{black}{\textbf{Snow Rem.}}
& \textcolor{black}{\textbf{Wtrmk Rem.}} \\

\midrule

\textcolor{black}{\textbf{HIVE~\cite{zhang2023hive}}}                    & \textcolor{black}{4.17} & \textcolor{black}{4.71} & \textcolor{black}{4.45} & \textcolor{black}{5.02} & \textcolor{black}{4.98} & \textcolor{black}{4.18} & \textcolor{black}{6.02} & \textcolor{black}{4.51} \\
\textcolor{black}{\textbf{InstructDiffusion~\cite{geng2023instructdiffusion}}}       & \textcolor{black}{3.65} & \textcolor{black}{4.68} & \textcolor{black}{4.45} & \textcolor{black}{5.11} & \textcolor{black}{4.63} & \textcolor{black}{4.01} & \textcolor{black}{5.69} & \textcolor{black}{4.97} \\
\textcolor{black}{\textbf{InstructPix2Pix~\cite{brooks2023instructpix2pix}}}         & \textcolor{black}{3.82} & \textcolor{black}{4.49} & \textcolor{black}{4.23} & \textcolor{black}{4.19} & \textcolor{black}{4.88} & \textcolor{black}{3.67} & \textcolor{black}{6.23} & \textcolor{black}{4.29} \\
\textcolor{black}{\textbf{MagicBrush~\cite{zhang2024magicbrush}}}                  & \textcolor{black}{3.87} & \textcolor{black}{4.59} & \textcolor{black}{4.41} & \textcolor{black}{4.07} & \textcolor{black}{4.82} & \textcolor{black}{3.63} & \textcolor{black}{6.07} & \textcolor{black}{4.41} \\
\textcolor{black}{\textbf{MGIE~\cite{fu2023guiding}}}                    & \textcolor{black}{3.91} & \textcolor{black}{4.63} & \textcolor{black}{4.57} & \textcolor{black}{4.80} & \textcolor{black}{4.74} & \textcolor{black}{4.19} & \textcolor{black}{6.04} & \textcolor{black}{4.77} \\
\textcolor{black}{\textbf{InstructEdit~\cite{wang2023instructedit}}}                & \textcolor{black}{3.79} & \textcolor{black}{4.57} & \textcolor{black}{3.82} & \textcolor{black}{4.77} & \textcolor{black}{4.87} & \textcolor{black}{4.06} & \textcolor{black}{6.46} & \textcolor{black}{4.77} \\
\textcolor{black}{\textbf{InstructAny2Pix~\cite{li2023instructany2pix}}}             & \textcolor{black}{5.79} & \textcolor{black}{5.85} & \textcolor{black}{5.74} & \textcolor{black}{6.26} & \textcolor{black}{6.04} & \textcolor{black}{5.60} & \textcolor{black}{6.46} & \textcolor{black}{5.88} \\
\textcolor{black}{\textbf{HQ-Edit~\cite{hui2024hq}}}                 & \textcolor{black}{4.98} & \textcolor{black}{5.08} & \textcolor{black}{5.55} & \textcolor{black}{5.81} & \textcolor{black}{5.31} & \textcolor{black}{4.31} & \textcolor{black}{6.04} & \textcolor{black}{5.02} \\

\midrule

\multicolumn{9}{c}{\textcolor{black}{\textbf{High-level Editing}}} \\

\midrule

\textcolor{black}{\textbf{Model}}
& \textcolor{black}{\textbf{Cnt.}}
& \textcolor{black}{\textbf{Dir. Perc.}}
& \textcolor{black}{\textbf{Obj. Rem.}}
& \textcolor{black}{\textbf{Obj. Repl.}}
& \textcolor{black}{\textbf{Bkg. Repl.}}
& \textcolor{black}{\textbf{Col. Alt.}}
& \textcolor{black}{\textbf{Sty. Alt.}}
& \textcolor{black}{\textbf{Reg. Acc.}} \\

\midrule

\textcolor{black}{\textbf{HIVE~\cite{zhang2023hive}}}              & \textcolor{black}{5.85} & \textcolor{black}{5.72} & \textcolor{black}{5.76} & \textcolor{black}{5.70} & \textcolor{black}{5.78} & \textcolor{black}{5.94} & \textcolor{black}{5.60} & \textcolor{black}{5.99} \\
\textcolor{black}{\textbf{InstructDiffusion~\cite{geng2023instructdiffusion}}}       & \textcolor{black}{5.65} & \textcolor{black}{5.57} & \textcolor{black}{5.52} & \textcolor{black}{5.37} & \textcolor{black}{5.73} & \textcolor{black}{5.81} & \textcolor{black}{5.65} & \textcolor{black}{5.71} \\
\textcolor{black}{\textbf{InstructPix2Pix~\cite{brooks2023instructpix2pix}}}         & \textcolor{black}{5.68} & \textcolor{black}{5.72} & \textcolor{black}{5.75} & \textcolor{black}{5.72} & \textcolor{black}{5.71} & \textcolor{black}{5.92} & \textcolor{black}{5.53} & \textcolor{black}{5.87} \\
\textcolor{black}{\textbf{MagicBrush~\cite{zhang2024magicbrush}}}                  & \textcolor{black}{5.76} & \textcolor{black}{5.72} & \textcolor{black}{5.68} & \textcolor{black}{5.63} & \textcolor{black}{6.09} & \textcolor{black}{5.93} & \textcolor{black}{5.77} & \textcolor{black}{5.83} \\
\textcolor{black}{\textbf{MGIE~\cite{fu2023guiding}}}                    & \textcolor{black}{5.76} & \textcolor{black}{5.62} & \textcolor{black}{5.58} & \textcolor{black}{5.48} & \textcolor{black}{5.73} & \textcolor{black}{5.85} & \textcolor{black}{5.73} & \textcolor{black}{5.79} \\
\textcolor{black}{\textbf{InstructEdit~\cite{wang2023instructedit}}}                & \textcolor{black}{5.73} & \textcolor{black}{5.63} & \textcolor{black}{5.93} & \textcolor{black}{5.74} & \textcolor{black}{5.68} & \textcolor{black}{5.94} & \textcolor{black}{5.62} & \textcolor{black}{5.85} \\
\textcolor{black}{\textbf{InstructAny2Pix~\cite{li2023instructany2pix}}}             & \textcolor{black}{6.29} & \textcolor{black}{6.09} & \textcolor{black}{6.35} & \textcolor{black}{6.19} & \textcolor{black}{6.35} & \textcolor{black}{6.40} & \textcolor{black}{6.48} & \textcolor{black}{6.41} \\
\textcolor{black}{\textbf{HQ-Edit~\cite{hui2024hq}}}                 & \textcolor{black}{5.81} & \textcolor{black}{5.91} & \textcolor{black}{5.83} & \textcolor{black}{5.98} & \textcolor{black}{6.06} & \textcolor{black}{6.02} & \textcolor{black}{5.66} & \textcolor{black}{6.04} \\

\midrule

\multicolumn{9}{c}{\textcolor{black}{\textbf{Multi-round Editing}}} \\

\midrule

\textcolor{black}{\textbf{Model}}
& \textcolor{black}{\textbf{Cnt.}}
& \textcolor{black}{\textbf{Dir. Perc.}}
& \textcolor{black}{\textbf{Obj. Rem.}}
& \textcolor{black}{\textbf{Obj. Repl.}}
& \textcolor{black}{\textbf{Bkg. Repl.}}
& \textcolor{black}{\textbf{Col. Alt.}}
& \textcolor{black}{\textbf{Sty. Alt.}}
& \textcolor{black}{\textbf{Reg. Acc.}} \\

\midrule

\textcolor{black}{\textbf{HIVE~\cite{zhang2023hive}}}  & \textcolor{black}{5.18} & \textcolor{black}{5.12} & \textcolor{black}{4.91} & \textcolor{black}{4.97} & \textcolor{black}{4.83} & \textcolor{black}{5.43} & \textcolor{black}{4.78} & \textcolor{black}{-} \\
\textcolor{black}{\textbf{InstructDiffusion~\cite{geng2023instructdiffusion}}}  & \textcolor{black}{5.11} & \textcolor{black}{5.01} & \textcolor{black}{4.76} & \textcolor{black}{4.88} & \textcolor{black}{5.14} & \textcolor{black}{5.46} & \textcolor{black}{4.93} & \textcolor{black}{-} \\
\textcolor{black}{\textbf{InstructPix2Pix~\cite{brooks2023instructpix2pix}}}  & \textcolor{black}{5.29} & \textcolor{black}{5.36} & \textcolor{black}{5.04} & \textcolor{black}{5.27} & \textcolor{black}{4.80} & \textcolor{black}{5.49} & \textcolor{black}{4.93} & \textcolor{black}{-} \\
\textcolor{black}{\textbf{MagicBrush~\cite{zhang2024magicbrush}}}  & \textcolor{black}{5.44} & \textcolor{black}{5.39} & \textcolor{black}{5.16} & \textcolor{black}{5.19} & \textcolor{black}{5.55} & \textcolor{black}{5.69} & \textcolor{black}{5.41} & \textcolor{black}{-} \\
\textcolor{black}{\textbf{MGIE~\cite{fu2023guiding}}}  & \textcolor{black}{5.33} & \textcolor{black}{5.26} & \textcolor{black}{4.94} & \textcolor{black}{4.94} & \textcolor{black}{4.81} & \textcolor{black}{5.53} & \textcolor{black}{4.90} & \textcolor{black}{-} \\
\textcolor{black}{\textbf{InstructEdit~\cite{wang2023instructedit}}}  & \textcolor{black}{5.37} & \textcolor{black}{5.41} & \textcolor{black}{5.71} & \textcolor{black}{5.28} & \textcolor{black}{5.41} & \textcolor{black}{5.83} & \textcolor{black}{5.07} & \textcolor{black}{-} \\
\textcolor{black}{\textbf{InstructAny2Pix~\cite{li2023instructany2pix}}}  & \textcolor{black}{6.40} & \textcolor{black}{6.27} & \textcolor{black}{6.46} & \textcolor{black}{6.41} & \textcolor{black}{6.56} & \textcolor{black}{6.57} & \textcolor{black}{6.68} & \textcolor{black}{-} \\
\textcolor{black}{\textbf{HQ-Edit~\cite{hui2024hq}}}  & \textcolor{black}{5.55} & \textcolor{black}{5.67} & \textcolor{black}{5.62} & \textcolor{black}{5.92} & \textcolor{black}{5.81} & \textcolor{black}{5.94} & \textcolor{black}{5.53} & \textcolor{black}{-} \\

\midrule

\end{tabular}
}
\label{tab:diverse_aes}
\end{table*}

\subsection{Evaluation Across Different Categories}

To examine the effect of various instruction categories—such as Animal, Object, Scenery, Plant, Human, and Global—on the outcomes, we conducted a statistical analysis of the scores for each instruction type.
Based on the data presented in Tab.~\ref{tab:category}, we observed the following:
\begin{itemize}
    \item The ``Scenery" and ``Global" categories consistently exhibit superior performance across all evaluated IIE models compared to other categories. This can be explained by their inherent tendency towards global editing, which reduces the need for accurate localization of specific target objects.
    
    \item Within the same evaluation metrics, the scores for multi-round editing are generally lower than those for single-round editing. This may result from the impact that the results of previous editing rounds exert on subsequent rounds, leading to a decrease in scores when multiple rounds are involved.
    
    \item Within the same evaluation metrics, switching from using original instructions to diverse instructions generally results in a performance decline for most models, evident in both single-round and multi-round editing scenarios.
\end{itemize}

\begin{table*}[!t]
\caption{
\textcolor{black}{
Comparison of Performance across Various Categories in Single-Round and Multi-Round Image Editing. Abbreviations: Ani. = Animal; Obj. = Object; Sce. = Scenery; Plt. = Plant; Hum. = Human; Glo. = Global.
}
}
\setlength{\tabcolsep}{4pt}
\centering
\resizebox{\textwidth}{!}{%
\begin{tabular}{l|cccccc|cccccc}
\toprule
\multicolumn{13}{c}{\textcolor{black}{\textbf{Original Instruction}}} \\ \midrule
\textcolor{black}{} & \multicolumn{6}{c|}{\textcolor{black}{\textbf{Single-round Editing}}} & \multicolumn{6}{c}{\textcolor{black}{\textbf{Multi-round Editing}}} \\ \midrule
\textcolor{black}{\textbf{Model}} & \textcolor{black}{\textbf{Ani.}} & \textcolor{black}{\textbf{Obj.}} & \textcolor{black}{\textbf{Sce.}} & \textcolor{black}{\textbf{Plt.}} & \textcolor{black}{\textbf{Hum.}} & \textcolor{black}{\textbf{Glo.}} & \textcolor{black}{\textbf{Ani.}} & \textcolor{black}{\textbf{Obj.}} & \textcolor{black}{\textbf{Sce.}} & \textcolor{black}{\textbf{Plt.}} & \textcolor{black}{\textbf{Hum.}} & \textcolor{black}{\textbf{Glo.}} \\ \midrule
\textcolor{black}{\textbf{HIVE~\cite{zhang2023hive}}}              & \textcolor{black}{32.72} & \textcolor{black}{36.93} & \textcolor{black}{42.53} & \textcolor{black}{35.74} & \textcolor{black}{30.55} & \textcolor{black}{38.69} & \textcolor{black}{16.38} & \textcolor{black}{17.57} & \textcolor{black}{20.52} & \textcolor{black}{12.88} & \textcolor{black}{17.00} & \textcolor{black}{23.65} \\
\textcolor{black}{\textbf{InstructDiffusion~\cite{geng2023instructdiffusion}}} & \textcolor{black}{33.86} & \textcolor{black}{37.59} & \textcolor{black}{40.91} & \textcolor{black}{38.70} & \textcolor{black}{33.59} & \textcolor{black}{38.82} & \textcolor{black}{13.94} & \textcolor{black}{14.11} & \textcolor{black}{23.81} & \textcolor{black}{10.43} & \textcolor{black}{11.04} & \textcolor{black}{21.03} \\
\textcolor{black}{\textbf{InstructPix2Pix~\cite{brooks2023instructpix2pix}}}   & \textcolor{black}{30.99} & \textcolor{black}{33.56} & \textcolor{black}{39.87} & \textcolor{black}{32.59} & \textcolor{black}{26.25} & \textcolor{black}{38.42} & \textcolor{black}{13.65} & \textcolor{black}{14.80} & \textcolor{black}{23.12} & \textcolor{black}{10.02} & \textcolor{black}{8.17} & \textcolor{black}{22.71} \\
\textcolor{black}{\textbf{MagicBrush~\cite{zhang2024magicbrush}}}          & \textcolor{black}{34.53} & \textcolor{black}{39.94} & \textcolor{black}{43.82} & \textcolor{black}{39.90} & \textcolor{black}{32.37} & \textcolor{black}{38.15} & \textcolor{black}{16.67} & \textcolor{black}{15.49} & \textcolor{black}{25.11} & \textcolor{black}{11.25} & \textcolor{black}{13.47} & \textcolor{black}{21.87} \\
\textcolor{black}{\textbf{MGIE~\cite{fu2023guiding}}}              & \textcolor{black}{31.82} & \textcolor{black}{39.11} & \textcolor{black}{45.48} & \textcolor{black}{40.22} & \textcolor{black}{33.08} & \textcolor{black}{37.62} & \textcolor{black}{14.37} & \textcolor{black}{13.55} & \textcolor{black}{19.83} & \textcolor{black}{11.25} & \textcolor{black}{12.14} & \textcolor{black}{22.81} \\
\textcolor{black}{\textbf{InstructEdit~\cite{wang2023instructedit}}}      & \textcolor{black}{24.83} & \textcolor{black}{32.27} & \textcolor{black}{40.18} & \textcolor{black}{31.54} & \textcolor{black}{29.05} & \textcolor{black}{33.97} & \textcolor{black}{7.90} & \textcolor{black}{5.81} & \textcolor{black}{22.86} & \textcolor{black}{3.68} & \textcolor{black}{6.84} & \textcolor{black}{18.94} \\
\textcolor{black}{\textbf{InstructAny2Pix~\cite{li2023instructany2pix}}} & \textcolor{black}{28.17} & \textcolor{black}{29.34} & \textcolor{black}{38.45} & \textcolor{black}{25.62} & \textcolor{black}{27.43} & \textcolor{black}{38.06} & \textcolor{black}{12.93} & \textcolor{black}{17.43} & \textcolor{black}{24.94} & \textcolor{black}{11.86} & \textcolor{black}{17.00} & \textcolor{black}{25.80} \\
\textcolor{black}{\textbf{HQ-Edit~\cite{hui2024hq}}}           & \textcolor{black}{26.74} & \textcolor{black}{30.53} & \textcolor{black}{39.98} & \textcolor{black}{24.91} & \textcolor{black}{24.66} & \textcolor{black}{26.91} & \textcolor{black}{11.49} & \textcolor{black}{12.45} & \textcolor{black}{23.55} & \textcolor{black}{6.95} & \textcolor{black}{11.04} & \textcolor{black}{23.88} \\ \midrule
\multicolumn{13}{c}{\textcolor{black}{\textbf{Diverse Instruction}}} \\ \midrule
\textcolor{black}{} & \multicolumn{6}{c|}{\textcolor{black}{\textbf{Single-round Editing}}} & \multicolumn{6}{c}{\textcolor{black}{\textbf{Multi-round Editing}}} \\ \midrule
\textcolor{black}{\textbf{Model}} & \textcolor{black}{\textbf{Ani.}} & \textcolor{black}{\textbf{Obj.}} & \textcolor{black}{\textbf{Sce.}} & \textcolor{black}{\textbf{Plt.}} & \textcolor{black}{\textbf{Hum.}} & \textcolor{black}{\textbf{Glo.}} & \textcolor{black}{\textbf{Ani.}} & \textcolor{black}{\textbf{Obj.}} & \textcolor{black}{\textbf{Sce.}} & \textcolor{black}{\textbf{Plt.}} & \textcolor{black}{\textbf{Hum.}} & \textcolor{black}{\textbf{Glo.}} \\ \midrule
\textcolor{black}{\textbf{HIVE~\cite{zhang2023hive}}}              & \textcolor{black}{28.50} & \textcolor{black}{34.31} & \textcolor{black}{45.31} & \textcolor{black}{34.52} & \textcolor{black}{29.06} & \textcolor{black}{37.63} & \textcolor{black}{13.22} & \textcolor{black}{11.34} & \textcolor{black}{19.22} & \textcolor{black}{10.63} & \textcolor{black}{10.60} & \textcolor{black}{21.76} \\
\textcolor{black}{\textbf{InstructDiffusion~\cite{geng2023instructdiffusion}}} & \textcolor{black}{25.04} & \textcolor{black}{32.44} & \textcolor{black}{40.54} & \textcolor{black}{31.98} & \textcolor{black}{29.10} & \textcolor{black}{34.11} & \textcolor{black}{11.78} & \textcolor{black}{10.51} & \textcolor{black}{23.29} & \textcolor{black}{8.38} & \textcolor{black}{9.93} & \textcolor{black}{19.79} \\
\textcolor{black}{\textbf{InstructPix2Pix~\cite{brooks2023instructpix2pix}}}   & \textcolor{black}{27.44} & \textcolor{black}{35.67} & \textcolor{black}{39.58} & \textcolor{black}{33.65} & \textcolor{black}{25.74} & \textcolor{black}{36.87} & \textcolor{black}{12.04} & \textcolor{black}{12.31} & \textcolor{black}{22.60} & \textcolor{black}{8.59} & \textcolor{black}{9.27} & \textcolor{black}{22.05} \\
\textcolor{black}{\textbf{MagicBrush~\cite{zhang2024magicbrush}}}          & \textcolor{black}{32.82} & \textcolor{black}{37.49} & \textcolor{black}{45.60} & \textcolor{black}{35.95} & \textcolor{black}{32.26} & \textcolor{black}{37.39} & \textcolor{black}{12.79} & \textcolor{black}{11.07} & \textcolor{black}{24.85} & \textcolor{black}{8.38} & \textcolor{black}{13.91} & \textcolor{black}{22.23} \\
\textcolor{black}{\textbf{MGIE~\cite{fu2023guiding}}}              & \textcolor{black}{32.58} & \textcolor{black}{35.77} & \textcolor{black}{44.02} & \textcolor{black}{36.29} & \textcolor{black}{35.22} & \textcolor{black}{37.76} & \textcolor{black}{12.50} & \textcolor{black}{10.24} & \textcolor{black}{21.56} & \textcolor{black}{8.59} & \textcolor{black}{7.95} & \textcolor{black}{21.49} \\
\textcolor{black}{\textbf{InstructEdit~\cite{wang2023instructedit}}}      & \textcolor{black}{32.16} & \textcolor{black}{36.01} & \textcolor{black}{40.45} & \textcolor{black}{35.94} & \textcolor{black}{31.11} & \textcolor{black}{36.15} & \textcolor{black}{8.76} & \textcolor{black}{4.98} & \textcolor{black}{22.25} & \textcolor{black}{3.89} & \textcolor{black}{6.40} & \textcolor{black}{18.86} \\
\textcolor{black}{\textbf{InstructAny2Pix~\cite{li2023instructany2pix}}} & \textcolor{black}{28.24} & \textcolor{black}{31.65} & \textcolor{black}{39.69} & \textcolor{black}{28.57} & \textcolor{black}{27.07} & \textcolor{black}{37.24} & \textcolor{black}{13.07} & \textcolor{black}{15.08} & \textcolor{black}{23.12} & \textcolor{black}{11.66} & \textcolor{black}{15.45} & \textcolor{black}{24.03} \\
\textcolor{black}{\textbf{HQ-Edit~\cite{hui2024hq}}}           & \textcolor{black}{27.33} & \textcolor{black}{30.96} & \textcolor{black}{41.47} & \textcolor{black}{26.07} & \textcolor{black}{23.22} & \textcolor{black}{27.05} & \textcolor{black}{13.65} & \textcolor{black}{11.76} & \textcolor{black}{23.72} & \textcolor{black}{8.38} & \textcolor{black}{11.26} & \textcolor{black}{23.67} \\ 
\bottomrule
\end{tabular}
}
\label{tab:category}
\end{table*}

\subsection{Human Evaluation.}

\begin{figure*}
  \centering
  \includegraphics[width=1.5\columnwidth]{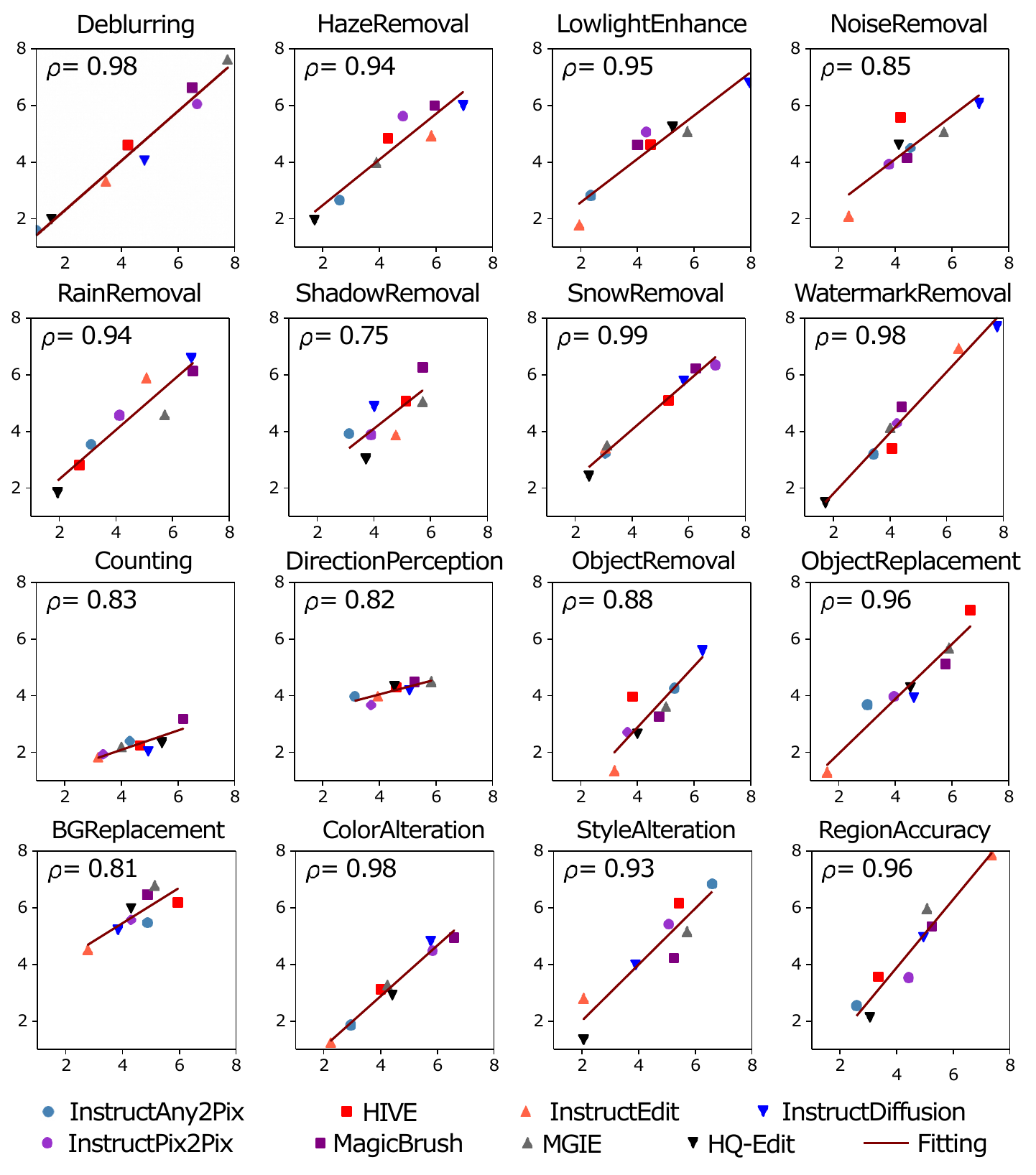}
  \caption{
    Visualization of the correlation between \benchname rank scores (Y-axis) and human evaluation scores (X-axis). A linear fit is applied to illustrate the relationship, and Spearman’s correlation coefficient ($\rho$) is calculated to quantify the correlation strength across each evaluated dimension.
  }
  \label{fig:ori_human}
\end{figure*}

In our study, we undertook a comprehensive evaluation of various models by carefully analyzing their \benchname ratings and computing their \benchname rank scores. This process adhered to the methodology detailed in Section~\ref{sec:human}, ensuring consistency and accuracy in our assessment approach.
Both the \benchname rank scores and the human evaluations utilize a standardized scale ranging from 1 to 8. This uniformity in measurement allows for a direct and meaningful comparison between the model-generated scores and human assessments, providing a clear framework for analysis.
To explore the relationship between these scores, we conducted an in-depth correlation analysis. The outcomes of this analysis are visually represented in Fig.~\ref{fig:ori_human}, offering an intuitive understanding of the data.
Remarkably, our findings reveal strong positive correlations between the \benchname rank scores and human evaluations across every dimension assessed. This consistency across all metrics suggests a high level of agreement between the computational benchmarks and human judgment.
These results furnish robust and compelling evidence that supports the alignment of our proposed benchmark framework with human perception. Such alignment underscores the validity and reliability of our model evaluation method, affirming its effectiveness as a tool for assessing model performance in a manner that resonates with human understanding.

\section{Insights}
\label{sec:insights}
Building on the findings presented in Section~\ref{sec:exp}, it becomes abundantly clear that the current IIE model has some significant limitations that necessitate ongoing enhancement and development. Through rigorous examination and analysis of benchmark experimental results, it is evident that the existing IIE model stands to gain substantial improvements in various dimensions, which could contribute to its overall enhancement and effectiveness.
\begin{itemize}
    
    \item As clearly shown in Tab.~\ref{tab:original} and Tab.~\ref{tab:diverse}, there is a noticeable and substantial degradation in performance across multiple dimensions when the model transitions from single-round editing to multi-round editing. This not only highlights the current model's inadequate robustness in handling successive rounds of edits but also suggests that its current architecture may not be fully optimized for complex, iterative editing tasks. \emph{Future iterations of the model must focus on enhancing its durability and resilience to maintain consistent performance across multiple edits.}
    
    \item The aesthetic quality of images produced by the model emerges as another critical aspect warranting attention from image editors and researchers alike. Presently, most models encounter a significant decline in aesthetic quality scores when moving from single-round editing to multi-round editing. This reflects a gap in the model's ability to sustain aesthetic standards under complex editing conditions. \emph{Hence, it is imperative that future research not only aims to achieve high accuracy in editing operations but also emphasizes the enhancement of aesthetic quality in the resulting images, ensuring that they meet or surpass user expectations and industry standards.}
    
    \item Our comprehensive analysis reveals that no single model dominates across all evaluation dimensions. Instead, various IIE models display a spectrum of strengths and weaknesses, with each one excelling in different facets of editing metrics. This indicates a need to acknowledge and address the inherent limitations within current models, paving the way for the development of a more cohesive IIE model that consistently demonstrates proficient and balanced performance across all dimensions. \emph{Going forward, research efforts should focus on reinforcing the robustness of these models in settings that involve multiple rounds and multiple dimensions of editing, thereby enhancing their practicality and applicability in real-world scenarios.}
    
    \item In specific evaluation metrics, such as object removal, we observe significant performance fluctuations among models including InstructPix2Pix\cite{brooks2023instructpix2pix}, HIVE~\cite{zhang2023hive}, InstructionDiffusion~\cite{geng2023instructdiffusion}, and MagicBrush~\cite{zhang2024magicbrush}. These variations depend largely on the nature of the instructions provided. \emph{Given the unpredictable and diverse nature of user-provided editing directives, there is an urgent need to develop an editing model equipped to handle instructions of variable complexity effectively, ensuring robust and reliable outputs regardless of the input complexity.}
    
    \item As illustrated in Tab.~\ref{tab:category}, significant variations in performance have been documented across different categories. Notably, the ``Scenery" and ``Global" categories consistently achieve superior performance when compared to other categories across all IIE models assessed. This phenomenon suggests an inherent alignment of these categories towards global editing, which inherently lessens the requirement for precise localization of target objects. \emph{These results underscore the necessity for future research to simultaneously take into account diverse editing content and contexts, facilitating a holistic approach to model development and refinement.}
    
\end{itemize}

\section{\textcolor{black}{Discussion}}

\subsection{\textcolor{black}{Discussion about Diversity Dimension}}

\textcolor{black}{
Diversity is widely acknowledged as a critical metric in text-to-image (T2I) generation~\cite{hall2024evalgim}. We also agree that it can be an important property for instruction-based image editing (IIE), especially because lack of diversity may expose systematic biases or mode collapse in edits. For instance, repeatedly inserting a white person when asked to add a person into a scene (e.g., a wealthy house) reflects an undesirable limitation of an editing model even if the edit is technically “consistent” across rounds.
}

\textcolor{black}{
In the current version of \benchname, our core objective is to evaluate controllability and reliability: whether the model performs the requested modification precisely, preserves non-target regions, and remains consistent across multiple rounds. Many instructions in our dataset are intentionally specific and goal-oriented (e.g., ``add a red apple on the shelf'' or ``remove the car''), which reduces the degrees of freedom where diversity would naturally manifest, and makes faithfulness/reliability the most directly measurable axis under a unique target outcome.
}

\textcolor{black}{
That said, we do not view diversity as irrelevant or orthogonal to IIE. Rather, incorporating diversity into \benchname\ is a meaningful and challenging direction for future work. Concretely, future extensions could introduce (i) underspecified or open-ended edit instructions (e.g., ``add a person'' or ``decorate the room'') where multiple valid realizations exist, and (ii) evaluation protocols that quantify both variation and fairness, potentially leveraging established T2I diversity measures~\cite{hall2024evalgim}. We believe this would complement our current focus by enabling a more holistic assessment that jointly considers faithfulness, multiround consistency, and diversity-related behaviors when the application requires it.
}

\subsection{\textcolor{black}{Discussion about Benchmark Size}}

\textcolor{black}{
The design of \benchname reflects a deliberate balance between data coverage, annotation quality, and practical feasibility for large-scale evaluation in instruction-based image editing. With over 2,000 images and 6,700 editing instructions, the benchmark spans 16 single-round and 7 multi-round editing dimensions, and covers a broad range of categories, including animals, objects, scenery, plants, humans, and global scenes. This scope notably exceeds the size and diversity of most publicly available benchmarks in the field, such as TedBench and EditBench, which are often limited in either sample count or dimension coverage.
}

\textcolor{black}{
High-quality human annotation was a central priority in constructing \benchname. Each instruction and category was carefully generated, reviewed, and diversified to ensure accuracy, representativeness, and reproducibility. While further increasing the dataset size might seem beneficial, it frequently introduces annotation inconsistencies or reduces the scientific rigor of the benchmark, potentially impacting the reliability of model evaluation results.
}

\textcolor{black}{
Practical considerations also played an important role in determining dataset scale. Instruction-based editing models typically require several seconds for inference per image, and thorough benchmarking across all dimensions demands significant computational resources and time. Excessive expansion of the dataset could therefore raise barriers to adoption and reproducibility within both academic and industrial communities.
}

\textcolor{black}{
Overall, \benchname is designed to achieve an optimal trade-off between comprehensive coverage and feasible evaluation, facilitating representative and high-quality assessment of instruction-based image editing models. Looking ahead, we will continue to enhance the scale and diversity of the dataset in future iterations, in response to community feedback and technological advancements.
}

\subsection{\textcolor{black}{Discussion about Benchmark Difficulty}}

\textcolor{black}{ The difficulty level of an evaluation benchmark is essential for distinguishing model capabilities and enabling meaningful scientific comparisons. In designing \benchname, we carefully considered the technical maturity of current instruction-based image editing models and the practical obstacles associated with highly complex editing scenarios. At present, most mainstream models still face notable limitations when handling particularly challenging tasks, such as fine-grained adjustments in crowded scenes or nuanced contextual edits. If the benchmark were to focus exclusively on these complex scenarios, it would be difficult for most models to generate robust results, which would undermine the fairness, scientific validity, and interpretability of comparative evaluations. }

\textcolor{black}{ To ensure representative and practical benchmarking, \benchname concentrates on core editing dimensions that are widely supported—such as objects, scenes, colors, and backgrounds—reflecting real-world usage and facilitating clearer differentiation between model performances. Challenging compositional instructions (for example, editing several objects or performing multi-stage modifications in a single step) present significant obstacles for objective and reproducible evaluation. Therefore, for fairness and clarity, each evaluation step in \benchname involves the manipulation of a single object or attribute per instruction. This design choice enables precise and unambiguous assessment of instruction compliance. }

\textcolor{black}{ However, we also recognize that atomic editing steps, while practical for consistent evaluation, may not fully capture the complexity of real-world editing workflows, which often require multi-object or compositional changes. To address this gap, \benchname incorporates multi-round editing tasks—simulating authentic usage by executing consecutive, staged instructions on the same image. This approach markedly elevates the evaluation difficulty and serves as a rigorous test of model robustness, consistency, and the capacity to handle accumulative modifications. In essence, multi-round editing builds a bridge from basic operations to complex editing pipelines, bringing our benchmark closer to reflecting the demands of actual applications. }

\textcolor{black}{ As instruction-based image editing models and annotation resources continue to evolve, we are committed to progressively extending the scope and difficulty of \benchname. A stepwise, scientifically grounded approach ensures that the benchmark remains representative, challenging, and adaptable to ongoing advancements in the field. }

\subsection{\textcolor{black}{Discussion about Instruction Robustness}}

\textcolor{black}{ The robustness of image editing models when confronted with ambiguous, erroneous, or infeasible instructions is a critical factor affecting their real-world applicability. Human inputs in practical scenarios are often imperfect—users may provide unclear intent, malformed syntax, or requests that are impossible to fulfill given the available image content. }

\textcolor{black}{ In the current design of \benchname, we intentionally employ explicit and feasible instructions as our primary evaluation protocol. This approach ensures that ground-truth answers can be defined clearly, results are reproducible, and model performance is fairly and objectively compared using standardized metrics. Introducing ambiguous, incorrect, or infeasible instructions poses significant challenges for scientific benchmarking: it complicates the establishment of annotation standards, the determination of correct outputs, and the formulation of quantitative evaluation metrics. These difficulties could undermine the validity of the benchmark and introduce substantial inconsistencies in comparative analysis. }

\textcolor{black}{ Nonetheless, we recognize the importance of evaluating how models handle ambiguous or confusing user inputs for true robustness. As research in model explainability, annotation methodology, and evaluation protocol progresses, an important future direction is to systematically incorporate scenarios involving imperfect instructions. We plan to explore dedicated modules for robustness and ambiguity analysis in future editions of \benchname, enabling comprehensive assessment of model resilience against human-like input complexity. }

\section{Conclusion}
\label{sec:conclusion}

In this study, we introduced \benchname, a thorough benchmark tailored for instruction-based image editing (IIE), accommodating both single-round and multi-round editing. Our benchmark encompasses a sizable dataset with over 2,000 images and upwards of 6,700 instructions, reflecting 16 unique evaluation metrics. To ascertain the effectiveness of \benchname, we carried out experiments with eight publicly available IIE models. Additionally, we augmented these experiments with detailed human assessments to establish the relationship between \benchname scores and human judgment.
Drawing from the findings obtained through \benchname, we offered crucial insights and suggestions aimed at enhancing IIE models.
We anticipate that the proposed \benchname will become an essential tool, significantly contributing to the development of IIE models and facilitating their performance assessment.

\bibliography{sn-bibliography}% common bib file


\begin{thebibliography}{80}
\ifx \bisbn   \undefined \def \bisbn  #1{ISBN #1}\fi
\ifx \binits  \undefined \def \binits#1{#1}\fi
\ifx \bauthor  \undefined \def \bauthor#1{#1}\fi
\ifx \batitle  \undefined \def \batitle#1{#1}\fi
\ifx \bjtitle  \undefined \def \bjtitle#1{#1}\fi
\ifx \bvolume  \undefined \def \bvolume#1{\textbf{#1}}\fi
\ifx \byear  \undefined \def \byear#1{#1}\fi
\ifx \bissue  \undefined \def \bissue#1{#1}\fi
\ifx \bfpage  \undefined \def \bfpage#1{#1}\fi
\ifx \blpage  \undefined \def \blpage #1{#1}\fi
\ifx \burl  \undefined \def \burl#1{\textsf{#1}}\fi
\ifx \doiurl  \undefined \def \doiurl#1{\url{https://doi.org/#1}}\fi
\ifx \betal  \undefined \def \betal{\textit{et al.}}\fi
\ifx \binstitute  \undefined \def \binstitute#1{#1}\fi
\ifx \binstitutionaled  \undefined \def \binstitutionaled#1{#1}\fi
\ifx \bctitle  \undefined \def \bctitle#1{#1}\fi
\ifx \beditor  \undefined \def \beditor#1{#1}\fi
\ifx \bpublisher  \undefined \def \bpublisher#1{#1}\fi
\ifx \bbtitle  \undefined \def \bbtitle#1{#1}\fi
\ifx \bedition  \undefined \def \bedition#1{#1}\fi
\ifx \bseriesno  \undefined \def \bseriesno#1{#1}\fi
\ifx \blocation  \undefined \def \blocation#1{#1}\fi
\ifx \bsertitle  \undefined \def \bsertitle#1{#1}\fi
\ifx \bsnm \undefined \def \bsnm#1{#1}\fi
\ifx \bsuffix \undefined \def \bsuffix#1{#1}\fi
\ifx \bparticle \undefined \def \bparticle#1{#1}\fi
\ifx \barticle \undefined \def \barticle#1{#1}\fi
\bibcommenthead
\ifx \bconfdate \undefined \def \bconfdate #1{#1}\fi
\ifx \botherref \undefined \def \botherref #1{#1}\fi
\ifx \url \undefined \def \url#1{\textsf{#1}}\fi
\ifx \bchapter \undefined \def \bchapter#1{#1}\fi
\ifx \bbook \undefined \def \bbook#1{#1}\fi
\ifx \bcomment \undefined \def \bcomment#1{#1}\fi
\ifx \oauthor \undefined \def \oauthor#1{#1}\fi
\ifx \citeauthoryear \undefined \def \citeauthoryear#1{#1}\fi
\ifx \endbibitem  \undefined \def \endbibitem {}\fi
\ifx \bconflocation  \undefined \def \bconflocation#1{#1}\fi
\ifx \arxivurl  \undefined \def \arxivurl#1{\textsf{#1}}\fi
\csname PreBibitemsHook\endcsname

%%% 1
\bibitem[\protect\citeauthoryear{Brooks
  et~al.}{2023}]{brooks2023instructpix2pix}
\begin{bchapter}
\bauthor{\bsnm{Brooks}, \binits{T.}},
\bauthor{\bsnm{Holynski}, \binits{A.}},
\bauthor{\bsnm{Efros}, \binits{A.A.}}:
\bctitle{Instructpix2pix: Learning to follow image editing instructions}.
In: \bbtitle{Proceedings of the IEEE/CVF Conference on Computer Vision and
  Pattern Recognition},
pp. \bfpage{18392}--\blpage{18402}
(\byear{2023})
\end{bchapter}
\endbibitem

%%% 2
\bibitem[\protect\citeauthoryear{Geng et~al.}{2023}]{geng2023instructdiffusion}
\begin{botherref}
\oauthor{\bsnm{Geng}, \binits{Z.}},
\oauthor{\bsnm{Yang}, \binits{B.}},
\oauthor{\bsnm{Hang}, \binits{T.}},
\oauthor{\bsnm{Li}, \binits{C.}},
\oauthor{\bsnm{Gu}, \binits{S.}},
\oauthor{\bsnm{Zhang}, \binits{T.}},
\oauthor{\bsnm{Bao}, \binits{J.}},
\oauthor{\bsnm{Zhang}, \binits{Z.}},
\oauthor{\bsnm{Hu}, \binits{H.}},
\oauthor{\bsnm{Chen}, \binits{D.}}, et al.:
Instructdiffusion: A generalist modeling interface for vision tasks.
arXiv preprint arXiv:2309.03895
(2023)
\end{botherref}
\endbibitem

%%% 3
\bibitem[\protect\citeauthoryear{Zhang et~al.}{2024}]{zhang2024magicbrush}
\begin{botherref}
\oauthor{\bsnm{Zhang}, \binits{K.}},
\oauthor{\bsnm{Mo}, \binits{L.}},
\oauthor{\bsnm{Chen}, \binits{W.}},
\oauthor{\bsnm{Sun}, \binits{H.}},
\oauthor{\bsnm{Su}, \binits{Y.}}:
Magicbrush: A manually annotated dataset for instruction-guided image editing.
Advances in Neural Information Processing Systems
\textbf{36}
(2024)
\end{botherref}
\endbibitem

%%% 4
\bibitem[\protect\citeauthoryear{Li et~al.}{2023}]{li2023instructany2pix}
\begin{botherref}
\oauthor{\bsnm{Li}, \binits{S.}},
\oauthor{\bsnm{Singh}, \binits{H.}},
\oauthor{\bsnm{Grover}, \binits{A.}}:
Instructany2pix: Flexible visual editing via multimodal instruction following.
arXiv preprint arXiv:2312.06738
(2023)
\end{botherref}
\endbibitem

%%% 5
\bibitem[\protect\citeauthoryear{Wang et~al.}{2023}]{wang2023instructedit}
\begin{botherref}
\oauthor{\bsnm{Wang}, \binits{Q.}},
\oauthor{\bsnm{Zhang}, \binits{B.}},
\oauthor{\bsnm{Birsak}, \binits{M.}},
\oauthor{\bsnm{Wonka}, \binits{P.}}:
Instructedit: Improving automatic masks for diffusion-based image editing with
  user instructions.
arXiv preprint arXiv:2305.18047
(2023)
\end{botherref}
\endbibitem

%%% 6
\bibitem[\protect\citeauthoryear{Zhang et~al.}{2023}]{zhang2023hive}
\begin{botherref}
\oauthor{\bsnm{Zhang}, \binits{S.}},
\oauthor{\bsnm{Yang}, \binits{X.}},
\oauthor{\bsnm{Feng}, \binits{Y.}},
\oauthor{\bsnm{Qin}, \binits{C.}},
\oauthor{\bsnm{Chen}, \binits{C.-C.}},
\oauthor{\bsnm{Yu}, \binits{N.}},
\oauthor{\bsnm{Chen}, \binits{Z.}},
\oauthor{\bsnm{Wang}, \binits{H.}},
\oauthor{\bsnm{Savarese}, \binits{S.}},
\oauthor{\bsnm{Ermon}, \binits{S.}}, et al.:
Hive: Harnessing human feedback for instructional visual editing.
arXiv preprint arXiv:2303.09618
(2023)
\end{botherref}
\endbibitem

%%% 7
\bibitem[\protect\citeauthoryear{Fu et~al.}{2024}]{fu2023guiding}
\begin{botherref}
\oauthor{\bsnm{Fu}, \binits{T.-J.}},
\oauthor{\bsnm{Hu}, \binits{W.}},
\oauthor{\bsnm{Du}, \binits{X.}},
\oauthor{\bsnm{Wang}, \binits{W.Y.}},
\oauthor{\bsnm{Yang}, \binits{Y.}},
\oauthor{\bsnm{Gan}, \binits{Z.}}:
Guiding instruction-based image editing via multimodal large language models.
International Conference on Learning Representations
(2024)
\end{botherref}
\endbibitem

%%% 8
\bibitem[\protect\citeauthoryear{Ho et~al.}{2020}]{ho2020denoising}
\begin{barticle}
\bauthor{\bsnm{Ho}, \binits{J.}},
\bauthor{\bsnm{Jain}, \binits{A.}},
\bauthor{\bsnm{Abbeel}, \binits{P.}}:
\batitle{Denoising diffusion probabilistic models}.
\bjtitle{Advances in neural information processing systems}
\bvolume{33},
\bfpage{6840}--\blpage{6851}
(\byear{2020})
\end{barticle}
\endbibitem

%%% 9
\bibitem[\protect\citeauthoryear{Sohl-Dickstein et~al.}{2015}]{sohl2015deep}
\begin{bchapter}
\bauthor{\bsnm{Sohl-Dickstein}, \binits{J.}},
\bauthor{\bsnm{Weiss}, \binits{E.}},
\bauthor{\bsnm{Maheswaranathan}, \binits{N.}},
\bauthor{\bsnm{Ganguli}, \binits{S.}}:
\bctitle{Deep unsupervised learning using nonequilibrium thermodynamics}.
In: \bbtitle{International Conference on Machine Learning (ICML)},
pp. \bfpage{2256}--\blpage{2265}
(\byear{2015}).
\bcomment{PMLR}
\end{bchapter}
\endbibitem

%%% 10
\bibitem[\protect\citeauthoryear{Welling and Teh}{2011}]{welling2011bayesian}
\begin{bchapter}
\bauthor{\bsnm{Welling}, \binits{M.}},
\bauthor{\bsnm{Teh}, \binits{Y.W.}}:
\bctitle{Bayesian learning via stochastic gradient langevin dynamics}.
In: \bbtitle{Proceedings of the 28th International Conference on Machine
  Learning (ICML)},
pp. \bfpage{681}--\blpage{688}
(\byear{2011})
\end{bchapter}
\endbibitem

%%% 11
\bibitem[\protect\citeauthoryear{Kulikov et~al.}{2023}]{kulikov2023sinddm}
\begin{bchapter}
\bauthor{\bsnm{Kulikov}, \binits{V.}},
\bauthor{\bsnm{Yadin}, \binits{S.}},
\bauthor{\bsnm{Kleiner}, \binits{M.}},
\bauthor{\bsnm{Michaeli}, \binits{T.}}:
\bctitle{Sinddm: A single image denoising diffusion model}.
In: \bbtitle{International Conference on Machine Learning (ICML)},
pp. \bfpage{17920}--\blpage{17930}
(\byear{2023}).
\bcomment{PMLR}
\end{bchapter}
\endbibitem

%%% 12
\bibitem[\protect\citeauthoryear{Liu et~al.}{2023a}]{liu2023improvedllava}
\begin{botherref}
\oauthor{\bsnm{Liu}, \binits{H.}},
\oauthor{\bsnm{Li}, \binits{C.}},
\oauthor{\bsnm{Li}, \binits{Y.}},
\oauthor{\bsnm{Lee}, \binits{Y.J.}}:
Improved Baselines with Visual Instruction Tuning.
arXiv:2310.03744
(2023)
\end{botherref}
\endbibitem

%%% 13
\bibitem[\protect\citeauthoryear{Liu et~al.}{2023b}]{liu2023llava}
\begin{botherref}
\oauthor{\bsnm{Liu}, \binits{H.}},
\oauthor{\bsnm{Li}, \binits{C.}},
\oauthor{\bsnm{Wu}, \binits{Q.}},
\oauthor{\bsnm{Lee}, \binits{Y.J.}}:
Visual Instruction Tuning.
NeurIPS
(2023)
\end{botherref}
\endbibitem

%%% 14
\bibitem[\protect\citeauthoryear{Fei et~al.}{2024a}]{fei2024enhancing}
\begin{botherref}
\oauthor{\bsnm{Fei}, \binits{H.}},
\oauthor{\bsnm{Wu}, \binits{S.}},
\oauthor{\bsnm{Zhang}, \binits{M.}},
\oauthor{\bsnm{Zhang}, \binits{M.}},
\oauthor{\bsnm{Chua}, \binits{T.-S.}},
\oauthor{\bsnm{Yan}, \binits{S.}}:
Enhancing video-language representations with structural spatio-temporal
  alignment.
IEEE Transactions on Pattern Analysis and Machine Intelligence
(2024)
\end{botherref}
\endbibitem

%%% 15
\bibitem[\protect\citeauthoryear{Fei et~al.}{2024b}]{fei2024video}
\begin{bchapter}
\bauthor{\bsnm{Fei}, \binits{H.}},
\bauthor{\bsnm{Wu}, \binits{S.}},
\bauthor{\bsnm{Ji}, \binits{W.}},
\bauthor{\bsnm{Zhang}, \binits{H.}},
\bauthor{\bsnm{Zhang}, \binits{M.}},
\bauthor{\bsnm{Lee}, \binits{M.-L.}},
\bauthor{\bsnm{Hsu}, \binits{W.}}:
\bctitle{Video-of-thought: Step-by-step video reasoning from perception to
  cognition}.
In: \bbtitle{Forty-first International Conference on Machine Learning}
(\byear{2024})
\end{bchapter}
\endbibitem

%%% 16
\bibitem[\protect\citeauthoryear{Fei et~al.}{2024c}]{fei2024vitron}
\begin{botherref}
\oauthor{\bsnm{Fei}, \binits{H.}},
\oauthor{\bsnm{Wu}, \binits{S.}},
\oauthor{\bsnm{Zhang}, \binits{H.}},
\oauthor{\bsnm{Chua}, \binits{T.-S.}},
\oauthor{\bsnm{Yan}, \binits{S.}}:
VITRON: A Unified Pixel-level Vision LLM for Understanding, Generating,
  Segmenting, Editing.
CoRR
(2024)
\end{botherref}
\endbibitem

%%% 17
\bibitem[\protect\citeauthoryear{Ma et~al.}{2024}]{ma2024inf}
\begin{botherref}
\oauthor{\bsnm{Ma}, \binits{Y.}},
\oauthor{\bsnm{Wang}, \binits{Z.}},
\oauthor{\bsnm{Sun}, \binits{X.}},
\oauthor{\bsnm{Lin}, \binits{W.}},
\oauthor{\bsnm{Zhou}, \binits{Q.}},
\oauthor{\bsnm{Ji}, \binits{J.}},
\oauthor{\bsnm{Ji}, \binits{R.}}:
Inf-llava: Dual-perspective perception for high-resolution multimodal large
  language model.
arXiv preprint arXiv:2407.16198
(2024)
\end{botherref}
\endbibitem

%%% 18
\bibitem[\protect\citeauthoryear{Huang et~al.}{2024}]{huang2023smartedit}
\begin{botherref}
\oauthor{\bsnm{Huang}, \binits{Y.}},
\oauthor{\bsnm{Xie}, \binits{L.}},
\oauthor{\bsnm{Wang}, \binits{X.}},
\oauthor{\bsnm{Yuan}, \binits{Z.}},
\oauthor{\bsnm{Cun}, \binits{X.}},
\oauthor{\bsnm{Ge}, \binits{Y.}},
\oauthor{\bsnm{Zhou}, \binits{J.}},
\oauthor{\bsnm{Dong}, \binits{C.}},
\oauthor{\bsnm{Huang}, \binits{R.}},
\oauthor{\bsnm{Zhang}, \binits{R.}}, et al.:
Smartedit: Exploring complex instruction-based image editing with multimodal
  large language models.
Proceedings of the IEEE/CVF Conference on Computer Vision and Pattern
  Recognition
(2024)
\end{botherref}
\endbibitem

%%% 19
\bibitem[\protect\citeauthoryear{Radford et~al.}{2021}]{radford2021learning}
\begin{bchapter}
\bauthor{\bsnm{Radford}, \binits{A.}},
\bauthor{\bsnm{Kim}, \binits{J.W.}},
\bauthor{\bsnm{Hallacy}, \binits{C.}},
\bauthor{\bsnm{Ramesh}, \binits{A.}},
\bauthor{\bsnm{Goh}, \binits{G.}},
\bauthor{\bsnm{Agarwal}, \binits{S.}},
\bauthor{\bsnm{Sastry}, \binits{G.}},
\bauthor{\bsnm{Askell}, \binits{A.}},
\bauthor{\bsnm{Mishkin}, \binits{P.}},
\bauthor{\bsnm{Clark}, \binits{J.}}, \betal:
\bctitle{Learning transferable visual models from natural language
  supervision}.
In: \bbtitle{International Conference on Machine Learning (ICML)},
pp. \bfpage{8748}--\blpage{8763}
(\byear{2021}).
\bcomment{PMLR}
\end{bchapter}
\endbibitem

%%% 20
\bibitem[\protect\citeauthoryear{Korhonen and You}{2012}]{korhonen2012peak}
\begin{bchapter}
\bauthor{\bsnm{Korhonen}, \binits{J.}},
\bauthor{\bsnm{You}, \binits{J.}}:
\bctitle{Peak signal-to-noise ratio revisited: Is simple beautiful?}
In: \bbtitle{2012 Fourth International Workshop on Quality of Multimedia
  Experience},
pp. \bfpage{37}--\blpage{38}
(\byear{2012}).
\bcomment{IEEE}
\end{bchapter}
\endbibitem

%%% 21
\bibitem[\protect\citeauthoryear{Wang et~al.}{2004}]{wang2004image}
\begin{barticle}
\bauthor{\bsnm{Wang}, \binits{Z.}},
\bauthor{\bsnm{Bovik}, \binits{A.C.}},
\bauthor{\bsnm{Sheikh}, \binits{H.R.}},
\bauthor{\bsnm{Simoncelli}, \binits{E.P.}}:
\batitle{Image quality assessment: from error visibility to structural
  similarity}.
\bjtitle{IEEE transactions on image processing}
\bvolume{13}(\bissue{4}),
\bfpage{600}--\blpage{612}
(\byear{2004})
\end{barticle}
\endbibitem

%%% 22
\bibitem[\protect\citeauthoryear{Zhang et~al.}{2018}]{zhang2018unreasonable}
\begin{bchapter}
\bauthor{\bsnm{Zhang}, \binits{R.}},
\bauthor{\bsnm{Isola}, \binits{P.}},
\bauthor{\bsnm{Efros}, \binits{A.A.}},
\bauthor{\bsnm{Shechtman}, \binits{E.}},
\bauthor{\bsnm{Wang}, \binits{O.}}:
\bctitle{The unreasonable effectiveness of deep features as a perceptual
  metric}.
In: \bbtitle{Proceedings of the IEEE Conference on Computer Vision and Pattern
  Recognition},
pp. \bfpage{586}--\blpage{595}
(\byear{2018})
\end{bchapter}
\endbibitem

%%% 23
\bibitem[\protect\citeauthoryear{Kawar et~al.}{2023}]{kawar2023imagic}
\begin{bchapter}
\bauthor{\bsnm{Kawar}, \binits{B.}},
\bauthor{\bsnm{Zada}, \binits{S.}},
\bauthor{\bsnm{Lang}, \binits{O.}},
\bauthor{\bsnm{Tov}, \binits{O.}},
\bauthor{\bsnm{Chang}, \binits{H.}},
\bauthor{\bsnm{Dekel}, \binits{T.}},
\bauthor{\bsnm{Mosseri}, \binits{I.}},
\bauthor{\bsnm{Irani}, \binits{M.}}:
\bctitle{Imagic: Text-based real image editing with diffusion models}.
In: \bbtitle{Proceedings of the IEEE/CVF Conference on Computer Vision and
  Pattern Recognition},
pp. \bfpage{6007}--\blpage{6017}
(\byear{2023})
\end{bchapter}
\endbibitem

%%% 24
\bibitem[\protect\citeauthoryear{Wang et~al.}{2023}]{wang2023imagen}
\begin{bchapter}
\bauthor{\bsnm{Wang}, \binits{S.}},
\bauthor{\bsnm{Saharia}, \binits{C.}},
\bauthor{\bsnm{Montgomery}, \binits{C.}},
\bauthor{\bsnm{Pont-Tuset}, \binits{J.}},
\bauthor{\bsnm{Noy}, \binits{S.}},
\bauthor{\bsnm{Pellegrini}, \binits{S.}},
\bauthor{\bsnm{Onoe}, \binits{Y.}},
\bauthor{\bsnm{Laszlo}, \binits{S.}},
\bauthor{\bsnm{Fleet}, \binits{D.J.}},
\bauthor{\bsnm{Soricut}, \binits{R.}}, \betal:
\bctitle{Imagen editor and editbench: Advancing and evaluating text-guided
  image inpainting}.
In: \bbtitle{Proceedings of the IEEE/CVF Conference on Computer Vision and
  Pattern Recognition},
pp. \bfpage{18359}--\blpage{18369}
(\byear{2023})
\end{bchapter}
\endbibitem

%%% 25
\bibitem[\protect\citeauthoryear{Basu et~al.}{2023}]{basu2023editval}
\begin{botherref}
\oauthor{\bsnm{Basu}, \binits{S.}},
\oauthor{\bsnm{Saberi}, \binits{M.}},
\oauthor{\bsnm{Bhardwaj}, \binits{S.}},
\oauthor{\bsnm{Chegini}, \binits{A.M.}},
\oauthor{\bsnm{Massiceti}, \binits{D.}},
\oauthor{\bsnm{Sanjabi}, \binits{M.}},
\oauthor{\bsnm{Hu}, \binits{S.X.}},
\oauthor{\bsnm{Feizi}, \binits{S.}}:
Editval: Benchmarking diffusion based text-guided image editing methods.
arXiv preprint arXiv:2310.02426
(2023)
\end{botherref}
\endbibitem

%%% 26
\bibitem[\protect\citeauthoryear{Huang et~al.}{2024}]{huang2024diffusion}
\begin{botherref}
\oauthor{\bsnm{Huang}, \binits{Y.}},
\oauthor{\bsnm{Huang}, \binits{J.}},
\oauthor{\bsnm{Liu}, \binits{Y.}},
\oauthor{\bsnm{Yan}, \binits{M.}},
\oauthor{\bsnm{Lv}, \binits{J.}},
\oauthor{\bsnm{Liu}, \binits{J.}},
\oauthor{\bsnm{Xiong}, \binits{W.}},
\oauthor{\bsnm{Zhang}, \binits{H.}},
\oauthor{\bsnm{Chen}, \binits{S.}},
\oauthor{\bsnm{Cao}, \binits{L.}}:
Diffusion model-based image editing: A survey.
arXiv preprint arXiv:2402.17525
(2024)
\end{botherref}
\endbibitem

%%% 27
\bibitem[\protect\citeauthoryear{Sheynin et~al.}{2023}]{sheynin2023emu}
\begin{botherref}
\oauthor{\bsnm{Sheynin}, \binits{S.}},
\oauthor{\bsnm{Polyak}, \binits{A.}},
\oauthor{\bsnm{Singer}, \binits{U.}},
\oauthor{\bsnm{Kirstain}, \binits{Y.}},
\oauthor{\bsnm{Zohar}, \binits{A.}},
\oauthor{\bsnm{Ashual}, \binits{O.}},
\oauthor{\bsnm{Parikh}, \binits{D.}},
\oauthor{\bsnm{Taigman}, \binits{Y.}}:
Emu edit: Precise image editing via recognition and generation tasks.
arXiv preprint arXiv:2311.10089
(2023)
\end{botherref}
\endbibitem

%%% 28
\bibitem[\protect\citeauthoryear{Ma et~al.}{2024}]{ma2024i2ebench}
\begin{bchapter}
\bauthor{\bsnm{Ma}, \binits{Y.}},
\bauthor{\bsnm{Ji}, \binits{J.}},
\bauthor{\bsnm{Ye}, \binits{K.}},
\bauthor{\bsnm{Lin}, \binits{W.}},
\bauthor{\bsnm{Zheng}, \binits{Y.}},
\bauthor{\bsnm{Zhou}, \binits{Q.}},
\bauthor{\bsnm{Sun}, \binits{X.}},
\bauthor{\bsnm{Ji}, \binits{R.}}, \betal:
\bctitle{I2ebench: A comprehensive benchmark for instruction-based image
  editing}.
In: \bbtitle{The Thirty-eighth Annual Conference on Neural Information
  Processing Systems}
(\byear{2024})
\end{bchapter}
\endbibitem

%%% 29
\bibitem[\protect\citeauthoryear{Goodfellow
  et~al.}{2014}]{goodfellow2014generative}
\begin{botherref}
\oauthor{\bsnm{Goodfellow}, \binits{I.}},
\oauthor{\bsnm{Pouget-Abadie}, \binits{J.}},
\oauthor{\bsnm{Mirza}, \binits{M.}},
\oauthor{\bsnm{Xu}, \binits{B.}},
\oauthor{\bsnm{Warde-Farley}, \binits{D.}},
\oauthor{\bsnm{Ozair}, \binits{S.}},
\oauthor{\bsnm{Courville}, \binits{A.}},
\oauthor{\bsnm{Bengio}, \binits{Y.}}:
Generative adversarial nets.
Advances in neural information processing systems
\textbf{27}
(2014)
\end{botherref}
\endbibitem

%%% 30
\bibitem[\protect\citeauthoryear{Karras et~al.}{2019}]{karras2019style}
\begin{bchapter}
\bauthor{\bsnm{Karras}, \binits{T.}},
\bauthor{\bsnm{Laine}, \binits{S.}},
\bauthor{\bsnm{Aila}, \binits{T.}}:
\bctitle{A style-based generator architecture for generative adversarial
  networks}.
In: \bbtitle{Proceedings of the IEEE/CVF Conference on Computer Vision and
  Pattern Recognition},
pp. \bfpage{4401}--\blpage{4410}
(\byear{2019})
\end{bchapter}
\endbibitem

%%% 31
\bibitem[\protect\citeauthoryear{Song et~al.}{2020}]{song2020denoising}
\begin{botherref}
\oauthor{\bsnm{Song}, \binits{J.}},
\oauthor{\bsnm{Meng}, \binits{C.}},
\oauthor{\bsnm{Ermon}, \binits{S.}}:
Denoising diffusion implicit models.
arXiv preprint arXiv:2010.02502
(2020)
\end{botherref}
\endbibitem

%%% 32
\bibitem[\protect\citeauthoryear{Saharia
  et~al.}{2022}]{saharia2022photorealistic}
\begin{barticle}
\bauthor{\bsnm{Saharia}, \binits{C.}},
\bauthor{\bsnm{Chan}, \binits{W.}},
\bauthor{\bsnm{Saxena}, \binits{S.}},
\bauthor{\bsnm{Li}, \binits{L.}},
\bauthor{\bsnm{Whang}, \binits{J.}},
\bauthor{\bsnm{Denton}, \binits{E.L.}},
\bauthor{\bsnm{Ghasemipour}, \binits{K.}},
\bauthor{\bsnm{Gontijo~Lopes}, \binits{R.}},
\bauthor{\bsnm{Karagol~Ayan}, \binits{B.}},
\bauthor{\bsnm{Salimans}, \binits{T.}}, \betal:
\batitle{Photorealistic text-to-image diffusion models with deep language
  understanding}.
\bjtitle{Advances in neural information processing systems}
\bvolume{35},
\bfpage{36479}--\blpage{36494}
(\byear{2022})
\end{barticle}
\endbibitem

%%% 33
\bibitem[\protect\citeauthoryear{Ramesh et~al.}{2021}]{ramesh2021zero}
\begin{bchapter}
\bauthor{\bsnm{Ramesh}, \binits{A.}},
\bauthor{\bsnm{Pavlov}, \binits{M.}},
\bauthor{\bsnm{Goh}, \binits{G.}},
\bauthor{\bsnm{Gray}, \binits{S.}},
\bauthor{\bsnm{Voss}, \binits{C.}},
\bauthor{\bsnm{Radford}, \binits{A.}},
\bauthor{\bsnm{Chen}, \binits{M.}},
\bauthor{\bsnm{Sutskever}, \binits{I.}}:
\bctitle{Zero-shot text-to-image generation}.
In: \bbtitle{International Conference on Machine Learning},
pp. \bfpage{8821}--\blpage{8831}
(\byear{2021}).
\bcomment{Pmlr}
\end{bchapter}
\endbibitem

%%% 34
\bibitem[\protect\citeauthoryear{Avrahami et~al.}{2022}]{avrahami2022blended}
\begin{bchapter}
\bauthor{\bsnm{Avrahami}, \binits{O.}},
\bauthor{\bsnm{Lischinski}, \binits{D.}},
\bauthor{\bsnm{Fried}, \binits{O.}}:
\bctitle{Blended diffusion for text-driven editing of natural images}.
In: \bbtitle{Proceedings of the IEEE/CVF Conference on Computer Vision and
  Pattern Recognition},
pp. \bfpage{18208}--\blpage{18218}
(\byear{2022})
\end{bchapter}
\endbibitem

%%% 35
\bibitem[\protect\citeauthoryear{Brown et~al.}{2020}]{brown2020language}
\begin{barticle}
\bauthor{\bsnm{Brown}, \binits{T.}},
\bauthor{\bsnm{Mann}, \binits{B.}},
\bauthor{\bsnm{Ryder}, \binits{N.}},
\bauthor{\bsnm{Subbiah}, \binits{M.}},
\bauthor{\bsnm{Kaplan}, \binits{J.D.}},
\bauthor{\bsnm{Dhariwal}, \binits{P.}},
\bauthor{\bsnm{Neelakantan}, \binits{A.}},
\bauthor{\bsnm{Shyam}, \binits{P.}},
\bauthor{\bsnm{Sastry}, \binits{G.}},
\bauthor{\bsnm{Askell}, \binits{A.}}, \betal:
\batitle{Language models are few-shot learners}.
\bjtitle{Advances in neural information processing systems}
\bvolume{33},
\bfpage{1877}--\blpage{1901}
(\byear{2020})
\end{barticle}
\endbibitem

%%% 36
\bibitem[\protect\citeauthoryear{Achiam et~al.}{2023}]{achiam2023gpt}
\begin{botherref}
\oauthor{\bsnm{Achiam}, \binits{J.}},
\oauthor{\bsnm{Adler}, \binits{S.}},
\oauthor{\bsnm{Agarwal}, \binits{S.}},
\oauthor{\bsnm{Ahmad}, \binits{L.}},
\oauthor{\bsnm{Akkaya}, \binits{I.}},
\oauthor{\bsnm{Aleman}, \binits{F.L.}},
\oauthor{\bsnm{Almeida}, \binits{D.}},
\oauthor{\bsnm{Altenschmidt}, \binits{J.}},
\oauthor{\bsnm{Altman}, \binits{S.}},
\oauthor{\bsnm{Anadkat}, \binits{S.}}, et al.:
Gpt-4 technical report.
arXiv preprint arXiv:2303.08774
(2023)
\end{botherref}
\endbibitem

%%% 37
\bibitem[\protect\citeauthoryear{Brack et~al.}{2024}]{brack2024ledits++}
\begin{bchapter}
\bauthor{\bsnm{Brack}, \binits{M.}},
\bauthor{\bsnm{Friedrich}, \binits{F.}},
\bauthor{\bsnm{Kornmeier}, \binits{K.}},
\bauthor{\bsnm{Tsaban}, \binits{L.}},
\bauthor{\bsnm{Schramowski}, \binits{P.}},
\bauthor{\bsnm{Kersting}, \binits{K.}},
\bauthor{\bsnm{Passos}, \binits{A.}}:
\bctitle{Ledits++: Limitless image editing using text-to-image models}.
In: \bbtitle{Proceedings of the IEEE/CVF Conference on Computer Vision and
  Pattern Recognition},
pp. \bfpage{8861}--\blpage{8870}
(\byear{2024})
\end{bchapter}
\endbibitem

%%% 38
\bibitem[\protect\citeauthoryear{Hui et~al.}{2024}]{hui2024hq}
\begin{botherref}
\oauthor{\bsnm{Hui}, \binits{M.}},
\oauthor{\bsnm{Yang}, \binits{S.}},
\oauthor{\bsnm{Zhao}, \binits{B.}},
\oauthor{\bsnm{Shi}, \binits{Y.}},
\oauthor{\bsnm{Wang}, \binits{H.}},
\oauthor{\bsnm{Wang}, \binits{P.}},
\oauthor{\bsnm{Zhou}, \binits{Y.}},
\oauthor{\bsnm{Xie}, \binits{C.}}:
Hq-edit: A high-quality dataset for instruction-based image editing.
arXiv preprint arXiv:2404.09990
(2024)
\end{botherref}
\endbibitem

%%% 39
\bibitem[\protect\citeauthoryear{Marino et~al.}{2019}]{marino2019ok}
\begin{bchapter}
\bauthor{\bsnm{Marino}, \binits{K.}},
\bauthor{\bsnm{Rastegari}, \binits{M.}},
\bauthor{\bsnm{Farhadi}, \binits{A.}},
\bauthor{\bsnm{Mottaghi}, \binits{R.}}:
\bctitle{Ok-vqa: A visual question answering benchmark requiring external
  knowledge}.
In: \bbtitle{Proceedings of the IEEE/cvf Conference on Computer Vision and
  Pattern Recognition},
pp. \bfpage{3195}--\blpage{3204}
(\byear{2019})
\end{bchapter}
\endbibitem

%%% 40
\bibitem[\protect\citeauthoryear{Hudson and Manning}{2019}]{hudson2019gqa}
\begin{bchapter}
\bauthor{\bsnm{Hudson}, \binits{D.A.}},
\bauthor{\bsnm{Manning}, \binits{C.D.}}:
\bctitle{Gqa: A new dataset for real-world visual reasoning and compositional
  question answering}.
In: \bbtitle{Proceedings of the IEEE/CVF Conference on Computer Vision and
  Pattern Recognition},
pp. \bfpage{6700}--\blpage{6709}
(\byear{2019})
\end{bchapter}
\endbibitem

%%% 41
\bibitem[\protect\citeauthoryear{Bigham et~al.}{2010}]{bigham2010vizwiz}
\begin{bchapter}
\bauthor{\bsnm{Bigham}, \binits{J.P.}},
\bauthor{\bsnm{Jayant}, \binits{C.}},
\bauthor{\bsnm{Ji}, \binits{H.}},
\bauthor{\bsnm{Little}, \binits{G.}},
\bauthor{\bsnm{Miller}, \binits{A.}},
\bauthor{\bsnm{Miller}, \binits{R.C.}},
\bauthor{\bsnm{Miller}, \binits{R.}},
\bauthor{\bsnm{Tatarowicz}, \binits{A.}},
\bauthor{\bsnm{White}, \binits{B.}},
\bauthor{\bsnm{White}, \binits{S.}}, \betal:
\bctitle{Vizwiz: nearly real-time answers to visual questions}.
In: \bbtitle{Proceedings of the 23nd Annual ACM Symposium on User Interface
  Software and Technology},
pp. \bfpage{333}--\blpage{342}
(\byear{2010})
\end{bchapter}
\endbibitem

%%% 42
\bibitem[\protect\citeauthoryear{Lu et~al.}{2022}]{lu2022learn}
\begin{barticle}
\bauthor{\bsnm{Lu}, \binits{P.}},
\bauthor{\bsnm{Mishra}, \binits{S.}},
\bauthor{\bsnm{Xia}, \binits{T.}},
\bauthor{\bsnm{Qiu}, \binits{L.}},
\bauthor{\bsnm{Chang}, \binits{K.-W.}},
\bauthor{\bsnm{Zhu}, \binits{S.-C.}},
\bauthor{\bsnm{Tafjord}, \binits{O.}},
\bauthor{\bsnm{Clark}, \binits{P.}},
\bauthor{\bsnm{Kalyan}, \binits{A.}}:
\batitle{Learn to explain: Multimodal reasoning via thought chains for science
  question answering}.
\bjtitle{Advances in Neural Information Processing Systems}
\bvolume{35},
\bfpage{2507}--\blpage{2521}
(\byear{2022})
\end{barticle}
\endbibitem

%%% 43
\bibitem[\protect\citeauthoryear{Li et~al.}{2023a}]{li2023evaluating}
\begin{botherref}
\oauthor{\bsnm{Li}, \binits{Y.}},
\oauthor{\bsnm{Du}, \binits{Y.}},
\oauthor{\bsnm{Zhou}, \binits{K.}},
\oauthor{\bsnm{Wang}, \binits{J.}},
\oauthor{\bsnm{Zhao}, \binits{W.X.}},
\oauthor{\bsnm{Wen}, \binits{J.-R.}}:
Evaluating object hallucination in large vision-language models.
arXiv preprint arXiv:2305.10355
(2023)
\end{botherref}
\endbibitem

%%% 44
\bibitem[\protect\citeauthoryear{Li et~al.}{2023b}]{li2023seed}
\begin{botherref}
\oauthor{\bsnm{Li}, \binits{B.}},
\oauthor{\bsnm{Wang}, \binits{R.}},
\oauthor{\bsnm{Wang}, \binits{G.}},
\oauthor{\bsnm{Ge}, \binits{Y.}},
\oauthor{\bsnm{Ge}, \binits{Y.}},
\oauthor{\bsnm{Shan}, \binits{Y.}}:
Seed-bench: Benchmarking multimodal llms with generative comprehension.
arXiv preprint arXiv:2307.16125
(2023)
\end{botherref}
\endbibitem

%%% 45
\bibitem[\protect\citeauthoryear{Li et~al.}{2023c}]{li2023mvbench}
\begin{botherref}
\oauthor{\bsnm{Li}, \binits{K.}},
\oauthor{\bsnm{Wang}, \binits{Y.}},
\oauthor{\bsnm{He}, \binits{Y.}},
\oauthor{\bsnm{Li}, \binits{Y.}},
\oauthor{\bsnm{Wang}, \binits{Y.}},
\oauthor{\bsnm{Liu}, \binits{Y.}},
\oauthor{\bsnm{Wang}, \binits{Z.}},
\oauthor{\bsnm{Xu}, \binits{J.}},
\oauthor{\bsnm{Chen}, \binits{G.}},
\oauthor{\bsnm{Luo}, \binits{P.}}, et al.:
Mvbench: A comprehensive multi-modal video understanding benchmark.
arXiv preprint arXiv:2311.17005
(2023)
\end{botherref}
\endbibitem

%%% 46
\bibitem[\protect\citeauthoryear{Yu et~al.}{2023}]{yu2023mm}
\begin{botherref}
\oauthor{\bsnm{Yu}, \binits{W.}},
\oauthor{\bsnm{Yang}, \binits{Z.}},
\oauthor{\bsnm{Li}, \binits{L.}},
\oauthor{\bsnm{Wang}, \binits{J.}},
\oauthor{\bsnm{Lin}, \binits{K.}},
\oauthor{\bsnm{Liu}, \binits{Z.}},
\oauthor{\bsnm{Wang}, \binits{X.}},
\oauthor{\bsnm{Wang}, \binits{L.}}:
Mm-vet: Evaluating large multimodal models for integrated capabilities.
arXiv preprint arXiv:2308.02490
(2023)
\end{botherref}
\endbibitem

%%% 47
\bibitem[\protect\citeauthoryear{Wu et~al.}{2023}]{wu2023q}
\begin{botherref}
\oauthor{\bsnm{Wu}, \binits{H.}},
\oauthor{\bsnm{Zhang}, \binits{Z.}},
\oauthor{\bsnm{Zhang}, \binits{E.}},
\oauthor{\bsnm{Chen}, \binits{C.}},
\oauthor{\bsnm{Liao}, \binits{L.}},
\oauthor{\bsnm{Wang}, \binits{A.}},
\oauthor{\bsnm{Li}, \binits{C.}},
\oauthor{\bsnm{Sun}, \binits{W.}},
\oauthor{\bsnm{Yan}, \binits{Q.}},
\oauthor{\bsnm{Zhai}, \binits{G.}}, et al.:
Q-bench: A benchmark for general-purpose foundation models on low-level vision.
arXiv preprint arXiv:2309.14181
(2023)
\end{botherref}
\endbibitem

%%% 48
\bibitem[\protect\citeauthoryear{Lin et~al.}{2014}]{lin2014microsoft}
\begin{bchapter}
\bauthor{\bsnm{Lin}, \binits{T.-Y.}},
\bauthor{\bsnm{Maire}, \binits{M.}},
\bauthor{\bsnm{Belongie}, \binits{S.}},
\bauthor{\bsnm{Hays}, \binits{J.}},
\bauthor{\bsnm{Perona}, \binits{P.}},
\bauthor{\bsnm{Ramanan}, \binits{D.}},
\bauthor{\bsnm{Doll{\'a}r}, \binits{P.}},
\bauthor{\bsnm{Zitnick}, \binits{C.L.}}:
\bctitle{Microsoft coco: Common objects in context}.
In: \bbtitle{Computer Vision--ECCV 2014: 13th European Conference, Zurich,
  Switzerland, September 6-12, 2014, Proceedings, Part V 13},
pp. \bfpage{740}--\blpage{755}
(\byear{2014}).
\bcomment{Springer}
\end{bchapter}
\endbibitem

%%% 49
\bibitem[\protect\citeauthoryear{Guo et~al.}{2023}]{guo2023sky}
\begin{bchapter}
\bauthor{\bsnm{Guo}, \binits{Y.}},
\bauthor{\bsnm{Xiao}, \binits{X.}},
\bauthor{\bsnm{Chang}, \binits{Y.}},
\bauthor{\bsnm{Deng}, \binits{S.}},
\bauthor{\bsnm{Yan}, \binits{L.}}:
\bctitle{From sky to the ground: A large-scale benchmark and simple baseline
  towards real rain removal}.
In: \bbtitle{Proceedings of the IEEE/CVF International Conference on Computer
  Vision},
pp. \bfpage{12097}--\blpage{12107}
(\byear{2023})
\end{bchapter}
\endbibitem

%%% 50
\bibitem[\protect\citeauthoryear{Martin et~al.}{2001}]{MartinFTM01}
\begin{bchapter}
\bauthor{\bsnm{Martin}, \binits{D.}},
\bauthor{\bsnm{Fowlkes}, \binits{C.}},
\bauthor{\bsnm{Tal}, \binits{D.}},
\bauthor{\bsnm{Malik}, \binits{J.}}:
\bctitle{A database of human segmented natural images and its application to
  evaluating segmentation algorithms and measuring ecological statistics}.
In: \bbtitle{Proc. 8th Int'l Conf. Computer Vision},
vol. \bseriesno{2},
pp. \bfpage{416}--\blpage{423}
(\byear{2001})
\end{bchapter}
\endbibitem

%%% 51
\bibitem[\protect\citeauthoryear{Chen et~al.}{2021}]{chen2021all}
\begin{bchapter}
\bauthor{\bsnm{Chen}, \binits{W.-T.}},
\bauthor{\bsnm{Fang}, \binits{H.-Y.}},
\bauthor{\bsnm{Hsieh}, \binits{C.-L.}},
\bauthor{\bsnm{Tsai}, \binits{C.-C.}},
\bauthor{\bsnm{Chen}, \binits{I.}},
\bauthor{\bsnm{Ding}, \binits{J.-J.}},
\bauthor{\bsnm{Kuo}, \binits{S.-Y.}}, \betal:
\bctitle{All snow removed: Single image desnowing algorithm using hierarchical
  dual-tree complex wavelet representation and contradict channel loss}.
In: \bbtitle{Proceedings of the IEEE/CVF International Conference on Computer
  Vision},
pp. \bfpage{4196}--\blpage{4205}
(\byear{2021})
\end{bchapter}
\endbibitem

%%% 52
\bibitem[\protect\citeauthoryear{Ancuti et~al.}{2019}]{ancuti2019dense}
\begin{bchapter}
\bauthor{\bsnm{Ancuti}, \binits{C.O.}},
\bauthor{\bsnm{Ancuti}, \binits{C.}},
\bauthor{\bsnm{Sbert}, \binits{M.}},
\bauthor{\bsnm{Timofte}, \binits{R.}}:
\bctitle{Dense-haze: A benchmark for image dehazing with dense-haze and
  haze-free images}.
In: \bbtitle{2019 IEEE International Conference on Image Processing (ICIP)},
pp. \bfpage{1014}--\blpage{1018}
(\byear{2019}).
\bcomment{IEEE}
\end{bchapter}
\endbibitem

%%% 53
\bibitem[\protect\citeauthoryear{Liu et~al.}{2021a}]{liu2021synthetic}
\begin{bchapter}
\bauthor{\bsnm{Liu}, \binits{Y.}},
\bauthor{\bsnm{Zhu}, \binits{L.}},
\bauthor{\bsnm{Pei}, \binits{S.}},
\bauthor{\bsnm{Fu}, \binits{H.}},
\bauthor{\bsnm{Qin}, \binits{J.}},
\bauthor{\bsnm{Zhang}, \binits{Q.}},
\bauthor{\bsnm{Wan}, \binits{L.}},
\bauthor{\bsnm{Feng}, \binits{W.}}:
\bctitle{From synthetic to real: Image dehazing collaborating with unlabeled
  real data}.
In: \bbtitle{Proceedings of the 29th ACM International Conference on
  Multimedia},
pp. \bfpage{50}--\blpage{58}
(\byear{2021})
\end{bchapter}
\endbibitem

%%% 54
\bibitem[\protect\citeauthoryear{Liu et~al.}{2021b}]{Liu_2021_WACV}
\begin{bchapter}
\bauthor{\bsnm{Liu}, \binits{Y.}},
\bauthor{\bsnm{Zhu}, \binits{Z.}},
\bauthor{\bsnm{Bai}, \binits{X.}}:
\bctitle{Wdnet: Watermark-decomposition network for visible watermark removal}.
In: \bbtitle{IEEE/CVF Winter Conference on Applications of Computer Vision
  (WACV)},
pp. \bfpage{3685}--\blpage{3693}
(\byear{2021})
\end{bchapter}
\endbibitem

%%% 55
\bibitem[\protect\citeauthoryear{Qu et~al.}{2017}]{qu2017deshadownet}
\begin{bchapter}
\bauthor{\bsnm{Qu}, \binits{L.}},
\bauthor{\bsnm{Tian}, \binits{J.}},
\bauthor{\bsnm{He}, \binits{S.}},
\bauthor{\bsnm{Tang}, \binits{Y.}},
\bauthor{\bsnm{Lau}, \binits{R.W.}}:
\bctitle{Deshadownet: A multi-context embedding deep network for shadow
  removal}.
In: \bbtitle{Proceedings of the IEEE Conference on Computer Vision and Pattern
  Recognition},
pp. \bfpage{4067}--\blpage{4075}
(\byear{2017})
\end{bchapter}
\endbibitem

%%% 56
\bibitem[\protect\citeauthoryear{Nah et~al.}{2017}]{Nah_2017_CVPR}
\begin{bchapter}
\bauthor{\bsnm{Nah}, \binits{S.}},
\bauthor{\bsnm{Kim}, \binits{T.H.}},
\bauthor{\bsnm{Lee}, \binits{K.M.}}:
\bctitle{Deep multi-scale convolutional neural network for dynamic scene
  deblurring}.
In: \bbtitle{The IEEE Conference on Computer Vision and Pattern Recognition
  (CVPR)}
(\byear{2017})
\end{bchapter}
\endbibitem

%%% 57
\bibitem[\protect\citeauthoryear{Shen et~al.}{2019}]{shen2019human}
\begin{bchapter}
\bauthor{\bsnm{Shen}, \binits{Z.}},
\bauthor{\bsnm{Wang}, \binits{W.}},
\bauthor{\bsnm{Lu}, \binits{X.}},
\bauthor{\bsnm{Shen}, \binits{J.}},
\bauthor{\bsnm{Ling}, \binits{H.}},
\bauthor{\bsnm{Xu}, \binits{T.}},
\bauthor{\bsnm{Shao}, \binits{L.}}:
\bctitle{Human-aware motion deblurring}.
In: \bbtitle{Proceedings of the IEEE/CVF International Conference on Computer
  Vision},
pp. \bfpage{5572}--\blpage{5581}
(\byear{2019})
\end{bchapter}
\endbibitem

%%% 58
\bibitem[\protect\citeauthoryear{Wei et~al.}{2018}]{wei2018deep}
\begin{botherref}
\oauthor{\bsnm{Wei}, \binits{C.}},
\oauthor{\bsnm{Wang}, \binits{W.}},
\oauthor{\bsnm{Yang}, \binits{W.}},
\oauthor{\bsnm{Liu}, \binits{J.}}:
Deep retinex decomposition for low-light enhancement.
arXiv preprint arXiv:1808.04560
(2018)
\end{botherref}
\endbibitem

%%% 59
\bibitem[\protect\citeauthoryear{Gao et~al.}{2024}]{gao2024sphinx}
\begin{botherref}
\oauthor{\bsnm{Gao}, \binits{P.}},
\oauthor{\bsnm{Zhang}, \binits{R.}},
\oauthor{\bsnm{Liu}, \binits{C.}},
\oauthor{\bsnm{Qiu}, \binits{L.}},
\oauthor{\bsnm{Huang}, \binits{S.}},
\oauthor{\bsnm{Lin}, \binits{W.}},
\oauthor{\bsnm{Zhao}, \binits{S.}},
\oauthor{\bsnm{Geng}, \binits{S.}},
\oauthor{\bsnm{Lin}, \binits{Z.}},
\oauthor{\bsnm{Jin}, \binits{P.}}, et al.:
Sphinx-x: Scaling data and parameters for a family of multi-modal large
  language models.
arXiv preprint arXiv:2402.05935
(2024)
\end{botherref}
\endbibitem

%%% 60
\bibitem[\protect\citeauthoryear{Chu et~al.}{2024}]{chu2024mobilevlm}
\begin{botherref}
\oauthor{\bsnm{Chu}, \binits{X.}},
\oauthor{\bsnm{Qiao}, \binits{L.}},
\oauthor{\bsnm{Zhang}, \binits{X.}},
\oauthor{\bsnm{Xu}, \binits{S.}},
\oauthor{\bsnm{Wei}, \binits{F.}},
\oauthor{\bsnm{Yang}, \binits{Y.}},
\oauthor{\bsnm{Sun}, \binits{X.}},
\oauthor{\bsnm{Hu}, \binits{Y.}},
\oauthor{\bsnm{Lin}, \binits{X.}},
\oauthor{\bsnm{Zhang}, \binits{B.}}, et al.:
Mobilevlm v2: Faster and stronger baseline for vision language model.
arXiv preprint arXiv:2402.03766
(2024)
\end{botherref}
\endbibitem

%%% 61
\bibitem[\protect\citeauthoryear{Zhu et~al.}{2024}]{zhu2024vislinginstruct}
\begin{botherref}
\oauthor{\bsnm{Zhu}, \binits{D.}},
\oauthor{\bsnm{Tang}, \binits{X.}},
\oauthor{\bsnm{Han}, \binits{W.}},
\oauthor{\bsnm{Lu}, \binits{J.}},
\oauthor{\bsnm{Zhao}, \binits{Y.}},
\oauthor{\bsnm{Xing}, \binits{G.}},
\oauthor{\bsnm{Wang}, \binits{J.}},
\oauthor{\bsnm{Yin}, \binits{D.}}:
Vislinginstruct: Elevating zero-shot learning in multi-modal language models
  with autonomous instruction optimization.
arXiv preprint arXiv:2402.07398
(2024)
\end{botherref}
\endbibitem

%%% 62
\bibitem[\protect\citeauthoryear{Dong et~al.}{2024}]{dong2024internlm}
\begin{botherref}
\oauthor{\bsnm{Dong}, \binits{X.}},
\oauthor{\bsnm{Zhang}, \binits{P.}},
\oauthor{\bsnm{Zang}, \binits{Y.}},
\oauthor{\bsnm{Cao}, \binits{Y.}},
\oauthor{\bsnm{Wang}, \binits{B.}},
\oauthor{\bsnm{Ouyang}, \binits{L.}},
\oauthor{\bsnm{Wei}, \binits{X.}},
\oauthor{\bsnm{Zhang}, \binits{S.}},
\oauthor{\bsnm{Duan}, \binits{H.}},
\oauthor{\bsnm{Cao}, \binits{M.}}, et al.:
Internlm-xcomposer2: Mastering free-form text-image composition and
  comprehension in vision-language large model.
arXiv preprint arXiv:2401.16420
(2024)
\end{botherref}
\endbibitem

%%% 63
\bibitem[\protect\citeauthoryear{Ma et~al.}{2022}]{ma2022x}
\begin{bchapter}
\bauthor{\bsnm{Ma}, \binits{Y.}},
\bauthor{\bsnm{Xu}, \binits{G.}},
\bauthor{\bsnm{Sun}, \binits{X.}},
\bauthor{\bsnm{Yan}, \binits{M.}},
\bauthor{\bsnm{Zhang}, \binits{J.}},
\bauthor{\bsnm{Ji}, \binits{R.}}:
\bctitle{X-clip: End-to-end multi-grained contrastive learning for video-text
  retrieval}.
In: \bbtitle{Proceedings of the 30th ACM International Conference on
  Multimedia},
pp. \bfpage{638}--\blpage{647}
(\byear{2022})
\end{bchapter}
\endbibitem

%%% 64
\bibitem[\protect\citeauthoryear{Ma et~al.}{2023}]{ma2023towards}
\begin{barticle}
\bauthor{\bsnm{Ma}, \binits{Y.}},
\bauthor{\bsnm{Ji}, \binits{J.}},
\bauthor{\bsnm{Sun}, \binits{X.}},
\bauthor{\bsnm{Zhou}, \binits{Y.}},
\bauthor{\bsnm{Ji}, \binits{R.}}:
\batitle{Towards local visual modeling for image captioning}.
\bjtitle{Pattern Recognition}
\bvolume{138},
\bfpage{109420}
(\byear{2023})
\end{barticle}
\endbibitem

%%% 65
\bibitem[\protect\citeauthoryear{Ji et~al.}{2022}]{ji2022knowing}
\begin{barticle}
\bauthor{\bsnm{Ji}, \binits{J.}},
\bauthor{\bsnm{Ma}, \binits{Y.}},
\bauthor{\bsnm{Sun}, \binits{X.}},
\bauthor{\bsnm{Zhou}, \binits{Y.}},
\bauthor{\bsnm{Wu}, \binits{Y.}},
\bauthor{\bsnm{Ji}, \binits{R.}}:
\batitle{Knowing what to learn: a metric-oriented focal mechanism for image
  captioning}.
\bjtitle{IEEE Transactions on Image Processing}
\bvolume{31},
\bfpage{4321}--\blpage{4335}
(\byear{2022})
\end{barticle}
\endbibitem

%%% 66
\bibitem[\protect\citeauthoryear{Reid et~al.}{2024}]{reid2024gemini}
\begin{botherref}
\oauthor{\bsnm{Reid}, \binits{M.}},
\oauthor{\bsnm{Savinov}, \binits{N.}},
\oauthor{\bsnm{Teplyashin}, \binits{D.}},
\oauthor{\bsnm{Lepikhin}, \binits{D.}},
\oauthor{\bsnm{Lillicrap}, \binits{T.}},
\oauthor{\bsnm{Alayrac}, \binits{J.-b.}},
\oauthor{\bsnm{Soricut}, \binits{R.}},
\oauthor{\bsnm{Lazaridou}, \binits{A.}},
\oauthor{\bsnm{Firat}, \binits{O.}},
\oauthor{\bsnm{Schrittwieser}, \binits{J.}}, et al.:
Gemini 1.5: Unlocking multimodal understanding across millions of tokens of
  context.
arXiv preprint arXiv:2403.05530
(2024)
\end{botherref}
\endbibitem

%%% 67
\bibitem[\protect\citeauthoryear{Bai et~al.}{2023}]{bai2023qwen}
\begin{botherref}
\oauthor{\bsnm{Bai}, \binits{J.}},
\oauthor{\bsnm{Bai}, \binits{S.}},
\oauthor{\bsnm{Yang}, \binits{S.}},
\oauthor{\bsnm{Wang}, \binits{S.}},
\oauthor{\bsnm{Tan}, \binits{S.}},
\oauthor{\bsnm{Wang}, \binits{P.}},
\oauthor{\bsnm{Lin}, \binits{J.}},
\oauthor{\bsnm{Zhou}, \binits{C.}},
\oauthor{\bsnm{Zhou}, \binits{J.}}:
Qwen-vl: A frontier large vision-language model with versatile abilities.
arXiv preprint arXiv:2308.12966
(2023)
\end{botherref}
\endbibitem

%%% 68
\bibitem[\protect\citeauthoryear{Wang et~al.}{2023}]{wang2023omni}
\begin{bchapter}
\bauthor{\bsnm{Wang}, \binits{H.}},
\bauthor{\bsnm{Chen}, \binits{X.}},
\bauthor{\bsnm{Ni}, \binits{B.}},
\bauthor{\bsnm{Liu}, \binits{Y.}},
\bauthor{\bsnm{Liu}, \binits{J.}}:
\bctitle{Omni aggregation networks for lightweight image super-resolution}.
In: \bbtitle{Proceedings of the IEEE/CVF Conference on Computer Vision and
  Pattern Recognition},
pp. \bfpage{22378}--\blpage{22387}
(\byear{2023})
\end{bchapter}
\endbibitem

%%% 69
\bibitem[\protect\citeauthoryear{Chen et~al.}{2023}]{chen2023masked}
\begin{bchapter}
\bauthor{\bsnm{Chen}, \binits{H.}},
\bauthor{\bsnm{Gu}, \binits{J.}},
\bauthor{\bsnm{Liu}, \binits{Y.}},
\bauthor{\bsnm{Magid}, \binits{S.A.}},
\bauthor{\bsnm{Dong}, \binits{C.}},
\bauthor{\bsnm{Wang}, \binits{Q.}},
\bauthor{\bsnm{Pfister}, \binits{H.}},
\bauthor{\bsnm{Zhu}, \binits{L.}}:
\bctitle{Masked image training for generalizable deep image denoising}.
In: \bbtitle{Proceedings of the IEEE/CVF Conference on Computer Vision and
  Pattern Recognition},
pp. \bfpage{1692}--\blpage{1703}
(\byear{2023})
\end{bchapter}
\endbibitem

%%% 70
\bibitem[\protect\citeauthoryear{Sanghvi et~al.}{2023}]{sanghvi2023structured}
\begin{bchapter}
\bauthor{\bsnm{Sanghvi}, \binits{Y.}},
\bauthor{\bsnm{Mao}, \binits{Z.}},
\bauthor{\bsnm{Chan}, \binits{S.H.}}:
\bctitle{Structured kernel estimation for photon-limited deconvolution}.
In: \bbtitle{Proceedings of the IEEE/CVF Conference on Computer Vision and
  Pattern Recognition},
pp. \bfpage{9863}--\blpage{9872}
(\byear{2023})
\end{bchapter}
\endbibitem

%%% 71
\bibitem[\protect\citeauthoryear{Chen et~al.}{2023}]{chen2023learning}
\begin{bchapter}
\bauthor{\bsnm{Chen}, \binits{X.}},
\bauthor{\bsnm{Li}, \binits{H.}},
\bauthor{\bsnm{Li}, \binits{M.}},
\bauthor{\bsnm{Pan}, \binits{J.}}:
\bctitle{Learning a sparse transformer network for effective image deraining}.
In: \bbtitle{Proceedings of the IEEE/CVF Conference on Computer Vision and
  Pattern Recognition},
pp. \bfpage{5896}--\blpage{5905}
(\byear{2023})
\end{bchapter}
\endbibitem

%%% 72
\bibitem[\protect\citeauthoryear{Wu et~al.}{2023}]{wu2023ridcp}
\begin{bchapter}
\bauthor{\bsnm{Wu}, \binits{R.-Q.}},
\bauthor{\bsnm{Duan}, \binits{Z.-P.}},
\bauthor{\bsnm{Guo}, \binits{C.-L.}},
\bauthor{\bsnm{Chai}, \binits{Z.}},
\bauthor{\bsnm{Li}, \binits{C.}}:
\bctitle{Ridcp: Revitalizing real image dehazing via high-quality codebook
  priors}.
In: \bbtitle{Proceedings of the IEEE/CVF Conference on Computer Vision and
  Pattern Recognition},
pp. \bfpage{22282}--\blpage{22291}
(\byear{2023})
\end{bchapter}
\endbibitem

%%% 73
\bibitem[\protect\citeauthoryear{Guo et~al.}{2023}]{guo2023shadowdiffusion}
\begin{bchapter}
\bauthor{\bsnm{Guo}, \binits{L.}},
\bauthor{\bsnm{Wang}, \binits{C.}},
\bauthor{\bsnm{Yang}, \binits{W.}},
\bauthor{\bsnm{Huang}, \binits{S.}},
\bauthor{\bsnm{Wang}, \binits{Y.}},
\bauthor{\bsnm{Pfister}, \binits{H.}},
\bauthor{\bsnm{Wen}, \binits{B.}}:
\bctitle{Shadowdiffusion: When degradation prior meets diffusion model for
  shadow removal}.
In: \bbtitle{Proceedings of the IEEE/CVF Conference on Computer Vision and
  Pattern Recognition},
pp. \bfpage{14049}--\blpage{14058}
(\byear{2023})
\end{bchapter}
\endbibitem

%%% 74
\bibitem[\protect\citeauthoryear{Kong et~al.}{2022}]{kong2022reflash}
\begin{bchapter}
\bauthor{\bsnm{Kong}, \binits{X.}},
\bauthor{\bsnm{Liu}, \binits{X.}},
\bauthor{\bsnm{Gu}, \binits{J.}},
\bauthor{\bsnm{Qiao}, \binits{Y.}},
\bauthor{\bsnm{Dong}, \binits{C.}}:
\bctitle{Reflash dropout in image super-resolution}.
In: \bbtitle{Proceedings of the IEEE/CVF Conference on Computer Vision and
  Pattern Recognition},
pp. \bfpage{6002}--\blpage{6012}
(\byear{2022})
\end{bchapter}
\endbibitem

%%% 75
\bibitem[\protect\citeauthoryear{Batifol et~al.}{2025}]{batifol2025flux}
\begin{botherref}
\oauthor{\bsnm{Batifol}, \binits{S.}},
\oauthor{\bsnm{Blattmann}, \binits{A.}},
\oauthor{\bsnm{Boesel}, \binits{F.}},
\oauthor{\bsnm{Consul}, \binits{S.}},
\oauthor{\bsnm{Diagne}, \binits{C.}},
\oauthor{\bsnm{Dockhorn}, \binits{T.}},
\oauthor{\bsnm{English}, \binits{J.}},
\oauthor{\bsnm{English}, \binits{Z.}},
\oauthor{\bsnm{Esser}, \binits{P.}},
\oauthor{\bsnm{Kulal}, \binits{S.}}, et al.:
Flux. 1 kontext: Flow matching for in-context image generation and editing in
  latent space.
arXiv e-prints,
2506
(2025)
\end{botherref}
\endbibitem

%%% 76
\bibitem[\protect\citeauthoryear{Wu et~al.}{2025}]{wu2025qwen}
\begin{botherref}
\oauthor{\bsnm{Wu}, \binits{C.}},
\oauthor{\bsnm{Li}, \binits{J.}},
\oauthor{\bsnm{Zhou}, \binits{J.}},
\oauthor{\bsnm{Lin}, \binits{J.}},
\oauthor{\bsnm{Gao}, \binits{K.}},
\oauthor{\bsnm{Yan}, \binits{K.}},
\oauthor{\bsnm{Yin}, \binits{S.-m.}},
\oauthor{\bsnm{Bai}, \binits{S.}},
\oauthor{\bsnm{Xu}, \binits{X.}},
\oauthor{\bsnm{Chen}, \binits{Y.}}, et al.:
Qwen-image technical report.
arXiv preprint arXiv:2508.02324
(2025)
\end{botherref}
\endbibitem

%%% 77
\bibitem[\protect\citeauthoryear{Bai et~al.}{2025}]{bai2025qwen3}
\begin{botherref}
\oauthor{\bsnm{Bai}, \binits{S.}},
\oauthor{\bsnm{Cai}, \binits{Y.}},
\oauthor{\bsnm{Chen}, \binits{R.}},
\oauthor{\bsnm{Chen}, \binits{K.}},
\oauthor{\bsnm{Chen}, \binits{X.}},
\oauthor{\bsnm{Cheng}, \binits{Z.}},
\oauthor{\bsnm{Deng}, \binits{L.}},
\oauthor{\bsnm{Ding}, \binits{W.}},
\oauthor{\bsnm{Gao}, \binits{C.}},
\oauthor{\bsnm{Ge}, \binits{C.}}, et al.:
Qwen3-vl technical report.
arXiv preprint arXiv:2511.21631
(2025)
\end{botherref}
\endbibitem

%%% 78
\bibitem[\protect\citeauthoryear{Touvron et~al.}{2023}]{touvron2023llama}
\begin{botherref}
\oauthor{\bsnm{Touvron}, \binits{H.}},
\oauthor{\bsnm{Lavril}, \binits{T.}},
\oauthor{\bsnm{Izacard}, \binits{G.}},
\oauthor{\bsnm{Martinet}, \binits{X.}},
\oauthor{\bsnm{Lachaux}, \binits{M.-A.}},
\oauthor{\bsnm{Lacroix}, \binits{T.}},
\oauthor{\bsnm{Rozi{\`e}re}, \binits{B.}},
\oauthor{\bsnm{Goyal}, \binits{N.}},
\oauthor{\bsnm{Hambro}, \binits{E.}},
\oauthor{\bsnm{Azhar}, \binits{F.}}, et al.:
Llama: Open and efficient foundation language models.
arXiv preprint arXiv:2302.13971
(2023)
\end{botherref}
\endbibitem

%%% 79
\bibitem[\protect\citeauthoryear{Schuhmann et~al.}{2022}]{schuhmann2022laion}
\begin{barticle}
\bauthor{\bsnm{Schuhmann}, \binits{C.}},
\bauthor{\bsnm{Beaumont}, \binits{R.}},
\bauthor{\bsnm{Vencu}, \binits{R.}},
\bauthor{\bsnm{Gordon}, \binits{C.}},
\bauthor{\bsnm{Wightman}, \binits{R.}},
\bauthor{\bsnm{Cherti}, \binits{M.}},
\bauthor{\bsnm{Coombes}, \binits{T.}},
\bauthor{\bsnm{Katta}, \binits{A.}},
\bauthor{\bsnm{Mullis}, \binits{C.}},
\bauthor{\bsnm{Wortsman}, \binits{M.}}, \betal:
\batitle{Laion-5b: An open large-scale dataset for training next generation
  image-text models}.
\bjtitle{Advances in Neural Information Processing Systems}
\bvolume{35},
\bfpage{25278}--\blpage{25294}
(\byear{2022})
\end{barticle}
\endbibitem

%%% 80
\bibitem[\protect\citeauthoryear{Hall et~al.}{2024}]{hall2024evalgim}
\begin{botherref}
\oauthor{\bsnm{Hall}, \binits{M.}},
\oauthor{\bsnm{Ma{\~n}as}, \binits{O.}},
\oauthor{\bsnm{Askari-Hemmat}, \binits{R.}},
\oauthor{\bsnm{Ibrahim}, \binits{M.}},
\oauthor{\bsnm{Ross}, \binits{C.}},
\oauthor{\bsnm{Astolfi}, \binits{P.}},
\oauthor{\bsnm{Ifriqi}, \binits{T.B.}},
\oauthor{\bsnm{Havasi}, \binits{M.}},
\oauthor{\bsnm{Benchetrit}, \binits{Y.}},
\oauthor{\bsnm{Ullrich}, \binits{K.}}, et al.:
Evalgim: A library for evaluating generative image models.
arXiv preprint arXiv:2412.10604
(2024)
\end{botherref}
\endbibitem

\end{thebibliography}

\end{document}